\documentclass[a4paper,fleqn]{cas-dc}

\usepackage[authoryear]{natbib}

\usepackage{pifont}
\usepackage{microtype} 
\usepackage[dvipsnames]{xcolor}
\usepackage[colorlinks]{hyperref}
\usepackage[skip=2pt]{subcaption}
\usepackage[all]{hypcap} 
\usepackage{algorithm}
\usepackage{algpseudocode}
\usepackage{placeins}
\newcommand{\cmark}{\ding{51}}
\newcommand{\xmark}{\ding{55}}
\hypersetup{citecolor=MidnightBlue,linkcolor=MidnightBlue,urlcolor=MidnightBlue}
\graphicspath{{./figures/}}
\DeclareMathOperator*{\argmax}{\arg\!\max}

\begin{document}
\let\WriteBookmarks\relax
\def\floatpagepagefraction{1}
\def\textpagefraction{.001}
\shorttitle{Analysing Deep Reinforcement Learning Agents Trained with Domain
Randomisation}
\shortauthors{T Dai et~al.}

\title[mode = title]{Analysing Deep Reinforcement Learning Agents Trained with Domain
Randomisation}                      

\author[1]{Tianhong Dai}
\cormark[1]
\ead{tianhong.dai15@imperial.ac.uk}
\author[1]{Kai Arulkumaran}
\cormark[1]
\ead{kailash.arulkumaran13@imperial.ac.uk}

\author[1]{Tamara Gerbert}
\author[1]{Samyakh Tukra}
\author[1]{Feryal Behbahani}
\author[1]{Anil Anthony Bharath}
\address[1]{BICI-Lab, Department of Bioengineering, Imperial College London,
Exhibition Road, London SW7 2AZ, United Kingdom}

\cortext[cor1]{Equal contributions. Correspondence to: BICI-Lab, Department of Bioengineering, Imperial College London,
Exhibition Road, London SW7 2AZ, United Kingdom.}

\begin{abstract}
Deep reinforcement learning has the potential to train robots to perform
complex tasks in the real world without requiring accurate models of the
robot or its environment. A practical approach is to train agents in
simulation, and then transfer them to the real world. One popular method
for achieving transferability is to use domain randomisation, which
involves randomly perturbing various aspects of a simulated environment
in order to make trained agents robust to the reality gap. However, less
work has gone into understanding such agents---which are deployed in the
real world---beyond task performance. In this work we examine such
agents, through qualitative and quantitative comparisons between agents
trained with and without visual domain randomisation. We train agents
for Fetch and Jaco robots on a visuomotor control task and evaluate how
well they generalise using different testing conditions. Finally, we
investigate the internals of the trained agents by using a suite of
interpretability techniques. Our results show that the primary outcome
of domain randomisation is more robust, entangled representations,
accompanied with larger weights with greater spatial structure;
moreover, the types of changes are heavily influenced by the task setup
and presence of additional proprioceptive inputs. Additionally, we
demonstrate that our domain randomised agents require higher sample
complexity, can overfit and more heavily rely on recurrent processing.
Furthermore, even with an improved saliency method introduced in this
work, we show that qualitative studies may not always correspond with
quantitative measures, necessitating the combination of inspection tools
in order to provide sufficient insights into the behaviour of trained
agents.
\end{abstract}




\maketitle

\hypertarget{introduction}{%
\section{Introduction}\label{introduction}}

Deep reinforcement learning (DRL) is currently one of the most prominent
subfields in AI, with applications to many domains
\citep{arulkumaran2017deep, franccois2018introduction}. One of the most
enticing possibilities that DRL affords is the ability to train robots
to perform complex tasks in the real world, all from raw sensory inputs.
For instance, while robotics has traditionally relied on hand-crafted
pipelines, each performing well-defined estimation tasks -- such as
ground-plane estimation, object detection, segmentation and
classification, \citep{kragic2009vision, martinez2014taxonomy} -- it is
now possible to learn visual perception and control in an ``end-to-end''
fashion
\citep{levine2016end, gu2017deep, zhu2017target, levine2018learning},
without explicit specification and training of networks for specific
sub-tasks.

A major advantage of using reinforcement learning (RL) versus the more
traditional approach to robotic system design based on optimal control
is that the latter requires a transition model for the task in order to
solve for the optimal sequence of actions. While optimal control, when
applicable, is more efficient, modelling certain classes of objects
(e.g., deformable objects) can require expensive simulation steps, and
often physical parameters (e.g., frictional coefficients) of real
objects that are not known in detail. Instead, approaches that use RL
can learn a direct mapping from observations to the optimal sequence of
actions, purely through interacting with the environment. Through the
powerful function approximation capabilities of neural networks (NNs),
deep learning (DL) has allowed RL algorithms to scale to domains with
significantly more complex input and action spaces than previously
considered tractable.

\begin{figure}
  \centering
  \includegraphics[width=\linewidth]{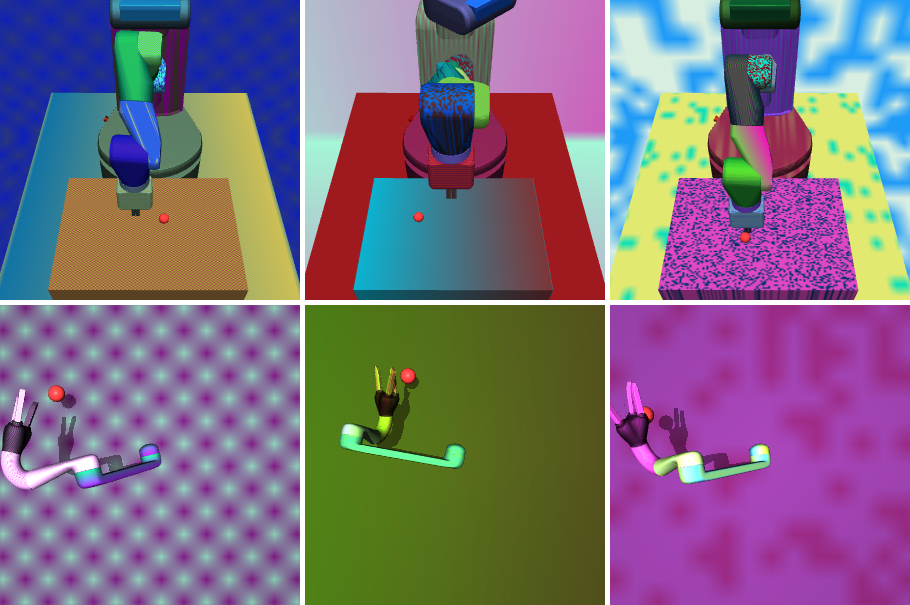}
  \caption{Examples of visual domain randomisation in our Fetch (top) and Jaco (bottom) robotics experiments.}
  \label{fig:dr_example}
\end{figure}
\setlength{\textfloatsep}{5pt}

The downside is that while DRL algorithms can learn complex control
policies from raw sensory data, they typically have poor sample
complexity. In practice, this means training DRL algorithms in
simulators before deploying them on real robots, which then introduces a
\emph{reality gap} \citep{jakobi1995noise} between the simulated and
real worlds---including not just differences in physics, but also visual
appearance. There are several solutions to this problem, including
fine-tuning a DRL agent on the real world \citep{rusu2017sim},
performing system identification to reduce the domain gap
\citep{chebotar2018closing}, and explicitly performing domain adaptation
\citep{tzeng2015towards, bousmalis2018using}.

One solution to increase the robustness of agents to potential
differences between simulators and the real world is to use \emph{domain
randomisation} (DR; pictured in Figure \ref{fig:dr_example}), in which
various properties of the simulation are varied, altering anything from
the positions or dynamical properties of objects to their visual
appearance. This extension of data augmentation to RL environments has
been used to successfully train agents for a range of different robots,
including robotic arms \citep{tobin2017domain, james2017transferring},
quadcopters \citep{sadeghi2017cad2rl}, and even humanoid robotic hands
\citep{andrychowicz2018learning}. While early uses of DR
\citep{tobin2017domain, james2017transferring} did not include
transition dynamics as a random property, we note that ``dynamics
randomisation'' \citep{peng2018sim} can now also be considered part of
the standard DR pipeline. Common practice is to design DR to incorporate
as many variations as possible, such that the real world would, ideally,
be a ``subset'' of the set of DR environments.

When the primary aim of this line of research is to enable the training
of agents that perform well in the real world, there is an obvious need
to characterise how these agents behave before they can be deployed ``in
the wild''. In particular, one can study how well these agents
\emph{generalise}---a criterion that has received considerable interest
in the DRL community recently
\citep{zhang2018dissection, zhang2018study, justesen2018procedural, witty2018measuring, packer2018assessing, cobbe2018quantifying, zhao2019investigating}.
To do so, we can construct unit tests that not only reflect the
conditions under which the agent has been trained, but also extrapolate
beyond; for instance, James et al. \citeyearpar{james2017transferring}
studied the test-time performance of agents trained with DR in the
presence of distractors or changed illumination. While adding robustness
to these extrapolation tests can be done by simply training under the
new conditions, we are interested in developing general procedures that
would still be useful when this option is not available. As we show
later (in Subsection \ref{sec:domain_shift}), depending on the training
conditions, we can even observe a failure of agents trained with DR to
generalise to the much simpler default visuals of the simulator.

While unit tests provide a quantitative measure by which we can probe
the performance of trained agents under various conditions, they treat
the trained agents as black boxes. However, with full access to the
internals of the trained models and even control over the training
process, we can dive even further into the models. Using common
interpretability tools such as saliency maps
\citep{morch1995visualization, simonyan2013deep, zeiler2014visualizing, selvaraju2017grad, sundararajan2017axiomatic}
and dimensionality reduction methods
\citep{pearson1901liii, maaten2008visualizing, mcinnes2018umap} for
visualising NN activations \citep{rauber2017visualizing}, we can obtain
information on why agents act the way they do. The results of these
methods work in tandem with unit tests, as matching performance to the
qualitative results allows us to have greater confidence in interpreting
the latter; in fact, this process allowed us to debug and improve upon
an existing saliency map method, as detailed in Subsection
\ref{sec:saliency_maps}. Through a combination of existing and novel
methods, we present here a more extensive look into DRL agents that have
been trained to perform control tasks using both visual and
proprioceptive inputs. In particular, under our set of experimental
conditions, we show that our agents trained with visual DR:

\begin{itemize}
\item require more representational learning capacity (Subsection \ref{sec:layer_ablations}),
\item are more robust to visual changes in the scene, exhibiting generalisation to unseen local/global perturbations (Subsection \ref{sec:test_scenarios}),
\item use a smaller set of more reliable visual cues when not provided proprioceptive inputs (Subsection \ref{sec:saliency_maps}),
\item more heavily rely on recurrent processing (Subsection \ref{sec:recurrent_ablation}),
\item have filters that have higher norms or greater spatial structure (Subsection \ref{sec:statistical_and_structural}), which respond to more complex spatial patterns (Subsection \ref{sec:activation_maximisation}),
\item learn more robust (Subsection \ref{sec:unit_ablations}) and \textit{entangled} \citep{frosst2019analyzing} representations (Subsection \ref{sec:activation_analysis}),
\item and can ``overfit'' to DR visuals (Subsection \ref{sec:domain_shift}).
\end{itemize}

\hypertarget{methods}{%
\section{Methods}\label{methods}}

\hypertarget{reinforcement-learning}{%
\subsection{Reinforcement Learning}\label{reinforcement-learning}}

In RL, the aim is to learn optimal behaviour in sequential decision
problems \citep{sutton2018reinforcement}, such as finding the best
trajectory for a manipulation task. It can formally be described by a
Markov decision process (MDP), whereby at every timestep \(t\) the agent
receives the state of the environment \(\mathbf{s}_t\), performs an
action \(\mathbf{a}_t\) sampled from its policy
\(\pi(\mathbf{a}_t|\mathbf{s}_t)\) (potentially parameterised by weights
\(\theta\)), and then receives the next state \(\mathbf{s}_{t+1}\) along
with a scalar reward \(r_{t+1}\). The goal of RL is to find the optimal
policy, \(\pi^*\), which maximises the expected return: \begin{align*}
\mathbb{E}[R_{t=0}] = \mathbb{E}\left[\sum_{t=0}^{T-1} \gamma^t r_{t+1} \right],
\end{align*} \noindent where in practice a discount value
\(\gamma \in [0, 1)\) is used to weight earlier rewards more heavily and
reduce the variance of the return over an episode of interaction with
the environment, ending at timestep \(T\).

Policy search methods, which are prevalent in robotics
\citep{deisenroth2013survey}, are one way of finding the optimal policy.
In particular, policy gradient methods that are commonly used with NNs
perform gradient ascent on \(\mathbb{E}_\pi[R]\) to optimise a
parameterised policy \(\pi(\cdot; \theta)\)
\citep{williams1991function}. Other RL methods rely on value functions,
which represent the future expected return from following a policy from
a given state: \(V_\pi(\mathbf{s}_t) = \mathbb{E}_\pi[R_t]\). The
combination of learned policy and value functions are known as
actor-critic methods, and utilise the critic (value function) in order
to reduce the variance of the training signal to the actor (policy)
\citep{barto1983neuronlike}. Instead of directly maximising the return
\(R_t\), the policy can then be trained to maximise the advantage
\(A_t = R_t - V_t\) (the difference between the empirical and predicted
return).

We note that in practice many problems are better described as
partially-observed MDPs, where the observation received by the agent
does not contain full information about the state of the environment. In
visuomotor object manipulation this can occur as the end effector blocks
the line of sight between the camera and the object, causing
self-occlusion. A common solution to this is to utilise recurrent
connections within the NN, allowing information about observations to
propagate from the beginning of the episode to the current timestep
\citep{wierstra2007solving}.

\hypertarget{proximal-policy-optimisation}{%
\subsubsection{Proximal Policy
Optimisation}\label{proximal-policy-optimisation}}

For our experiments we train our agents using proximal policy
optimisation (PPO) \citep{schulman2017proximal}, a widely used and
performant RL algorithm.\footnote{In particular, PPO has been used with
  DR to train a policy that was applied to a Shadow Dexterous Hand in
  the real world \citep{andrychowicz2018learning}.} Rather than training
the policy to maximise the advantage directly, PPO instead maximises the
surrogate objective: \begin{align*}
\mathcal{L}_{clip} = \mathbb{E}_t\left[\min(\rho_t(\theta)A_t, \text{clip}(\rho_t(\theta), 1 - \epsilon, 1 + \epsilon)A_t)\right],
\end{align*} with \begin{align*}
\rho_t(\theta) = \frac{\pi(\mathbf{a}_t|\mathbf{s}_t; \theta)}{\pi_{old}(\mathbf{a}_t|\mathbf{s}_t; \theta_{old})},
\end{align*} \noindent where \(\rho_t(\theta)\) is the ratio between the
current policy and the old policy, \(\epsilon\) is the clip ratio which
restricts the change in the policy distribution, and \(A_{t}\) is the
advantage, which we choose to be the Generalised Advantage Estimate
(GAE): \begin{align*}
A_{t} = \delta_t + (\gamma\lambda)\delta_{t+1} + \ldots + (\gamma\lambda)^{T-t+1}\delta_{T-1},
\end{align*} \noindent that mixes Monte Carlo returns \(R_t\) and
temporal difference errors
\(\delta_t = r_t + \gamma V_\pi(\mathbf{s}_{t+1}) - V_\pi(\mathbf{s}_t)\)
with hyperparameter \(\lambda\) \citep{schulman2015high}.

In practice, both the actor and the critic can be combined into a single
NN with two output heads, parameterised by \(\theta\)
\citep{mnih2016asynchronous}. The full PPO objective involves maximising
\(\mathcal{L}_{clip}\), minimising the squared error between the learned
value function and the empirical return: \begin{align*}
\mathcal{L}_{value} = \mathbb{E}_t\left[(V_\pi(\mathbf{s}_t; \theta) - R_t)^2\right],
\end{align*} \noindent and maximising the (Shannon) entropy of the
policy, which for discrete action sets of size \(|\mathcal{A}|\), is
defined as: \begin{align*}
\mathcal{L}_{entropy} = \mathbb{E}_t\left[-\sum_{n=1}^{|\mathcal{A}|} \pi(a_n|\mathbf{s}_t; \theta) \log\left(\pi(a_n|\mathbf{s}_t; \theta)\right)\right].
\end{align*} \noindent Entropy regularisation prevents the policy from
prematurely collapsing to a deterministic solution and aids exploration
\citep{williams1991function}.

Using a parallelised implementation of PPO, we are able to train our
agents to strong performance on all training setups within a reasonable
amount of time. Training details are described in Subsection
\ref{sec:networks_training}.

\hypertarget{neural-network-interpretability}{%
\subsection{Neural Network
Interpretability}\label{neural-network-interpretability}}

The recent success of machine learning (ML) methods has led to a renewed
interest in trying to interpret trained models, whereby an explanation
of a model's ``reasoning'' may be used as a way to understand other
properties, such as safety, fairness, reliability, or simply to provide
an explanation of the model's behaviour \citep{doshi2017towards}. In
this work, we are primarily concerned with scientific understanding, but
our considerations are grounded in other properties necessary for
eventual real-world deployment, such as robustness.

The challenge that we face is that, unlike other ML algorithms that are
considered interpretable by design (such as decision trees or nearest
neighbours \citep{freitas2014comprehensible}), standard NNs are
generally considered \emph{black boxes}. However, given decades of
research into methods for interpreting NNs
\citep{morch1995visualization, craven1996extracting}, we now have a
range of techniques at our disposal \citep{guidotti2018survey}. Beyond
simply looking at test performance (a measure of interpretability in its
own right \citep{doshi2017towards}), we will focus on a variety of
techniques that will let us examine trained NNs both in the context of,
and independently of, task performance. In particular, we discuss
saliency maps (Subsection \ref{sec:saliency_maps}), activation
maximisation (Subsection \ref{sec:act_max}), weight visualisations
(Subsection \ref{sec:weight_vis}), statistical and structural weight
characterisations (Subsection \ref{sec:stat_struct_weight}), unit
ablations (Subsection \ref{sec:unit_abl}), layer re-initialisation
(Subsection \ref{sec:layer_abl}) and activation analysis (Subsection
\ref{sec:act_an}). By utilising a range of techniques we hope to cover
various points along the trade-off between fidelity and interpretability
\citep{ribeiro2016should}.

\hypertarget{saliency-maps}{%
\subsubsection{Saliency Maps}\label{saliency-maps}}

\label{sec:saliency_maps}

\begin{figure}
  \centering
  \begin{subfigure}{0.32\columnwidth}
    \includegraphics[width=\linewidth]{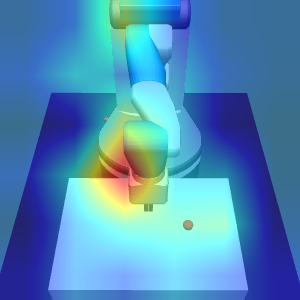}
    \subcaption{Fetch (GC)}
  \end{subfigure}
  \begin{subfigure}{0.32\columnwidth}
    \includegraphics[width=\linewidth]{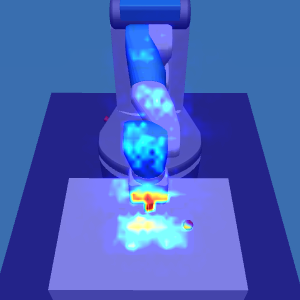}
    \subcaption{Fetch (IG)}
  \end{subfigure}
  \begin{subfigure}{0.32\columnwidth}
    \includegraphics[width=\linewidth]{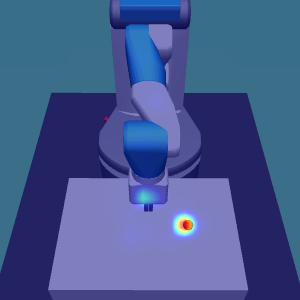}
    \subcaption{Fetch (occlusion)}
  \end{subfigure}
\\
  \begin{subfigure}{0.32\columnwidth}
    \includegraphics[width=\linewidth]{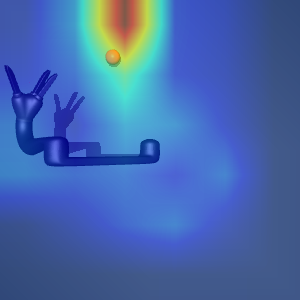}
    \subcaption{Jaco (GC)}
  \end{subfigure}
  \begin{subfigure}{0.32\columnwidth}
    \includegraphics[width=\linewidth]{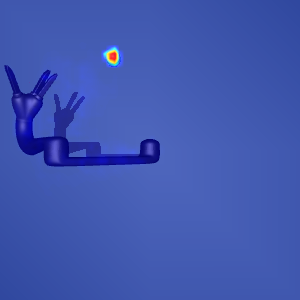}
    \subcaption{Jaco (IG)}
  \end{subfigure}
  \begin{subfigure}{0.32\columnwidth}
    \includegraphics[width=\linewidth]{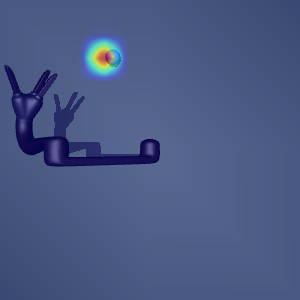}
    \subcaption{Jaco (occlusion)}
  \end{subfigure}
  \caption{Comparison of saliency map methods on Fetch (a-c) and Jaco (d-f) tasks. While Grad-CAM (GC; a, d) and IG (b, e) create somewhat interpretable saliency maps, the occlusion method (c, f) subjectively works best overall.}
  \label{fig:saliency_comparison}
\end{figure}

Saliency maps are one of the most common techniques used for
understanding the decisions made by NNs, and in particular,
convolutional NNs (CNNs). The most common methods are gradient-based,
and utilise the derivative of the network output with respect to the
inputs, indicating, for images, how changing the pixel intensities at
each location will affect the output \citep{simonyan2013deep}. We
investigated the use of two popular, more advanced variants of this
technique---gradient-weighted class activation mapping (Grad-CAM)
\citep{selvaraju2017grad} and integrated gradients (IG)
\citep{sundararajan2017axiomatic}---as well as an occlusion-based
method, which masks parts of the image and performs a sensitivity
analysis with respect to the change in the network's outputs
\citep{zeiler2014visualizing}. As shown in Figure
\ref{fig:saliency_comparison}, the latter technique gave the most
``interpretable'' saliency maps across all trained agents, so we utilise
it alone when analysing our trained agents in latter sections.\footnote{This
  observation is also in line with prior work on interpreting DRL agents
  \citep{greydanus2018visualizing}.} In light of the unreliability of
saliency methods \citep{kindermans2017reliability}, we include a
discussion and comparison of these methods to illuminate the importance
of checking the outputs of qualitative methods. As a final remark we
note that clustering methods have been used to automatically find groups
of strategies via collections of saliency maps
\citep{lapuschkin2019unmasking}, but, given the relative visual
simplicity of our tasks, highlighting individual examples is
sufficiently informative.

\hypertarget{cam}{%
\paragraph{CAM}\label{cam}}

The class average map (CAM) \citep{zhou2016learning} was developed as a
saliency method for CNNs with global average pooling
\citep{lin2013network} trained for the purpose of object recognition.
The value of the saliency map \(S^c_{m,n}\) for class \(c\) at spatial
location \(m, n\) is calculated by summing over the activations
\(\mathbf{A}^k\) of the final convolutional layer (with \(k\) channels)
and the corresponding class weights \(w_k^c\): \begin{align*}
S^c_{m,n} = \sum_{k}w_{k}^{c}A^k_{m,n}
\end{align*}

\hypertarget{grad-cam}{%
\paragraph{Grad-CAM}\label{grad-cam}}

Given a network \(F\) and input \(\mathbf{x}\), Grad-CAM extends CAM
from fully-convolutional NNs to generic CNNs by instead constructing
class weights \(\omega_k^c\) by using the partial derivative for the
output of a class \(c\), \(\partial F(\mathbf{x})^c\), with respect to
the \(k\) feature maps \(\mathbf{A}^k\) of any convolutional layer. The
Grad-CAM saliency map for a class, \(\mathbf{S}^c\), is the positive
component of the linear combination of class weights \(\omega_k^c\) and
feature maps \(\mathbf{A}^k\): \begin{align*}
\mathbf{S}^c = \max\left(\sum_k \omega_k^c \mathbf{A}^k, 0\right),\\
\text{with } \omega_k^c = \frac{1}{mn}\sum_m\sum_n \frac{\partial F(\mathbf{x})^c}{\partial A_{m,n}^k},
\end{align*} \noindent where \(\omega_k^c\) is formed by averaging over
spatial locations \(m, n\).

\begin{figure}
  \centering
  \begin{subfigure}{0.32\columnwidth}
    \includegraphics[width=\linewidth]{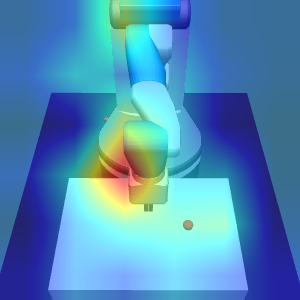}
    \subcaption{Action 1}
  \end{subfigure}
  \begin{subfigure}{0.32\columnwidth}
    \includegraphics[width=\linewidth]{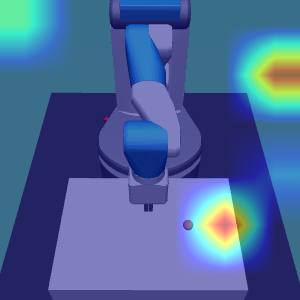}
    \subcaption{Action 2}
  \end{subfigure}
  \begin{subfigure}{0.32\columnwidth}
    \includegraphics[width=\linewidth]{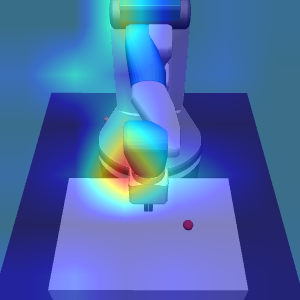}
    \subcaption{Action 3}
  \end{subfigure}
  \caption{Grad-CAM saliency maps for all actions (up/down, left/right, forward/backward) for a trained Fetch agent.}
  \label{fig:grad_cam}
\end{figure}

In place of a given class \(c\), we use Grad-CAM to create a saliency
map per (output) action (Figure \ref{fig:grad_cam}). As in this case, it
is not always clear how to interpret Grad-CAM saliency maps for our
trained agents. There are many reasons such techniques might ``fail'',
such as the mixing of both positive and negative contributions towards
the network outputs \citep{springenberg2015striving,bach2015pixel}.

\hypertarget{integrated-gradients}{%
\paragraph{Integrated Gradients}\label{integrated-gradients}}

Sundarajan et al. \citeyearpar{sundararajan2017axiomatic} proposed that
attribution methods (saliency maps in our case) should be:

\begin{description}
\item [Sensitive] If an input and the baseline differ in one feature and have different outputs, the differing feature should have a non-zero attribution
\item [Invariant to implementation] Attributions should be identical for two functionally equivalent models
\end{description}

Prior gradient-based methods break the first property. Their method, IG,
achieves both by constructing the saliency value \(S_n\) for each input
dimension \(n\) from the path integral of the gradients along the linear
interpolation between input \(\mathbf{x}\) and a baseline input
\(\mathbf{x}^{base}\): \begin{align*}
S_m = \left(x_n - x_n^{base}\right)\int_{\alpha = 0}^1\frac{\partial F\left(\mathbf{x}^{base} + \alpha\left(\mathbf{x} - \mathbf{x}^{base}\right)\right)}{\partial x_n} d\alpha.
\end{align*} \noindent Although Sundarajan et al.
\citeyearpar{sundararajan2017axiomatic} suggested that a black image can
be used as the baseline, we found that using the (dataset) \emph{average
input}, provided superior results. Contemporaneous work has examined the
use of more advanced baselines for gradient-based saliency map methods
\citep{sturmfels2020visualizing}.

\hypertarget{occlusion}{%
\paragraph{Occlusion}\label{occlusion}}

As an alternative to gradient-based methods, Zeiler et al.
\citeyearpar{zeiler2014visualizing} proposed running a (grey, square)
mask over the input and tracking how the network's outputs change in
response. Greydanus et al. \citeyearpar{greydanus2018visualizing}
applied this method to understanding actor-critic-based DRL agents,
using the resulting saliency maps to examine strong and overfitting
policies; they however noted that a grey square may be perceived as part
of a grey object, and instead used a localised Gaussian blur to add
``spatial uncertainty''. The saliency value for each input location is
the Euclidean distance between the original output\footnote{In practice
  taken to be the logits for a categorical policy. We considered that
  the Kullback-Leibler divergence between policy distributions might be
  more meaningful, but found that it produces qualitatively similar
  saliency maps.} and the output given the input
\(\mathbf{x}_{m, n}^{occ}\) which has been occluded at location
\((m, n)\): \begin{align*}
S_{m, n} = \lVert F(\mathbf{x}) - F(\mathbf{x}_{m, n}^{occ}) \rVert_2,
\end{align*} \noindent where \(\lVert\cdot\rVert_p\) denotes the
\(\ell_p\)-norm.

\begin{figure}
  \centering
  \begin{subfigure}{0.32\columnwidth}
    \includegraphics[width=\linewidth]{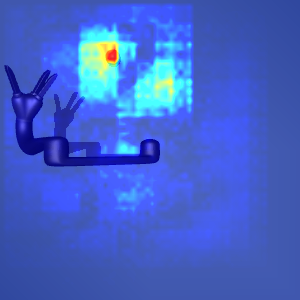}
    \subcaption{IG (black $\mathbf{x}^{base}$)}
  \end{subfigure}
  \begin{subfigure}{0.32\columnwidth}
    \includegraphics[width=\linewidth]{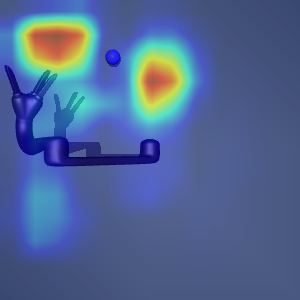}
    \subcaption{Occ. (Gaussian)}
  \end{subfigure}
  \begin{subfigure}{0.32\columnwidth}
    \includegraphics[width=\linewidth]{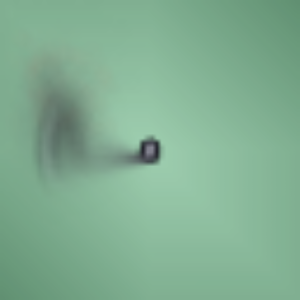}
    \subcaption{Avg. $\mathbf{x}^{base}$}
  \end{subfigure}

  \begin{subfigure}{0.32\columnwidth}
    \includegraphics[width=\linewidth]{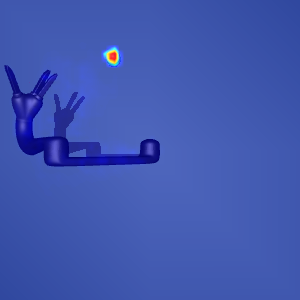}
    \subcaption{IG (avg. $\mathbf{x}^{base}$)}
  \end{subfigure}
  \begin{subfigure}{0.32\columnwidth}
    \includegraphics[width=\linewidth]{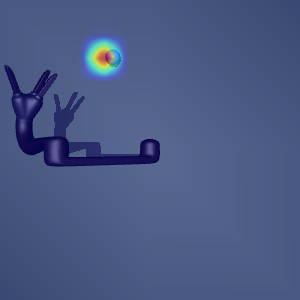}
    \subcaption{Occ. (avg. $\mathbf{x}^{base}$)}
  \end{subfigure}
  \begin{subfigure}{0.32\columnwidth}
    \includegraphics[width=\linewidth]{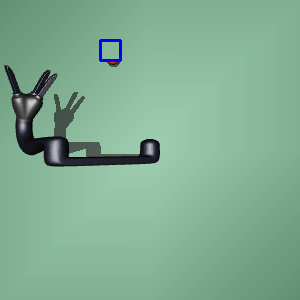}
    \subcaption{Avg. $\mathbf{x}^{base}$ mask}
  \end{subfigure}
  \caption{Saliency map methods (a, b) improved (d, e) by the use of an \textit{average image} (c, f). IG with the \textit{default black baseline} (a) and occlusion (occ.) with the \textit{default Gaussian blur} (b) show additional artifacts. By creating an average image (c) from a large set of trajectories, we can form an improved baseline for IG (d) or occlusion (e); the usage of the average image as the occlusion (with a blue outline added for emphasis) is pictured in (f).}
  \label{fig:saliency_baseline}
\end{figure}

However, we found that certain trained agents sometimes confused the
blurred location with the target location---a failing of the attribution
method against noise/distractors \citep{kindermans2016investigating},
and not necessarily the model itself. Motivated by the methods that
compute interpretations against reference inputs
\citep{bach2015pixel, ribeiro2016should, shrikumar2017learning, sundararajan2017axiomatic, lundberg2017unified},
we replaced the Gaussian blur with a mask\footnote{Replacing a circular
  region of 5px radius around the \((m, n)\) location.} derived from a
baseline input, which roughly represents what the model would expect to
see on average. Intuitively, this acts as a counterfactual, revealing
what would happen if the specific part of the input was not there. For
this we averaged over frames collected from our standard evaluation
protocol (see Subsection \ref{sec:environments} for details), creating
an average input to be used as an improved baseline for IG, as well as
the source of the mask for the occlusion-based method (Figure
\ref{fig:saliency_baseline}). Unless specified otherwise, we use our
average input baseline for all IG and occlusion-based saliency maps.

\hypertarget{activation-maximisation}{%
\subsubsection{Activation Maximisation}\label{activation-maximisation}}

\label{sec:act_max}

Gradients can also be used to try and visualise what maximises the
activation of a given neuron/channel. This can be formulated as an
optimisation problem, using projected\footnote{After every gradient step
  the input is clamped back to within \([0, 1]\).} gradient ascent in
the input space \citep{erhan2009visualizing}. Although this would
ideally show what a neuron/channel is selective for, unconstrained
optimisation may end up in solutions far from the training manifold
\citep{mahendran2015understanding}, and so a variety of regularisation
techniques have been suggested for making qualitatively better
visualisations. We experimented with some of the ``weak regularisers''
\citep{olah2017feature}, and found that a combination of frequency
penalisation (Gaussian blur) \citep{nguyen2015deep} and transformation
robustness (random scaling and translation/jitter)
\citep{mordvintsev2015inceptionism} worked best, although they were not
sufficient to completely rid the resulting visualisations of the high
frequency patterns caused by strided convolutions
\citep{odena2016deconvolution}. We performed the optimisation procedure
for activation maximisation for 20 iterations, applying the
regularisation transformations and taking gradient steps in the
\(\ell_2\)-norm \citep{madry2018towards} with a step size of 0.1.
Pseudocode for our method, applied to a trained network \(f\), is
detailed in Algorithm \ref{alg:act_max}.

\begin{algorithm}
\caption{Activation maximisation procedure with transformation robustness, frequency penalisation and $\ell_2$-norm gradient updates.}
\label{alg:act_max}
\begin{algorithmic}
\State $f' \gets \text{network } f \text{ truncated at intermediate layer}$
\State $i \gets \text{optimisation iterations}$
\State $n \gets \text{neuron/channel index}$
\State $\alpha \gets \text{step size}$
\State $x \sim U(0, 1) \text{ with dimensionality } 3 \times \text{height} \times \text{width}$
\Loop { $i$ steps}
  \State $x \gets RandomScale(x)$
  \State $x \gets RandomJitter(x)$
  \State $x \gets GaussianBlur(x)$
  \State $\mathcal{L} \gets \text{mean}(f'(x)_n)$
  \State $x \gets x + \alpha\frac{\nabla\mathcal{L}_x}{\Vert \nabla\mathcal{L}_x \Vert_2}$
  \State $x \gets \min(\max(x, 0), 1)$
\EndLoop
\State \Return $x$
\end{algorithmic}
\end{algorithm}

\hypertarget{weight-visualisations}{%
\subsubsection{Weight Visualisations}\label{weight-visualisations}}

\label{sec:weight_vis}

It is possible to visualise both convolutional filters and
fully-connected weight matrices as images. Part of the initial
excitement around DL was the observation that CNNs trained on object
recognition would learn frequency-, orientation- and colour-selective
filters \citep{krizhevsky2012imagenet}, and more broadly might reflect
the hierarchical feature extraction within the visual cortex
\citep{yamins2016using}. However, as demonstrated by Such et al.
\citeyearpar{such2018atari}, DRL agents can perform well with spatially
unstructured filters, although they did find a positive correlation
between spatial structure and performance for RL agents trained with
gradients\footnote{Intriguingly, agents trained using evolutionary
  algorithms did not develop spatially structured filters, even when
  achieving competitive performance.}. We also found this to be the
case, and hence developed quantitative measures to compare filters,
which we discuss below. Similarly, even more sophisticated
visualisations of weight matrices for fully-connected layers
\citep{hinton1991lesioning} are difficult to reason about, and so we
turned to statistical measures for these as well.

\hypertarget{statistical-and-structural-weight-characterisations}{%
\subsubsection{Statistical and Structural Weight
Characterisations}\label{statistical-and-structural-weight-characterisations}}

\label{sec:stat_struct_weight}

\hypertarget{magnitude}{%
\paragraph{Magnitude}\label{magnitude}}

\label{sec:magnitude}

A traditional measure for the ``importance'' of individual neurons in a
weight matrix is their magnitude, as exemplified by utilising weight
decay as a regulariser \citep{hanson1989comparing}. Similarly,
convolutional filters, considered as one unit, can be characterised by
their \(\ell_1\)-norms. Given that NN weights are typically randomly
initialised with small but non-zero values
\citep{lecun1998efficient, glorot2010understanding, he2015delving}, the
presence of many zeros or large values indicate significant changes
during training. We can compare these both across trained agents, and
across the training process (although change in magnitude may not
correspond with a change in task performance \citep{zhang2019all}).

\hypertarget{distribution}{%
\paragraph{Distribution}\label{distribution}}

\label{sec:distribution}

The set of weights in a layer can be considered as a distribution of
values, and analysed as such. Early connectionist work studied the
distributions of weights of trained networks, finding generally
non-normal distributions using goodness-of-fit tests and higher order
moments (skew and kurtosis)
\citep{hanson1990connectionist, bellido1993backpropagation}.

\hypertarget{spectral-analysis}{%
\paragraph{Spectral Analysis}\label{spectral-analysis}}

\label{sec:spectral}

Convolutional filters are typically initialised pseudo-randomly, so that
there exists little or no spatial correlation within a single unit. We
hence propose using the 2D discrete power spectral density (PSD) as a
way of assessing the spatial organisation of convolutional filters, and
the power spectral entropy (PSE) as a measure of their complexity. Given
the mean-centred\footnote{Offsetting the mean exposes the
  \emph{relative} spatial structure.} 2D spatial-domain filter,
\(\mathbf{W}_{m, n}\), its corresponding spectral representation,
\(\mathbf{\hat{W}}_{u, v}\), can be calculated via the 2D discrete
Fourier transform of the original filter pattern (\(j=\sqrt{-1}\)):
\begin{align*}
\mathbf{\hat{W}}_{u, v} = \sum_{m=0}^{M-1}\sum_{n=0}^{N-1}W_{m, n} \exp \left [ -\frac{j 2\pi}{MN}(u m + v n) \right ],
\end{align*} \noindent and its PSD, \(\mathbf{S}_{u, v}\), from the
normalised squared amplitude of the spectrum: \begin{align*}
\mathbf{S}_{u, v} = \frac{1}{U V} \left|\mathbf{\hat{W}}_{u, v}\right|^2,
\end{align*} \noindent where \((m, n)\) are spatial indices, \((u, v)\)
are frequency indices, \((M, N)\) is the spatial extent of the filter,
and \((U, V)\) is the frequency extent of the filter.

When renormalised such that the sum of the PSD is 1, the PSD may be
thought of as a probability mass function over a dictionary of
components from a spatial Fourier transform. We can treat each location
\((u, v)\) in Fourier space as a symbol, and its corresponding value at
\(\mathbf{S}_{u, v}\) as the probability of that symbol appearing. The
PSE is then simply the Shannon entropy of this distribution, which we
use as a measure of spatial (dis)organisation. In our analysis
(Subsection \ref{sec:statistical_and_structural}), we include statistics
calculated over randomly initialised networks as a baseline. As the
initial weights for units are typically drawn independently from a
normal or uniform distribution, this leads to a fairly flat PSD with PSE
close to \(\log(MN)\)---an upper-bound on PSE.

\begin{figure}
  \centering
  \begin{subfigure}{0.3\columnwidth}
    \includegraphics[width=\linewidth]{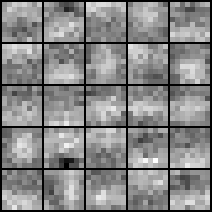}
    \subcaption{PSE  = 2.63}
  \end{subfigure}
  \begin{subfigure}{0.3\columnwidth}
    \includegraphics[width=\linewidth]{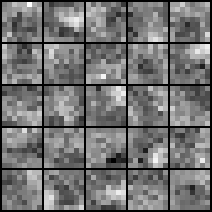}
    \subcaption{PSE = 3.19}
  \end{subfigure}
  \begin{subfigure}{0.3\columnwidth}
    \includegraphics[width=\linewidth]{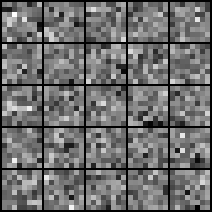}
    \subcaption{PSE = 3.86}
  \end{subfigure}
  \caption{Convolutional filters from models trained with DR, with varying power spectral entropies, ranked from the lowest (a) to highest (c). The PSE value refers to the centre filter.}
  \label{fig:entropy_vis}
\end{figure}

One weakness of spectral analysis is that these measures will fail to
pick up strongly localised spatial features, as such filters would also
result in a roughly uniform PSD. In practice, global structure is still
useful to quantify, and matches well with human intuition (Figure
\ref{fig:entropy_vis}).

Entropy as an information-theoretic measure has been used in DL in many
functions, from predicting neural network ensemble performance
\citep{hansen1990neural} to usage as a regulariser
\citep{khabou1999entropy} or pruning criteria \citep{luo2017entropy}
when applied to activations. Spectral entropy has been used as an input
feature for NNs
\citep{zheng1996digital, krkic1996eeg, misra2004spectral, srinivasan2005artificial},
but, to the best of our knowledge, not for quantifying aspects of the
network itself.

\hypertarget{unit-ablations}{%
\subsubsection{Unit Ablations}\label{unit-ablations}}

\label{sec:unit_abl}

Another way to characterise the importance of a single
neuron/convolutional filter is to remove it and observe how this affects
the performance of the NN: a large drop indicates that a particular unit
is by itself very important to the task at hand. More generally, rather
than only looking at performance, one might look for a large change in
the output. It is also possible to extend this to pairs or higher-order
groups of neurons, checking for redundancy among units
\citep{sietsma1988neural}, but this process can then become
combinatorially expensive.

This process is highly related to that of pruning---a methodology for
model compression. Pruning involves removing connections or even entire
units while minimising performance loss
\citep{sietsma1988neural, reed1993pruning}. Some statistical and
structural weight characterisations used for pruning include the
\(\ell_1\)-norm (for individual neurons \citep{han2015learning} and for
convolutional filters \citep{li2017pruning}) and discrete cosine
transform coefficients (for convolutional filters
\citep{liu2018frequency}). More broadly, one might consider redundancy
in activation space \citep{sietsma1988neural, sietsma1991creating}, or
(indirectly) change in task performance, using criteria such as the
(second) derivative of the objective function with respect to the
parameters \citep{lecun1990optimal, hassibi1993second}. As such, we
combine unit ablation studies---which give empirical results---with
these quantitative metrics.

\hypertarget{layer-re-initialisation}{%
\subsubsection{Layer Re-initialisation}\label{layer-re-initialisation}}

\label{sec:layer_abl}

One can extend the concept of ablations to entire layers, and use this
to study the \emph{re-initialisation robustness} of trained networks
\citep{zhang2019all}. Typical neural network architectures, as used in
our work, are compositions of multiple parameterised layers, with
parameters \(\{\theta_1, \theta_2, \ldots, \theta_L\}\), where \(L\) is
the depth of the network. Using \(\theta_l^t\) to denote the set of
parameters of layer \(l \in [1, L]\) at training epoch \(t \in [1, T]\)
over a maximum of \(T\) epochs, we can study the evolution of each
layer's parameters over time---for example through the change in the
\(\ell_\infty\)- or \(\ell_2\)-norm of the set of parameters.

Zhang et al. \citeyearpar{zhang2019all} proposed re-initialisation
robustness as a measure of how important a layer's parameters are with
respect to task performance over the span of the optimisation procedure.
After training, for a given layer \(l\), re-initialisation robustness is
measured by replacing the parameters \(\theta_l^T\) with parameters
checkpointed from a previous timepoint \(t\), that is, setting
\(\theta_l^T \leftarrow \theta_l^t\), and then re-measuring task
performance. They observed that for common CNN architectures trained for
object classification, while the parameters of the latter layers of the
networks tended to change a lot by the \(\ell_\infty\)- and
\(\ell_2\)-norms, the same layers were robust to re-initialisation at
checkpoints early during the optimisation procedure, and even to the
initialisation at \(t = 0\). In the latter case, the parameters are
independent of the training data, which means that the effective number
of parameters is lower than the total number of parameters. Given that
the effective number of parameters is a better measure for model
complexity than total number, this potentially allows us to
differentiate between models with the same architecture. Unlike Zhang et
al. \citeyearpar{zhang2019all}, we use re-initialisation robustness to
study the effect of task complexity (training with and without DR, and
with and without proprioceptive inputs), but with networks of similar
capacity.

\hypertarget{recurrent-ablation}{%
\subsubsection{Recurrent Ablation}\label{recurrent-ablation}}

\label{sec:recurrent_abl}

When using recurrent units in the network architecture, we can test if
non-trivial recurrent dynamics are being used by forcing the hidden
state to be constant. If the performance of the agent degrades, then it
is somehow using the recurrent dynamics to perform the task---although
it is difficult to say what the exact ``strategy'' might be. However, if
the performance drop is zero or minimal, then the recurrency is not
being utilised. The constant values of the hidden states should be set
to the empirical average of the values during normal operation, as
naively setting all values to zero could cause a considerable shift in
the distribution of expected inputs---as the hypothesis is that the
network may have learned a constant offset, rather than completely
ignoring the hidden state.

\hypertarget{entanglement}{%
\subsubsection{Entanglement}\label{entanglement}}

\label{sec:act_an}

Finally, we consider analysing the internal activations of trained
networks. One of the primary methods for examining activations is to
take the high-dimensional vectors and project them to a
lower-dimensional space (commonly \(\mathbb{R}^2\) for visualisation
purposes) using dimensionality reduction methods that try and preserve
the structure of the original data \citep{rauber2017visualizing}. Common
choices for visualising activations include both principal components
analysis (PCA; a linear projection)
\citep{pearson1901liii, elman1989representation, aubry2015understanding}
and t-distributed stochastic neighbor embedding (t-SNE; a nonlinear
projection)
\citep{maaten2008visualizing, hamel2010learning, mohamed2012understanding, donahue2014decaf, mnih2015human}.

While it is possible to qualitatively examine the projections of the
activations for a single network, or compare them across trained
networks, one can also use the projections quantitatively, by for
instance looking at class overlap in the projected space
\citep{rauber2017visualizing}. In our RL setting there is no native
concept of a ``class'', but we can instead use activations taken under
different generalisation test scenarios (Subsection
\ref{sec:test_scenarios}) to see (beyond the generalisation performance)
how the internal representations of the trained networks vary under the
different scenarios. Specifically, we measure \emph{entanglement} (``how
close pairs of representations from the same class are, relative to
pairs of representations from different classes''
\citep{frosst2019analyzing}) using the soft nearest neighbour loss,
\(\mathcal{L}_{SNN}\), \citep{salakhutdinov2007learning}, defined over a
batch of size \(B\) with samples \(\mathbf{x}\) and classes \(y\) (where
in our case \(\mathbf{x}\) is a projected activation and \(y\) is a test
scenario) with temperature \(T\) (and using \(\delta_{i,j}\) as the
Kronecker-delta):

\begin{eqnarray}
\mathcal{L}_{SNN} &=& \frac{1}{B} \sum_{n=1}^B \left ( \log \left [ \sum_{b=1}^B (1-\delta_{b,n}) \cdot e^{-\frac{\lVert \mathbf{x}_n - \mathbf{x}_b \rVert_2^2}{T}} \right ] \right . \nonumber \\
 & & \left . -\log \left [\sum_{a=1}^B (1-\delta_{a,n}) \cdot \delta_{y_a,y_n} \cdot e^{-\frac{\lVert \mathbf{x}_n - \mathbf{x}_a \rVert_2^2}{T}} \right ] \right ) \nonumber
\end{eqnarray}

In particular, if representations between different test scenarios are
highly entangled, this indicates that the network is largely invariant
to the factors of variation between the different scenarios. Considering
DR as a form of data augmentation, this is what we might expect of
networks trained with DR.

\hypertarget{experiments}{%
\section{Experiments}\label{experiments}}

\hypertarget{environments}{%
\subsection{Environments}\label{environments}}

\label{sec:environments}

\begin{figure}
  \centering
  \begin{subfigure}{0.49\columnwidth}
    \includegraphics[width=\linewidth]{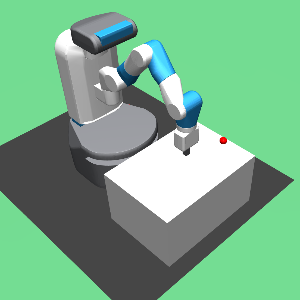}
    \subcaption{Fetch enviroment}
  \end{subfigure}
  \begin{subfigure}{0.49\columnwidth}
    \includegraphics[width=\linewidth]{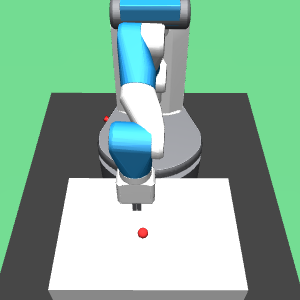}
    \subcaption{Fetch camera view}
  \end{subfigure}
  \\
  \begin{subfigure}{0.49\columnwidth}
    \includegraphics[width=\linewidth]{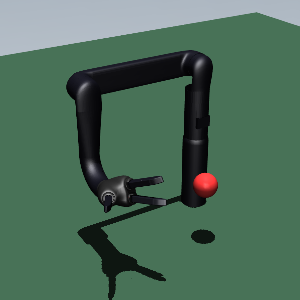}
    \subcaption{Jaco environment}
  \end{subfigure}
  \begin{subfigure}{0.49\columnwidth}
    \includegraphics[width=\linewidth]{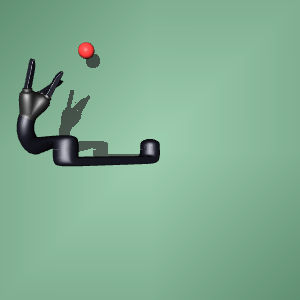}
    \subcaption{Jaco camera view}
  \end{subfigure}
  \caption{Fetch (a) and Jaco (c) environments, with associated camera views (b, d) that are provided as input to the agents.}
  \label{fig:robot_setups}
\end{figure}

In order to test the effects of DR, we base our experiments on reaching
tasks with visuomotor control. The tasks involve moving the end effector
of a robot arm to reach a randomly positioned target during each
episode, with visual (one RGB camera view) and sometimes proprioceptive
(joint positions, angles and velocities) input provided to the agent.
Unlike many DRL experiments where the position of the joints and the
target are explicitly provided, in our setup the agent must infer the
position of the target, and sometimes itself, purely through vision.
Importantly, we use two robotic arms---the Fetch Mobile Manipulator and
the KINOVA JACO Assistive robotic arm (pictured in Figure
\ref{fig:robot_setups}; henceforth referred to as Fetch and Jaco,
respectively)---which have different control schemes and different
visual appearances. This leads to changes in the relative importance of
the visual and proprioceptive inputs, which we explore in several of our
experiments.

The Fetch has a 7 degrees-of-freedom (DoF) arm, not including the
two-finger gripper. The original model and reaching task setup were
modified from the \texttt{FetchReach} task in OpenAI Gym
\citep{brockman2016openai, plappert2018multi} in order to provide an
additional camera feed for the agent (while also removing the
coordinates of the target from the input). The target can appear
anywhere on the 2D table surface. The agent has 3 sets of actions,
corresponding to position control of the end effector ({[}-5, 5{]} cm in
the x, y and z directions; gripper control is disabled).

The Jaco has been configured to be 6 DoF, with the 3 fingers disabled.
The target can appear anywhere within a 3D area to one side of the
robot's base. The agent has 6 sets of actions, corresponding to velocity
control of the arm joints ({[}-0.6, +0.6{]} rad/s). Due to the
difference in control schemes, 2D versus 3D target locations, and
homogeneous appearance of the Jaco, reaching tasks with the Jaco are
more challenging---particularly when proprioceptive input is not
provided to the agent. A summary of the different settings for the Fetch
and Jaco environments is provided in Table \ref{tbl:exp_setup}.

\begin{table}
  \caption{Summary of Fetch and Jaco experimental setups.}
  \label{tbl:exp_setup}
  \centering
  \begin{tabular}{c|cc}
    \toprule
    Setting & Fetch & Jaco \\
    \midrule
    Active (Total) DoF & 7 & 6 (9) \\
    Target Range & $21 \times 31 \text{cm}^2$ & $40 \times 40 \times 40 \text{cm}^3$ \\
    Num. Test Targets & 80 & 250 \\
    Vision Input & $3 \times 64 \times 64$ & $3 \times 64 \times 64$ \\
    Proprioceptive Inputs & 30 & 18 \\
    Control Type & Position & Velocity \\
    Num. Actions & 3 & 6 \\
    Action Discretisation & 5 & 5 \\
    Control Frequency & 6.67Hz & 6.67Hz \\
    \bottomrule
  \end{tabular}
\end{table}

During training, target positions are sampled uniformly from within the
set range, with episodes terminating once the target is reached (within
10 cm of the target centre), or otherwise timing out in 100 timesteps.
The reward is sparse, with the only nonzero reward being +1 when the
target is reached. During testing, a fixed set of target positions,
covering a uniform grid over all possible target positions, are used; 80
positions in a 2D grid are used for Fetch, and 250 positions in a 3D
grid are used for Jaco. By using a deterministic policy and averaging
performance over the entire set of test target positions, we obtain an
empirical estimate of the probability of task success. Test episodes are
set to time out within 20 timesteps in order to minimise false positives
from the policy accidentally reaching the target.

We only randomise initial positions (for all agents) and visuals (for
some agents), but not dynamics, as this is still a sufficiently rich
task setup to explore. Henceforth we refer to agents trained with visual
randomisations as being under the DR condition, whereas agents trained
without are the standard (baseline) condition. Apart from the target, we
randomise the visuals of all other objects in the environment: the
robots, the table, the floor and the skybox. At the start of every
episode and at each timestep, we randomly alter the RGB colours,
textures and colour gradients of all surfaces (Figure
\ref{fig:dr_example} for example visual observations).

Importantly, there are several aspects that are not altered, as we also
want to test extrapolation to out-of-distribution scenarios (Subsection
\ref{sec:test_scenarios}). For example, one of the tests that we apply
to probe generalisation is to change a previously static
property---surface reflectivity, which is completely disabled during
training---and see how this affects the trained agents. All environments
were constructed in MuJoCo \citep{todorov2012mujoco}, a fast and
accurate physics simulator that is commonly used for DRL experiments.

\hypertarget{networks-and-training}{%
\subsection{Networks and Training}\label{networks-and-training}}

\label{sec:networks_training}

\begin{figure*}
  \centering
  \includegraphics[width=0.85\linewidth]{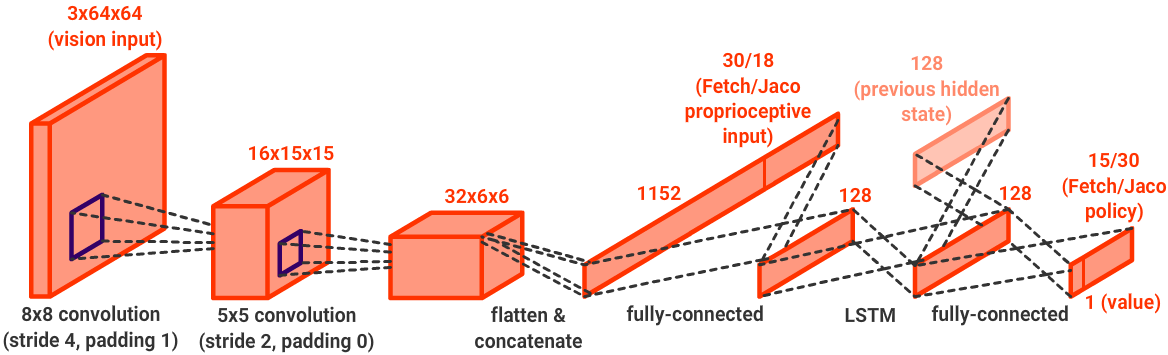}
  \caption{Actor-critic network architecture.}
  \label{fig:network}
\end{figure*}

We utilise the same basic actor-critic network architecture for each
experiment, based on the recurrent architecture used by Rusu et al.
\citeyearpar{rusu2017sim} for their Jaco experiments. The architecture
has 2 convolutional layers, a fully-connected layer, a long short-term
memory (LSTM) layer \citep{hochreiter1997long, gers2000learning}, and a
final fully-connected layer for the policy and value outputs; rectified
linear units \citep{nair2010rectified} were used at the output of the
convolutional layers and first fully-connected layer. Proprioceptive
inputs, when provided, were concatenated with the outputs of the
convolutional layers before being input into the first
fully-connected-layer. The policy, \(\pi(\cdot; \theta)\), is a product
of independent categorical distributions, with one distribution per
action dimension. Weights were initialised using orthogonal weight
initialisation \citep{saxe2014exact, ilyas2018deep} and biases were set
to zero. The specifics of the architecture are detailed in Figure
\ref{fig:network}.

During training, a stochastic policy
\(\mathbf{a} \sim \pi(\mathbf{a}|\mathbf{s}; \theta)\) is used and
trained with PPO with clip ratio \(\epsilon = 0.1\), GAE trace decay
\(\lambda = 0.95\) and discount \(\gamma = 0.99\). Each epoch of
training consists of 32 worker processes collecting 128 timesteps worth
of data each, then 4 PPO updates with a minibatch size of 1024. We train
for up to \(5 \times 10^3\) epochs, using the Adam optimiser
\citep{kingma2014adam} with learning rate \(= 2.5 \times 10^{-4}\),
\(\beta\)s \(= \{0.9, 0.999\}\), and \(\epsilon = 1 \times 10^{-5}\).
\(\mathcal{L}_{value}\) is weighted by 0.5 and \(\mathcal{L}_{entropy}\)
is weighted by 0.01. If the max \(\ell_2\)-norm of the gradients exceeds
0.5 they are rescaled to have a max \(\ell_2\)-norm of 0.5
\citep{pascanu2013difficulty}. During testing, the deterministic policy
\(\mathbf{a} = \argmax_\mathbf{a} \pi(\mathbf{a}|\mathbf{s}; \theta)\)
is used. Our training was implemented using PyTorch
\citep{paszke2017automatic}. Training each model (each seed) for the
full number of timesteps takes 1 day on a GTX 1080Ti.

\hypertarget{domain-shift}{%
\subsection{Domain Shift}\label{domain-shift}}

\label{sec:domain_shift}

\begin{figure*}
  \centering
  \hspace{0.025\textwidth}
  \begin{subfigure}{0.45\textwidth}
    \includegraphics[width=\textwidth]{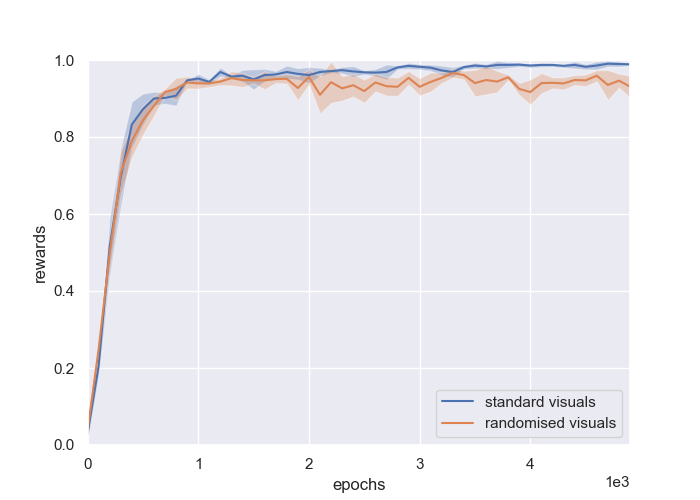}
    \caption{Proprioceptive inputs}
  \end{subfigure}
  \hfill
  \begin{subfigure}{0.45\textwidth}
    \includegraphics[width=\textwidth]{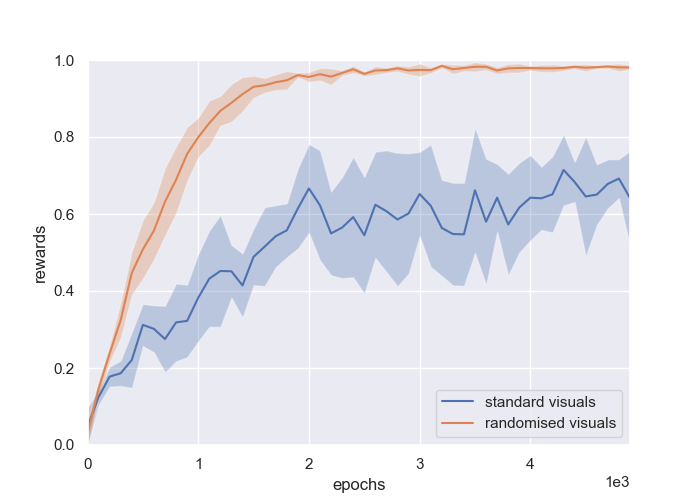}
    \caption{No proprioceptive inputs}
  \end{subfigure}
  \hspace{0.025\textwidth}
  \caption{Test performance of Jaco agents trained with DR and (a) with or (b) without proprioceptive inputs; the agents are tested against both standard and randomised visuals. Without proprioceptive inputs, the agents fail to fully deal with the domain gap between the randomised and standard visuals. Statistics (median and 95\% confidence interval) are calculated over all models (seeds) and test target locations.}
  \label{fig:domain_shift}
\end{figure*}

Once agents are successfully trained on each of the different conditions
(Fetch/Jaco, DR/no DR, proprioceptive/no proprioceptive inputs), we can
perform further tests to see how they generalise. However, while the
agents achieve practically perfect test performance on the conditions
that they were trained under, the Jaco agents trained with DR but
without proprioceptive inputs fare worse when tested under the
simulator's standard visuals (Figure \ref{fig:domain_shift}),
demonstrating a drop in performance under domain shift. It is both
assumed and observed that domain shift occurs when transferring models
trained with DR to the more complex and noisy visuals of the real world,
but it is somewhat unexpected to see this happen when shifting to
simpler visuals, which are expected to be a subset of DR visuals---this
indicates that the agent may in some sense be overfitting to the DR
visuals. Because of this, it is not completely straightforward to
compare performance between different agents, but the change in
performance of a single agent over differing test conditions is still
highly meaningful.

We also trained agents with visual DR where the visuals were only
randomised at the beginning of each episode, and kept fixed during.
These agents exhibited the same gap in performance between the standard
and randomised visuals, indicating that this is not an issue of temporal
consistency in the DR setup.

\hypertarget{test-scenarios}{%
\subsection{Test Scenarios}\label{test-scenarios}}

\label{sec:test_scenarios}

In order to test how the agents generalise to different held-out
conditions, we constructed a suite of tests for the trained agents
(Figure \ref{fig:tests} for observations for Fetch under the different
conditions\footnote{Simulation environment parameters of the Mujoco can
  be referenced from \url{http://www.mujoco.org/book/XMLreference.html}.},
and Table \ref{tbl:tests} for the results):

\begin{figure*}
  \centering
  \begin{subfigure}{0.19\textwidth}
    \includegraphics[width=\textwidth]{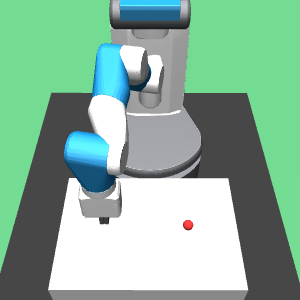}
    \caption{Standard}
  \end{subfigure}
  \begin{subfigure}{0.19\textwidth}
    \includegraphics[width=\textwidth]{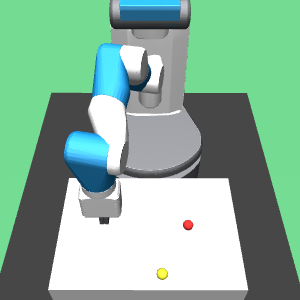}
    \caption{Colour (Y)}
  \end{subfigure}
    \begin{subfigure}{0.19\textwidth}
    \includegraphics[width=\textwidth]{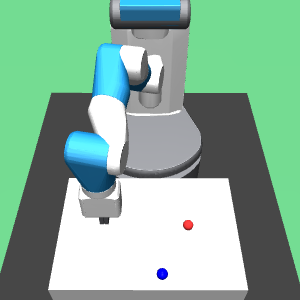}
    \caption{Colour (B)}
  \end{subfigure}
  \begin{subfigure}{0.19\textwidth}
    \includegraphics[width=\textwidth]{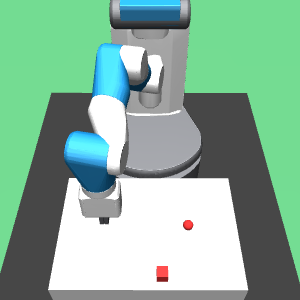}
    \caption{Shape}
  \end{subfigure}
  \begin{subfigure}{0.19\textwidth}
    \includegraphics[width=\textwidth]{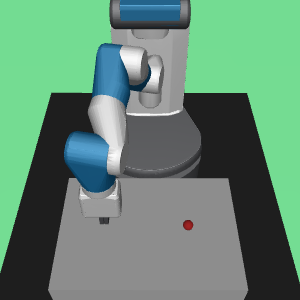}
    \caption{Illumination}
  \end{subfigure}

  \begin{subfigure}{0.19\textwidth}
    \includegraphics[width=\textwidth]{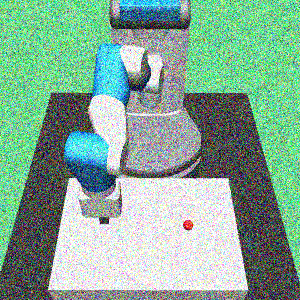}
    \caption{Noise}
  \end{subfigure}
  \begin{subfigure}{0.19\textwidth}
    \includegraphics[width=\textwidth]{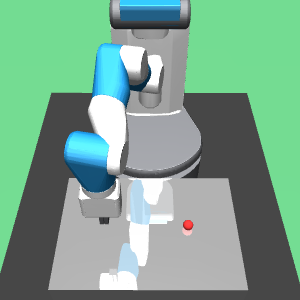}
    \caption{Reflection}
  \end{subfigure}
  \begin{subfigure}{0.19\textwidth}
    \includegraphics[width=\textwidth]{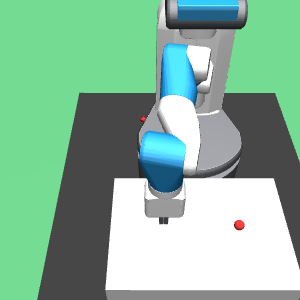}
    \caption{Translation}
  \end{subfigure}
  \begin{subfigure}{0.19\textwidth}
    \includegraphics[width=\textwidth]{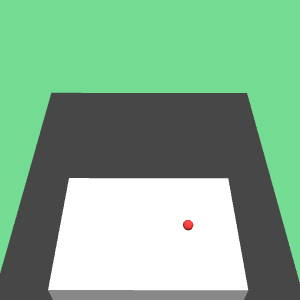}
    \caption{Invisibility}
  \end{subfigure}
  \caption{Camera observations for Fetch under different test conditions.}
  \label{fig:tests}
\end{figure*}

\begin{description}
\item[Standard.] This is the standard evaluation procedure with the default simulator visuals, where the deterministic policy is applied to all test target positions and the performance is averaged (1.0 means that all targets were reached within 20 timesteps). 
\item[Colour (Y).] This introduces a yellow sphere distractor object that is the same size and shape as the target. This specifically tests the sensitivity of the policy to localising the target given another object of a different colour in the scene, given our training regime (no distractors). Under an additive colour scheme (RGB), the yellow sphere contains both red and green components.
\item[Colour (B).] This introduces a blue sphere distractor object that is the same size and shape as the target. This tests the sensitivity of the policy to localising the target given another object of a different colour--- but in this case without any red component.
\item[Shape.] This introduces a red cube distractor object that is the same width and colour as the target, but a different shape.
\item[Illumination.] This changes the diffuse colour of the main light from 1.2 to 0.1 for Jaco, or from 0.8 to 0.0 for Fetch.
\item[Noise.] This adds Gaussian noise $\sim N(0, 0.25)$ to the visual observations. 
\item[Reflection.] This sets the table (for Fetch) or ground (for Jaco) to be reflective. This introduces reflections of the robot (and the target for Jaco) in the input.
\item[Translation.] This offsets the RGB camera by 20cm in the x direction for Jaco or 20cm in the y direction for Fetch.
\item[Invisibility.] This makes the robot transparent; this is not a realistic alteration, but is instead used to test the importance of the visual inputs for self-localisation.
\end{description}

\begin{table*}
  \caption{Test performance of all models with local visual changes (distractors), global visual changes, and invisibility (visual self-localisation test). Checkmarks and crosses indicate enabling/disabling DR and proprioceptive inputs (Prop.), respectively. Statistics are calculated over all models (seeds) and test target locations.}
  \label{tbl:tests}
  \resizebox{\linewidth}{!}{
  \begin{tabular}{c|cc|c|ccc|cccc|c}
    \toprule
    Robot & DR & Prop. & Standard & Colour (Y) & Colour (B) & Shape & Illumination & Noise & Reflection & Translation & Invisibility\\
    \midrule
    Fetch & \textcolor{red}{\xmark}   & \textcolor{red}{\xmark}   & 1.000$\pm 0.000$ & 0.993$\pm 0.007$ & 1.000$\pm 0.000$ & 0.775$\pm 0.085$ & 0.467$\pm 0.067$ & 0.980$\pm 0.006$ & 0.447$\pm 0.039$ & 0.008$\pm 0.004$ & 0.000$\pm 0.000$\\
    Fetch & \textcolor{red}{\xmark}   & \textcolor{green}{\cmark} & 1.000$\pm 0.000$ & 0.875$\pm 0.088$ & 0.995$\pm 0.004$ & 0.243$\pm 0.064$ & 0.325$\pm 0.115$ & 0.988$\pm 0.004$ & 0.570$\pm 0.078$ & 0.000$\pm 0.000$ & 0.000$\pm 0.000$\\
    Fetch & \textcolor{green}{\cmark} & \textcolor{red}{\xmark}   & 0.983$\pm 0.004$ & 0.970$\pm 0.011$ & 0.970$\pm 0.009$ & 0.913$\pm 0.042$ & 0.893$\pm 0.013$ & 0.985$\pm 0.007$ & 0.972$\pm 0.011$ & 0.093$\pm 0.040$ & 0.000$\pm 0.000$\\
    Fetch & \textcolor{green}{\cmark} & \textcolor{green}{\cmark} & 0.997$\pm 0.002$ & 0.995$\pm 0.003$ & 0.997$\pm 0.002$ & 0.963$\pm 0.020$ & 0.983$\pm 0.006$ & 0.970$\pm 0.008$ & 0.985$\pm 0.005$ & 0.153$\pm 0.055$ & 0.023$\pm 0.015$\\
    \hline
    Jaco  & \textcolor{red}{\xmark}   & \textcolor{red}{\xmark}   & 0.995$\pm 0.003$ & 0.281$\pm 0.067$ & 0.720$\pm 0.090$ & 0.274$\pm 0.077$ & 0.874$\pm 0.034$ & 0.635$\pm 0.028$ & 0.734$\pm 0.032$ & 0.394$\pm 0.055$ & 0.000$\pm 0.000$\\
    Jaco  & \textcolor{red}{\xmark}   & \textcolor{green}{\cmark} & 0.995$\pm 0.001$ & 0.451$\pm 0.040$ & 0.914$\pm 0.051$ & 0.258$\pm 0.044$ & 0.587$\pm 0.043$ & 0.478$\pm 0.059$ & 0.618$\pm 0.061$ & 0.399$\pm 0.040$ & 0.001$\pm 0.001$\\
    Jaco  & \textcolor{green}{\cmark} & \textcolor{red}{\xmark}   & 0.650$\pm 0.056$ & 0.640$\pm 0.046$ & 0.650$\pm 0.056$ & 0.636$\pm 0.040$ & 0.473$\pm 0.049$ & 0.575$\pm 0.040$ & 0.429$\pm 0.060$ & 0.141$\pm 0.034$ & 0.007$\pm 0.002$\\
    Jaco  & \textcolor{green}{\cmark} & \textcolor{green}{\cmark} & 0.991$\pm 0.004$ & 0.987$\pm 0.005$ & 0.991$\pm 0.003$ & 0.970$\pm 0.017$ & 0.442$\pm 0.018$ & 0.896$\pm 0.007$ & 0.946$\pm 0.006$ & 0.356$\pm 0.029$ & 0.916$\pm 0.022$\\
    \bottomrule
  \end{tabular}}
\end{table*}

\hypertarget{local-visual-changes}{%
\subsubsection{Local Visual Changes}\label{local-visual-changes}}

Noting that the baseline performance of the Jaco model trained with DR
but without proprioception is lower under standard visuals, across both
robots, DR confers robustness to both the colour and shape distractors
(Table \ref{tbl:tests}). However, there is not as consistent a pattern
between agents trained without DR.

With Fetch, both colour distractors have little effect on the agents,
but the shape distractor diminishes the performance of the non-DR agent
trained without proprioception somewhat, and the non-DR agent trained
with proprioception significantly. Given this, it seems that the latter
agent relies mainly on colour detection in order to locate the ball. As
a result of self-localising based on visual input alone, the former
agent develops more sophisticated vision, allowing the model to somewhat
distinguish shapes.

With Jaco, both non-DR agents suffer noticeable drops in performance in
the presence of distractors with a red component, whilst both DR agents
experience only a very small decrease in performance across all local
distractors. While the non-DR agents also have reduced success with the
blue sphere distractor, it is less pronounced, indicating that non-DR
Jaco agents are primarily detecting large red components as the target
object.

In order to test that the location of the distractor does not also
influence the models' responses, we varied this and recorded the
corresponding success rates. The low standard deviations shown in Table
\ref{tbl:vary_distractor} indicate that the location only has a minimal
impact on the results.

\begin{table}
  \caption{Test performance of a single model with distractors locations varying over 9 different on the ground plane (Jaco) and table (Fetch). Checkmarks and crosses indicate enabling/disabling DR and proprioceptive inputs (Prop.), respectively. Statistics are calculated for the best model (seed), over all test target locations and all distractor locations.}
  \label{tbl:vary_distractor}
  \centering
  \begin{tabular}{c|cc|cc}
    \toprule
    Robot & DR & Prop. & Colour (Y) & Shape\\
    \midrule
    Fetch & \textcolor{red}{\xmark}   & \textcolor{red}{\xmark}   & 0.79$\pm 0.06$ & 0.44$\pm 0.08$\\
    Fetch & \textcolor{red}{\xmark}   & \textcolor{green}{\cmark} & 0.75$\pm 0.08$ & 0.35$\pm 0.06$\\
    Fetch & \textcolor{green}{\cmark} & \textcolor{red}{\xmark}   & 0.84$\pm 0.06$ & 0.50$\pm 0.11$\\
    Fetch & \textcolor{green}{\cmark} & \textcolor{green}{\cmark} & 0.86$\pm 0.05$ & 0.49$\pm 0.10$\\
    \hline
    Jaco  & \textcolor{red}{\xmark}   & \textcolor{red}{\xmark}   & 0.31$\pm 0.05$ & 0.18$\pm 0.04$\\
    Jaco  & \textcolor{red}{\xmark}   & \textcolor{green}{\cmark} & 0.45$\pm 0.07$ & 0.20$\pm 0.04$\\
    Jaco  & \textcolor{green}{\cmark} & \textcolor{red}{\xmark}   & 0.69$\pm 0.02$ & 0.41$\pm 0.05$\\
    Jaco  & \textcolor{green}{\cmark} & \textcolor{green}{\cmark} & 0.91$\pm 0.02$ & 0.46$\pm 0.09$\\
    \bottomrule
  \end{tabular}
\end{table}

\hypertarget{global-visual-changes}{%
\subsubsection{Global Visual Changes}\label{global-visual-changes}}

Referring to Table \ref{tbl:tests}, DR generally confers more
robustness, although this time the DR agents do exhibit noticeable drops
in performance across many of these tests.

Reducing the illumination does drop the performance of all agents,
although the Fetch agents trained with DR are the most robust.
Intriguingly, the Jaco agents trained without proprioception are more
robust with respect to this change, as compared to the agents trained
with. Their need to self-localise visually necessitates a more complex
visual system, whereas simpler visual processing may be thrown off by
the reduction in contrast or even simply the change in the pixel values
of the target. Given that the DR agents trained with proprioception tend
to be the most robust across most of the test conditions, this motivates
an additional consideration for training---when performing sensor fusion
within a model, the combination of information should be more resilient
to the loss or faulty functioning of any individual sensory input.

Additive Gaussian noise has very little effect on the Fetch agents, but
reduces the performance of the Jaco agents---by over 30\% for agents
trained without DR, but only by about 10\% for agents trained with DR.

Making the table surface reflective throws off the Fetch agents trained
without DR, with an approximately 50\% drop in performance, but with DR
the agents are resilient to this change. The Jaco agents trained without
DR also incur a significant, yet smaller drop in performance. A likely
explanation for this difference is that the size of the robots relative
to the image differs, and the reflection of the Jaco arm simply changes
the input less. When given proprioceptive inputs, both the Fetch and the
Jaco agent trained with DR display similar levels of resilience.

Translating the camera causes a dramatic drop in performance in all
agents. DR confers a minimal amount of resilience to this for the Fetch
agents, with the best performance at 15\%. The performance of most Jaco
agents drops approximately 60\%, apart from the agent trained with DR
but without proprioception, for which the drop is 50\%. In the absence
of DR, the Fetch agents fail completely, whilst the Jaco agents achieve
a success rate of 39\%, suggesting that all Jaco agents manage to learn
a degree of translation invariance for their policies. One hypothesis
for this is that the requirement to reach a target in 3D confers a more
generalisable representation of space.

\hypertarget{visual-self-localisation}{%
\subsubsection{Visual
Self-localisation}\label{visual-self-localisation}}

For nearly all agents, rendering the robot invisible drops performance
to zero. There are four non-zero performance scores, but three of these
are low enough to be attributable to chance. This test indicates that
perhaps either directly or indirectly the position of the robot is
inferred visually, although we cannot rule out that the drop in
performance is due to the domain shift that results from rendering the
arm invisible. The standout is the Jaco agent with proprioceptive inputs
and DR training, which only incurs a small drop in performance---this
agent is able to self-localise solely based on proprioceptive input.

\hypertarget{tests-summary}{%
\subsubsection{Tests Summary}\label{tests-summary}}

\begin{figure}
  \centering
  \begin{subfigure}{0.45\textwidth}
    \includegraphics[width=\linewidth]{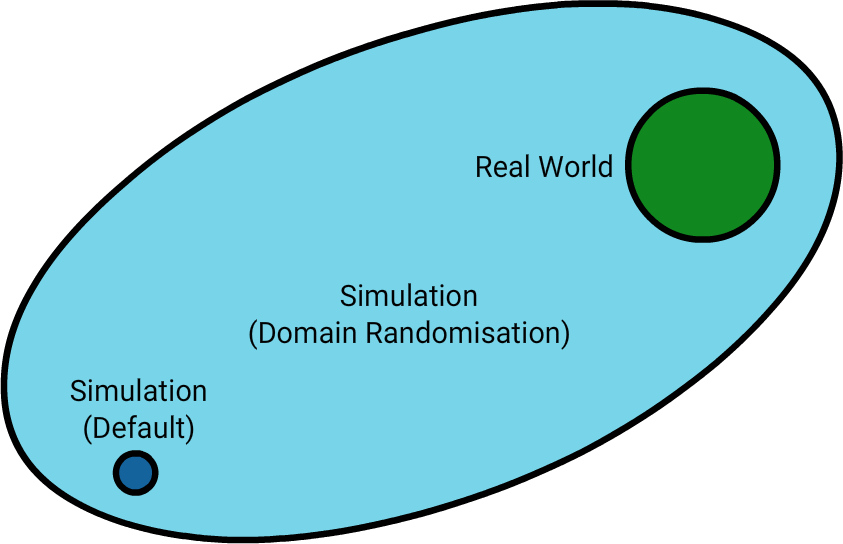}
    \subcaption{Idealised theory}
  \end{subfigure}
  \hspace{0.05\textwidth}
  \begin{subfigure}{0.45\textwidth}
    \includegraphics[width=\linewidth]{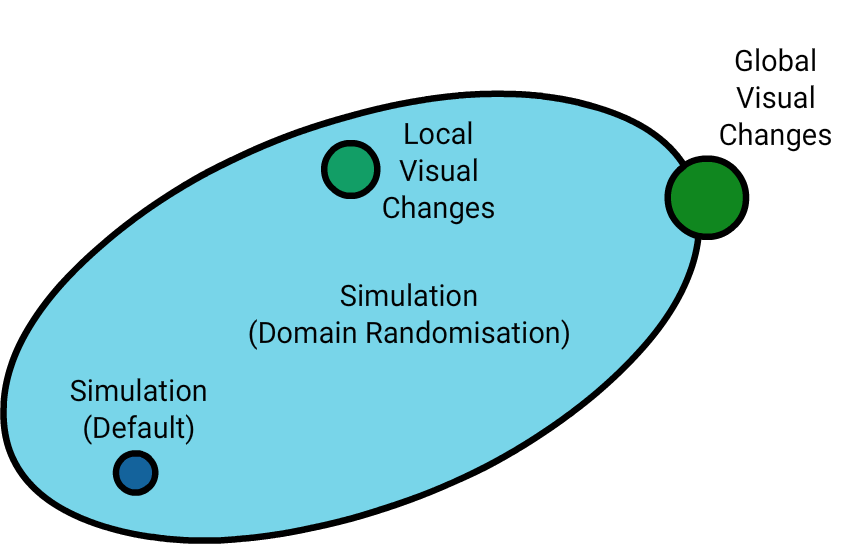}
    \subcaption{Our experiments for testing out-of-domain generalisation}
  \end{subfigure}
  \caption[Domain randomisation methodology.]{Domain randomisation methodology. (a) In theory, the range of simulation parameters should be varied in such a way as to successfully encompass visual/physical properties that would be encountered in the real world. (b) In practice, we limited the scope of variation to test generalisation to out-of-domain inputs within simulation, and observed different levels of success depending on the type of change. The local and global visual changes are designed to test generalisation in a way that is reflective of the real world.}
  \label{fig:dr_concept}
\end{figure}

There is no single clear result from our evaluation of different setups
with different types of tests, beyond the general importance of sensor
fusion and DR to improve the ability for agents to generalise. The type
of DR used during training---randomising colours and textures---allows
generalisation to localised changes---distractor objects---but fails to
reliably improve generalisation across the more global changes, such as
illumination or translation (Figure \ref{fig:dr_concept}). This should
not come as a surprise given that our DR never changed the position of
the robot, nor the illumination of the target. The takeaway is that
``generalisation'' is more nuanced, and performing systematic tests can
help probe what strategies networks might be using to operate. Finding
failure cases for ``weaker'' agents can still be a useful exercise for
evaluating more robust agents, as it enables adversarial evaluation
\citep{uesato2018rigorous}, and can inform us about the design of DR.

\hypertarget{model-analysis}{%
\section{Model Analysis}\label{model-analysis}}

The unit tests that we constructed can be used to evaluate the
performance of an arbitrary black box policy under differing conditions,
but we also have the ability to inspect the internals of our trained
agents. Although we cannot obtain a complete explanation for the learned
policies, we can still glean further information from both the learned
parameters and the sets of activations in the networks.

\hypertarget{saliency-maps-1}{%
\subsection{Saliency Maps}\label{saliency-maps-1}}

\label{sec:saliency_maps}

One of the first tests usually conducted is to examine saliency maps to
infer which aspects of the input influence the output of the agent. We
use the occlusion-based technique with average baseline, and focus on
distractors: we show saliency maps for both the standard test setup, and
with either the different colour (Y) or different shape distractors.

\begin{figure*}
  \centering
  \begin{subfigure}{0.32\columnwidth}
    \includegraphics[width=\linewidth]{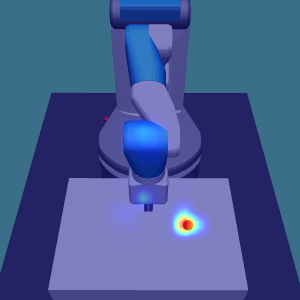}
    \subcaption{Standard (no DR, no proprioception)}
  \end{subfigure}
  \begin{subfigure}{0.32\columnwidth}
    \includegraphics[width=\linewidth]{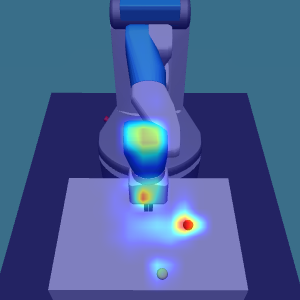}
    \subcaption{Colour (no DR, no proprioception)}
  \end{subfigure}
  \begin{subfigure}{0.32\columnwidth}
    \includegraphics[width=\linewidth]{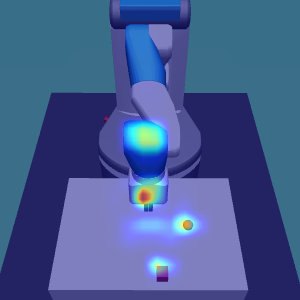}
    \subcaption{Shape (no DR, no proprioception)}
  \end{subfigure}
  \begin{subfigure}{0.32\columnwidth}
    \includegraphics[width=\linewidth]{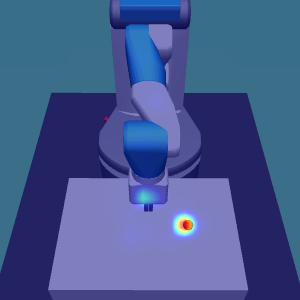}
    \subcaption{Standard (no DR, proprioception)}
  \end{subfigure}
  \begin{subfigure}{0.32\columnwidth}
    \includegraphics[width=\linewidth]{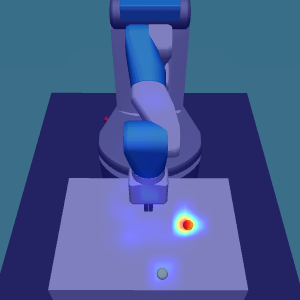}
    \subcaption{Colour (no DR, proprioception)}
  \end{subfigure}
  \begin{subfigure}{0.32\columnwidth}
    \includegraphics[width=\linewidth]{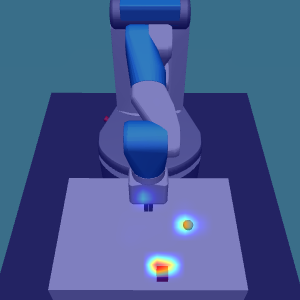}
    \subcaption{Shape (no DR, proprioception)}
  \end{subfigure}

  \begin{subfigure}{0.32\columnwidth}
    \includegraphics[width=\linewidth]{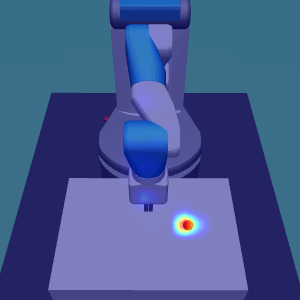}
    \subcaption{Standard (DR, no proprioception)}
  \end{subfigure}
  \begin{subfigure}{0.32\columnwidth}
    \includegraphics[width=\linewidth]{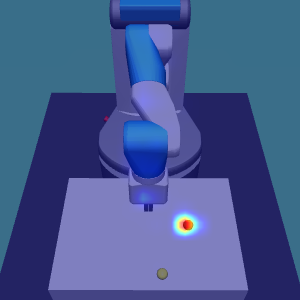}
    \subcaption{Colour (DR, no proprioception)}
  \end{subfigure}
  \begin{subfigure}{0.32\columnwidth}
    \includegraphics[width=\linewidth]{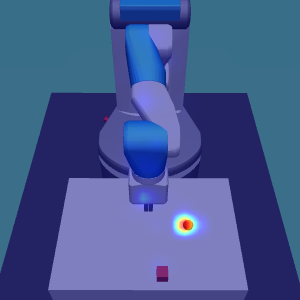}
    \subcaption{Shape (DR, no proprioception)}
  \end{subfigure}
  \begin{subfigure}{0.32\columnwidth}
    \includegraphics[width=\linewidth]{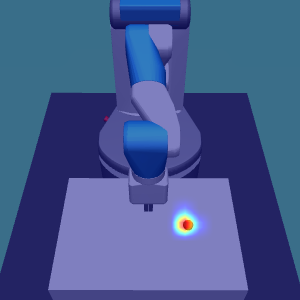}
    \subcaption{Standard (DR, proprioception)}
  \end{subfigure}
  \begin{subfigure}{0.32\columnwidth}
    \includegraphics[width=\linewidth]{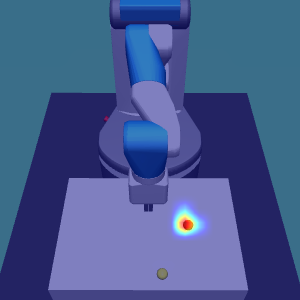}
    \subcaption{Colour (DR, proprioception)}
  \end{subfigure}
  \begin{subfigure}{0.32\columnwidth}
    \includegraphics[width=\linewidth]{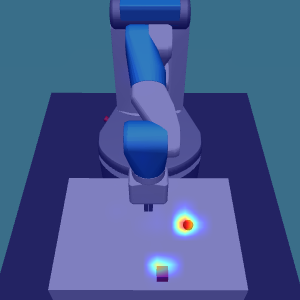}
    \subcaption{Shape (DR, proprioception)}
  \end{subfigure}
  \caption{Occlusion-based saliency maps with Fetch models trained with (g-l) or without (a-f) DR and with (d-f, j-l) or without proprioception (a-c, g-i) in three different distractor conditions. The best Fetch model was used for each training condition.}
  \label{fig:saliency_fetch_distractor}
\end{figure*}

\begin{figure*}
  \centering
  \begin{subfigure}{0.32\columnwidth}
    \includegraphics[width=\linewidth]{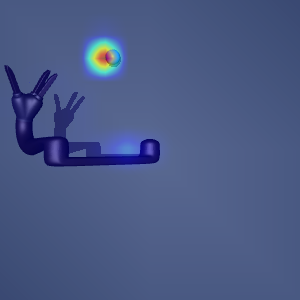}
    \subcaption{Standard (no DR, no proprioception)}
  \end{subfigure}
  \begin{subfigure}{0.32\columnwidth}
    \includegraphics[width=\linewidth]{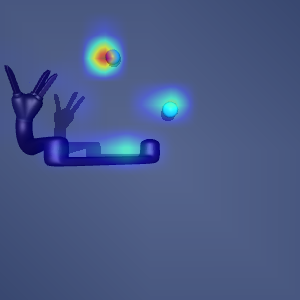}
    \subcaption{Colour (no DR, no proprioception)}
  \end{subfigure}
  \begin{subfigure}{0.32\columnwidth}
    \includegraphics[width=\linewidth]{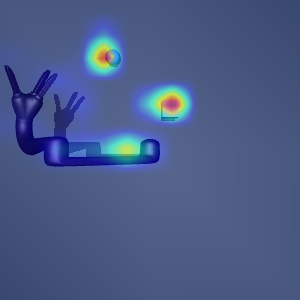}
    \subcaption{Shape (no DR, no proprioception)}
  \end{subfigure}
  \begin{subfigure}{0.32\columnwidth}
    \includegraphics[width=\linewidth]{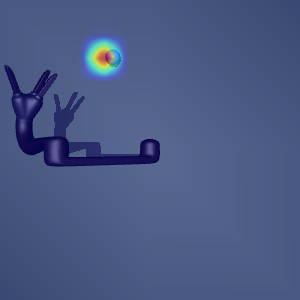}
    \subcaption{Standard (no DR, proprioception)}
  \end{subfigure}
  \begin{subfigure}{0.32\columnwidth}
    \includegraphics[width=\linewidth]{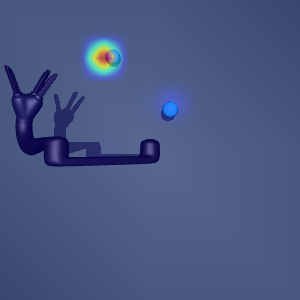}
    \subcaption{Colour (no DR, proprioception)}
  \end{subfigure}
  \begin{subfigure}{0.32\columnwidth}
    \includegraphics[width=\linewidth]{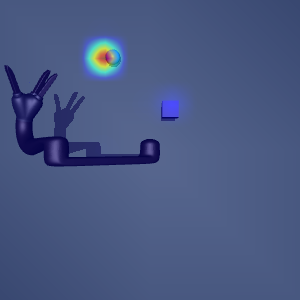}
    \subcaption{Shape (no DR, proprioception)}
  \end{subfigure}

  \begin{subfigure}{0.32\columnwidth}
    \includegraphics[width=\linewidth]{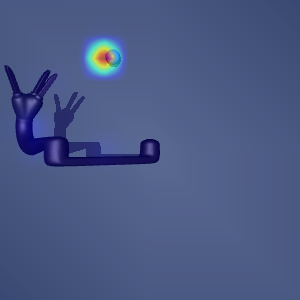}
    \subcaption{Standard (DR, no proprioception)}
  \end{subfigure}
  \begin{subfigure}{0.32\columnwidth}
    \includegraphics[width=\linewidth]{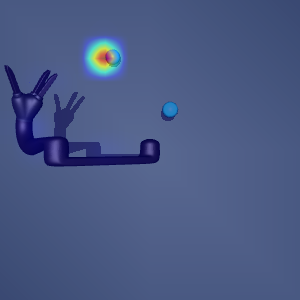}
    \subcaption{Colour (DR, no proprioception)}
  \end{subfigure}
  \begin{subfigure}{0.32\columnwidth}
    \includegraphics[width=\linewidth]{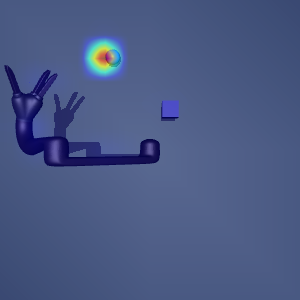}
    \subcaption{Shape (DR, no proprioception)}
  \end{subfigure}
  \begin{subfigure}{0.32\columnwidth}
    \includegraphics[width=\linewidth]{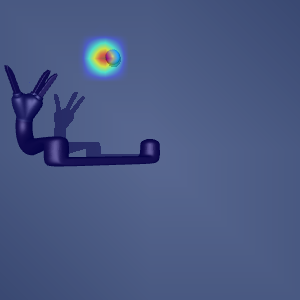}
    \subcaption{Standard (DR, proprioception)}
  \end{subfigure}
  \begin{subfigure}{0.32\columnwidth}
    \includegraphics[width=\linewidth]{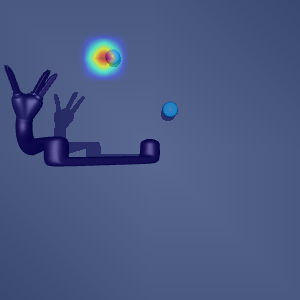}
    \subcaption{Colour (DR, proprioception)}
  \end{subfigure}
  \begin{subfigure}{0.32\columnwidth}
    \includegraphics[width=\linewidth]{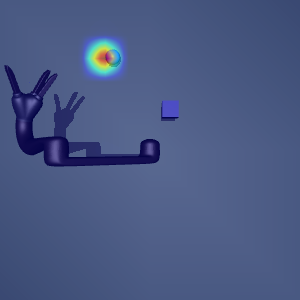}
    \subcaption{Shape (DR, proprioception)}
  \end{subfigure}
  \caption{Occlusion-based saliency maps with Jaco models trained with (g-l) or without (a-f) DR and with (d-f, j-l) or without proprioception (a-c, g-i) in three different distractor conditions. The best Jaco model was used for each training condition.}
  \label{fig:saliency_jaco_distractor}
\end{figure*}

The saliency maps for the Fetch agents (Figure
\ref{fig:saliency_fetch_distractor}) differs between all models. Apart
from the model trained with DR and with proprioception (Figure
\ref{fig:saliency_fetch_distractor}j-l), all agents seem to use the
gripper to self-localise. Despite having access to clean proprioceptive
inputs, the Fetch agent trained without DR still pays attention to its
own body in the image---so it is not necessarily the case that agents
will even utilise the inputs that we may expect. The Fetch agents
trained without DR show saliency on the distractors (Figure
\ref{fig:saliency_fetch_distractor}a-f), while the agents trained with
DR do not (with the exception of the model trained with DR and
proprioception on the shape distractor, as seen in Figure
\ref{fig:saliency_fetch_distractor}).

The saliency maps for the Jaco agents (Figure
\ref{fig:saliency_jaco_distractor}) are more homogeneous, with a large
amount of attention on the target, and little elsewhere. The saliency
for the agent trained without DR and without proprioception clearly
shows some attention around the base of the arm (Figure
\ref{fig:saliency_jaco_distractor}a-c) that would indicate visual
self-localisation. On an initial inspection, it may appear that there is
no saliency around the arm for the agent trained with DR and without
proprioception, although we know that in order to succeed it must be
relying on visual self-localisation. Indeed, there is saliency present
around the arm (Figure \ref{fig:saliency_jaco_distractor}g-i), but it is
difficult to perceive. This example indicates the subjective nature of
interpreting saliency maps, and hence why they should not be the sole
tool for analysis.

This recommendation is also borne out by the mismatch between the
saliency maps and performance. For the Fetch agents trained with DR, the
agent with proprioception shows saliency over the shape distractor
(Figure \ref{fig:saliency_fetch_distractor}l) in contrast to without
proprioception (Figure \ref{fig:saliency_fetch_distractor}i);
conversely, the performance drop is greater in the latter than the
former. Similarly for the Jaco agents trained without DR, the agent
without proprioception shows a large amount of saliency over the shape
distractor (Figure \ref{fig:saliency_jaco_distractor}c), while the agent
with proprioception demonstrates only a minimal amount of saliency
(Figure \ref{fig:saliency_jaco_distractor}f); however, they both have
the same drop in performance (\textgreater{} 70\%).

\hypertarget{activation-maximisation-1}{%
\subsection{Activation Maximisation}\label{activation-maximisation-1}}

\label{sec:activation_maximisation}

\begin{figure*}
  \centering
  \begin{subfigure}{0.49\textwidth}
    \includegraphics[width=\textwidth]{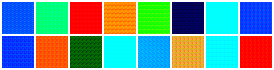}
    \caption{Convolution 1 (no DR, no proprioception)}
  \end{subfigure}
  \begin{subfigure}{0.49\textwidth}
    \includegraphics[width=\textwidth]{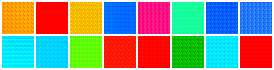}
    \caption{Convolution 1 (no DR, proprioception)}
  \end{subfigure}

  \begin{subfigure}{0.49\textwidth}
    \includegraphics[width=\textwidth]{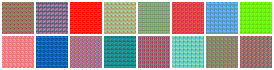}
    \caption{Convolution 1 (DR, no proprioception)}
  \end{subfigure}
  \begin{subfigure}{0.49\textwidth}
    \includegraphics[width=\textwidth]{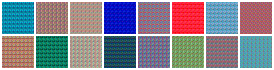}
    \caption{Convolution 1 (DR, proprioception)}
  \end{subfigure}

  \begin{subfigure}{0.49\textwidth}
    \includegraphics[width=\textwidth]{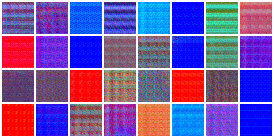}
    \caption{Convolution 2 (no DR, no proprioception)}
  \end{subfigure}
  \begin{subfigure}{0.49\textwidth}
    \includegraphics[width=\textwidth]{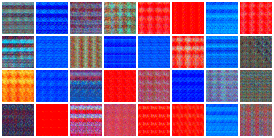}
    \caption{Convolution 2 (no DR, proprioception)}
  \end{subfigure}

  \begin{subfigure}{0.49\textwidth}
    \includegraphics[width=\textwidth]{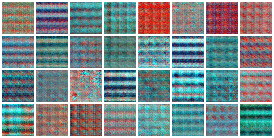}
    \caption{Convolution 2 (DR, no proprioception)}
  \end{subfigure}
  \begin{subfigure}{0.49\textwidth}
    \includegraphics[width=\textwidth]{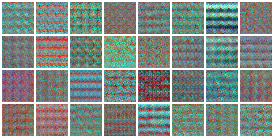}
    \caption{Convolution 2 (DR, proprioception)}
  \end{subfigure}
  
  \begin{subfigure}{0.49\textwidth}
    \centering
    \includegraphics[width=0.63\textwidth]{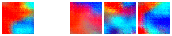}
    \caption{Value + Policy (no DR, no proprioception)}
  \end{subfigure}
  \begin{subfigure}{0.49\textwidth}
    \centering
    \includegraphics[width=0.63\textwidth]{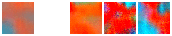}
    \caption{Value + Policy (no DR, proprioception)}
  \end{subfigure}

  \begin{subfigure}{0.49\textwidth}
    \centering
    \includegraphics[width=0.63\textwidth]{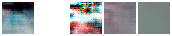}
    \caption{Value + Policy (DR, no proprioception)}
  \end{subfigure}
  \begin{subfigure}{0.49\textwidth}
    \centering
    \includegraphics[width=0.63\textwidth]{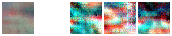}
    \caption{Value + Policy (DR, proprioception)}
  \end{subfigure}
  \caption{Activation maximisation for trained Fetch agents: first convolutional layer (a-d); second convolutional layer (e-h); value and policy outputs (i-l). The best Fetch model was used for each training condition. Proprioceptive inputs and hidden state for value and policy visualisations are set to zero. Agents trained without DR have many red filters (the colour of the target) in the second layer (e, f), while agents trained with DR have more structured oriented red-blue filters (g, h). In comparison, the Jaco task induces more structured filters even without DR (see Figure \ref{fig:act_max_jaco}).}
  \label{fig:act_max_fetch}
\end{figure*}

\begin{figure*}
  \centering
  \begin{subfigure}{0.49\textwidth}
    \includegraphics[width=\textwidth]{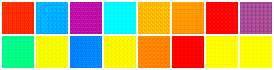}
    \caption{Convolution 1 (no DR, no proprioception)}
  \end{subfigure}
  \begin{subfigure}{0.49\textwidth}
    \includegraphics[width=\textwidth]{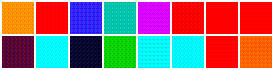}
    \caption{Convolution 1 (no DR, proprioception)}
  \end{subfigure}

  \begin{subfigure}{0.49\textwidth}
    \includegraphics[width=\textwidth]{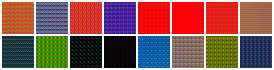}
    \caption{Convolution 1 (DR, no proprioception)}
  \end{subfigure}
  \begin{subfigure}{0.49\textwidth}
    \includegraphics[width=\textwidth]{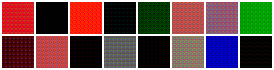}
    \caption{Convolution 1 (DR, proprioception)}
  \end{subfigure}

  \begin{subfigure}{0.49\textwidth}
    \includegraphics[width=\textwidth]{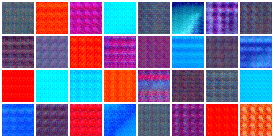}
    \caption{Convolution 2 (no DR, no proprioception)}
  \end{subfigure}
  \begin{subfigure}{0.49\textwidth}
    \includegraphics[width=\textwidth]{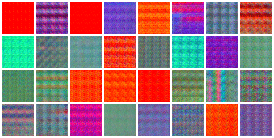}
    \caption{Convolution 2 (no DR, proprioception)}
  \end{subfigure}

  \begin{subfigure}{0.49\textwidth}
    \includegraphics[width=\textwidth]{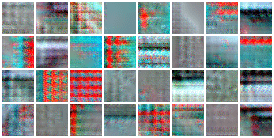}
    \caption{Convolution 2 (DR, no proprioception)}
  \end{subfigure}
  \begin{subfigure}{0.49\textwidth}
    \includegraphics[width=\textwidth]{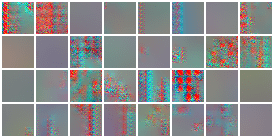}
    \caption{Convolution 2 (DR, proprioception)}
  \end{subfigure}

  \begin{subfigure}{0.49\textwidth}
    \includegraphics[width=\textwidth]{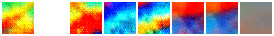}
    \caption{Value + Policy (no DR, no proprioception)}
  \end{subfigure}
  \begin{subfigure}{0.49\textwidth}
    \includegraphics[width=\textwidth]{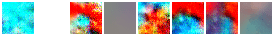}
    \caption{Value + Policy (no DR, proprioception)}
  \end{subfigure}

  \begin{subfigure}{0.49\textwidth}
    \includegraphics[width=\textwidth]{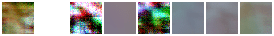}
    \caption{Value + Policy (DR, no proprioception)}
  \end{subfigure}
  \begin{subfigure}{0.49\textwidth}
    \includegraphics[width=\textwidth]{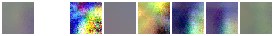}
    \caption{Value + Policy (DR, proprioception)}
  \end{subfigure}
  \caption{Activation maximisation for trained Jaco agents: first convolutional layer (a-d); second convolutional layer (e-h); value and policy outputs (i-l). The best Jaco model was used for each training condition. Proprioceptive inputs and hidden state for value and policy visualisations are set to zero. All agents have colour-gradient filters in the second layer (e-h), indicating more visual complexity than needed for the Fetch task (Figure \ref{fig:act_max_fetch}).}
  \label{fig:act_max_jaco}
\end{figure*}

In line with Such et al. \citeyearpar{such2018atari}, activation
maximisation applied to the first convolutional layer results in edge
detectors, with larger-scale spatial structure in the latter layers
(Figure \ref{fig:act_max_fetch} and Figure \ref{fig:act_max_jaco}).
There are several trends that apply to both the Fetch and Jaco agents.
Firstly, the agents trained without DR develop simpler, more colourful
filters in both layers. In contrast, the agents trained with DR develop
more edge-like detectors, with higher contrast, in their first
convolutional layers. In their second convolutional layers, the feature
detectors resemble the red target itself, surrounded by a complementary
blue-green. This style of detector is consistent across both the Fetch
and Jaco agents, which suggests that it was not developed in response to
the green floor in the Jaco environment.

Across the layer 1 images, an outlier is the Jaco agent trained with DR
and proprioception---it appears to have several ``dead'' filters in the
first layer (Figure \ref{fig:act_max_jaco}d), and filters in the second
layer which do not respond to any particular pattern (Figure
\ref{fig:act_max_jaco}h). However, even though the pattern from the
second filter in the first layer is almost black, it results in a drop
in performance from 81\% to 72\% when ablated (Subsection
\label{sec:unit_ablations}), showing that it is not possible to properly
judge the importance of a filter using activation maximisation.

Finally, there is a more global, but largely uninterpretable structure
when maximising the value function or policy outputs (choosing the unit
that corresponds to the largest positive movement per action output).
For Fetch agents without DR, the visualisations are dominated by red
(the target colour), but with DR there is a wider spectrum of colours.
This trend is the same for the Jaco agents, although without DR and
without proprioceptive inputs the colours that maximise the value output
are purple and green (a constant hue shift on the usual red and blue).
The agents trained with DR but without proprioception have the most
plain activation maximisation images for the policy, perhaps suggesting
a more factorised control scheme. For the Fetch agent, only the first
and third actuators are activated by strong visual inputs (given zeroes
as the proprioceptive inputs and hidden state), which correspond to the
most important joints for accomplishing this reaching task (the rotating
base and the elbow).

As a reminder we note that activation maximisation may not (and is
practically unlikely to) converge to images within the training data
manifold \citep{mahendran2015understanding}---a disadvantage addressed
by the complementary technique of finding image patches within the
training data that maximally activate individual neurons
\citep{girshick2014rich}.

\hypertarget{statistical-and-structural-weight-characterisations-1}{%
\subsection{Statistical and Structural Weight
Characterisations}\label{statistical-and-structural-weight-characterisations-1}}

\label{sec:statistical_and_structural}

\begin{figure*}
  \centering
  \begin{subfigure}{0.24\textwidth}
    \includegraphics[width=\textwidth]{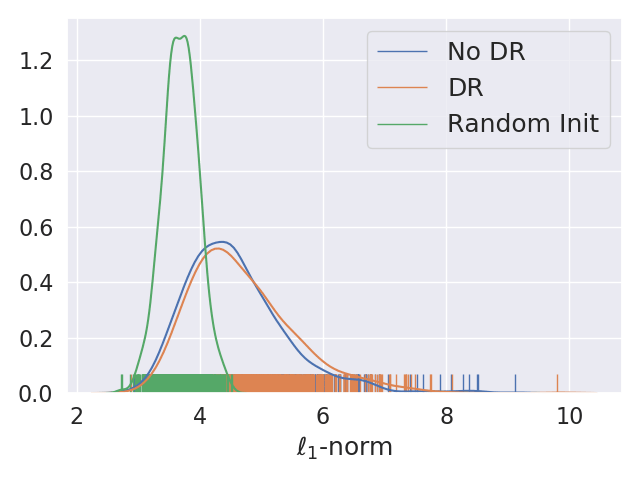}
    \caption{$\ell_1$-norm (layer 1)}
  \end{subfigure}
  \begin{subfigure}{0.24\textwidth}
    \includegraphics[width=\textwidth]{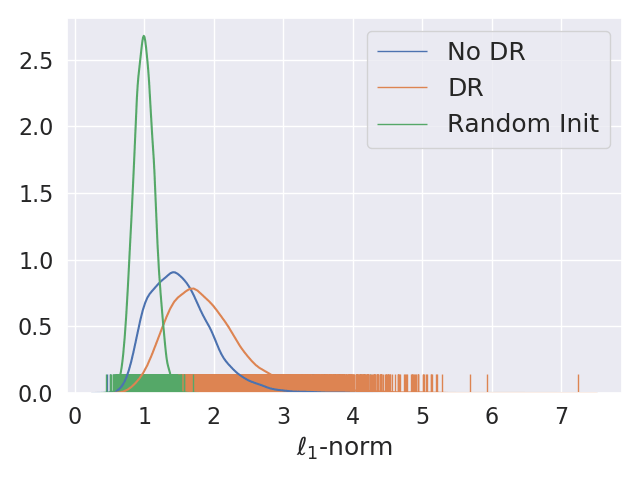}
    \caption{$\ell_1$-norm (layer 2)}
  \end{subfigure}
  \begin{subfigure}{0.24\textwidth}
    \includegraphics[width=\textwidth]{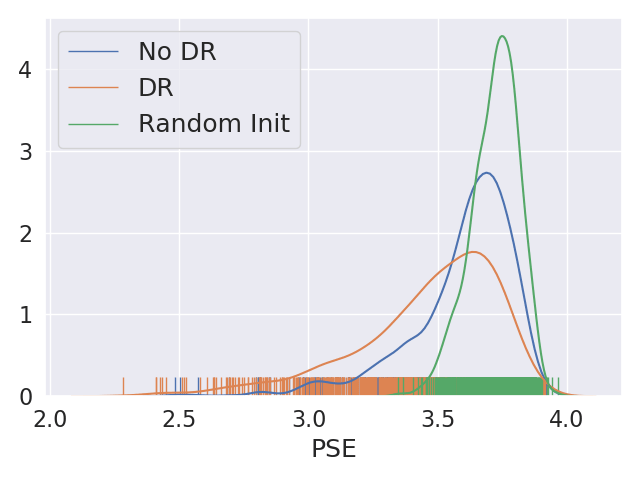}
    \caption{PSE (layer 1)}
  \end{subfigure}
  \begin{subfigure}{0.24\textwidth}
    \includegraphics[width=\textwidth]{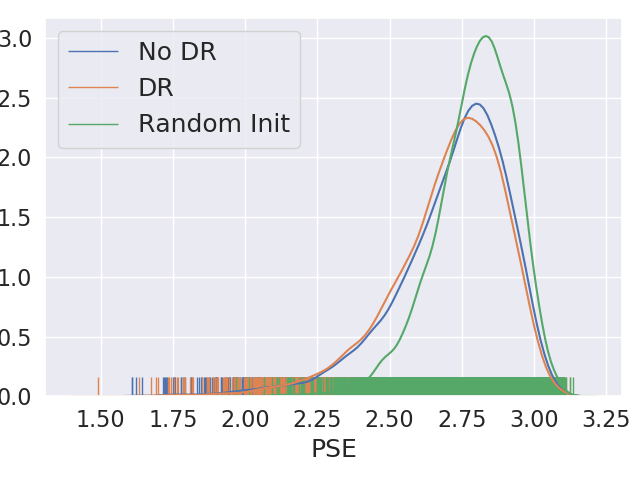}
    \caption{PSE (layer 2)}
  \end{subfigure}
  \caption{Effect of DR on statistical and structural characterisiations of convolutional filters, using all filters from all models, along with models with randomly initialised weights. This effect is layer-dependent, with a large change in $\ell_1$-norm for layer 2, but not layer 1, and a relatively larger change in PSE for layer 1 as compared to layer 2.}
  \label{fig:norms_entropy}
\end{figure*}

We calculated statistical and structural weight characteristics over all
trained models (Fetch and Jaco, with/without proprioception,
with/without DR, 5 seeds), which allows us to average over 40 conditions
to examine the effects of DR. We analysed the norms (Subsection
\ref{sec:magnitude}) and moments (Subsection \ref{sec:distribution}) of
all of the weights of the trained agents, and could not find consistent
trends across all layers. The most meaningful characterisations were the
\(\ell_1\)-norm and the power spectral entropy, PSE, (Subsection
\ref{sec:spectral}), applied to the convolutional filters.

Figure \ref{fig:norms_entropy} shows a KDE of the \(\ell_1\)-norms and
PSEs of all of the 2D filters within the first and second convolutional
layers. For the \(\ell_1\)-norm, in layer 2 the distribution is skewed
towards higher values when the model is trained with DR. For the PSE, in
both layers, but particularly layer 1, the distribution is skewed
towards lower values when the model is trained with DR. Using the
nonparametric Kolmogorov-Smirnov (K-S) two-sided test between the two
distributions (DR versus non-DR), the \(p\)-value of the
\(\ell_1\)-norms is 0.014 (K-S statistic 0.072) for layer 1 and
\(\sim 0\) (K-S statistic 0.285) for layer 2, and the \(p\)-value of the
PSEs is \(5.71\times10^{-27}\) (K-S statistic 0.251) for layer 1 and
\(3.32\times10^{-9}\) (K-S statistic 0.044) for layer 2. Given the same
weight initialisation distributions across all models, this difference
indicates that DR causes a significant change in the final distribution
of weights, with both larger weights and greater spatial structure.

\hypertarget{unit-ablations-1}{%
\subsection{Unit Ablations}\label{unit-ablations-1}}

\label{sec:unit_ablations}

\begin{figure*}
  \centering
  \begin{subfigure}{0.49\linewidth}
    \includegraphics[width=\textwidth]{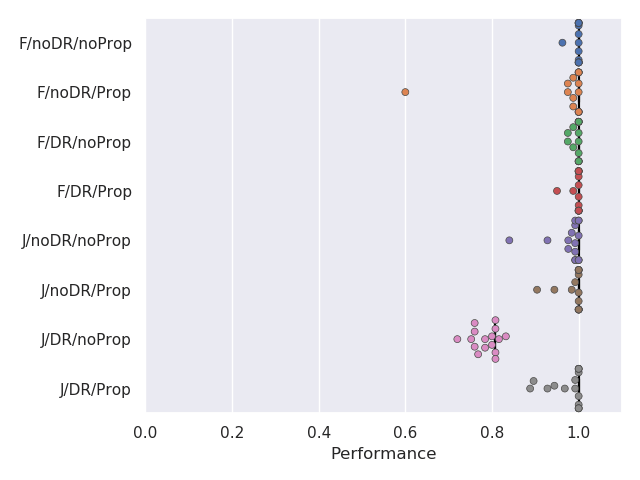}
    \caption{Standard environment (layer 1)}
  \end{subfigure}
  \begin{subfigure}{0.49\linewidth}
    \includegraphics[width=\textwidth]{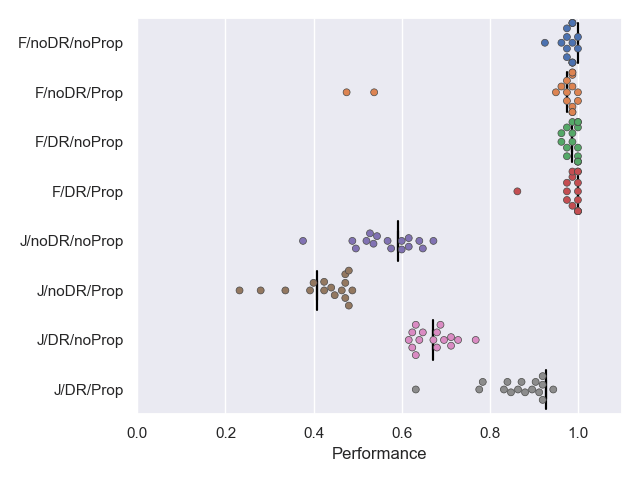}
    \caption{Noisy environment (layer 1)}
  \end{subfigure}

  \begin{subfigure}{0.49\linewidth}
    \includegraphics[width=\textwidth]{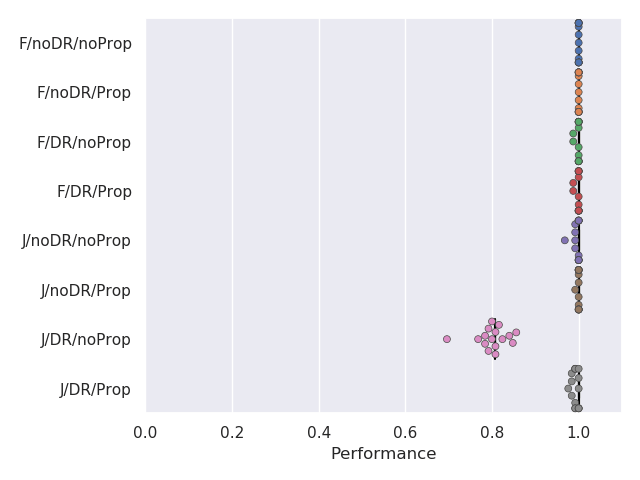}
    \caption{Standard environment (layer 2)}
  \end{subfigure}
  \begin{subfigure}{0.49\linewidth}
    \includegraphics[width=\textwidth]{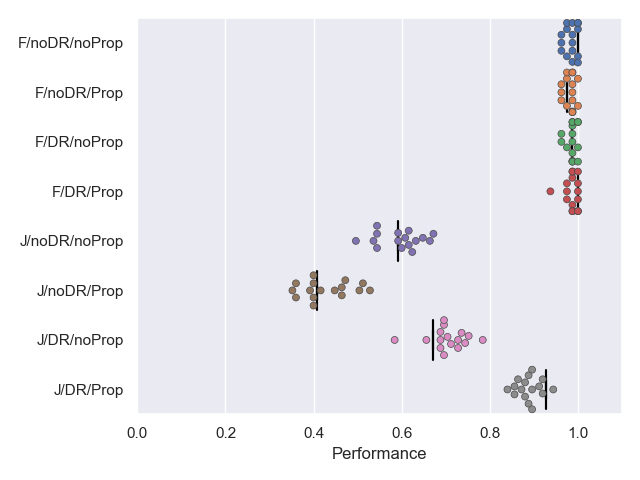}
    \caption{Noisy environment (layer 2)}
  \end{subfigure}
  \caption{Unit-wise ablation tests in two different visual test environments. Each point corresponds to one unit in layer 1 (a, b) or layer 2 (c, d), with the vertical bars representing baseline performance in the test environment. The training settings correspond to the Fetch (F) and Jaco (J) robots, whether additional proprioceptive inputs are available (Prop), and if DR was used. The best model was used for each training condition. Note that the Jaco agent trained with DR but without proprioception already has a lower base performance on the standard visuals than the other models (see Table \ref{tbl:tests}).}
  \label{fig:ablations}
\end{figure*}

Given access to the trained models, unit ablations allow us to perform a
quantitative, white box analysis. To ablate units, we manually zero the
activations of one of the output channels in either the first or second
convolutional layers, iterating the process over every channel. We then
re-evaluate the modified agents for each of the 8 training settings,
using the agent with the best performance over all 5 seeds for each one
(noting that the performance of the best Jaco agent trained with DR and
without proprioception is significantly higher than the average, as
reported in Table \ref{tbl:tests}). These agents are tested on a single
\(x-y\) plane of the fixed test targets---the full 80 for Fetch, and 125
for Jaco---and both the standard visual and additive Gaussian noise test
scenarios (see Subsection \ref{sec:test_scenarios}), as the latter is
often used to mimic sensor noise in robotic learning tasks
\citep{jakobi1995noise}. The results of the ablations are presented in
Figure \ref{fig:ablations}.

We can make several observations from the plots in Figure
\ref{fig:ablations}. Firstly, the Fetch agents are barely affected by
unit ablations, whereas they have varying effects on the Jaco agents.
The higher variability for Jaco agents could be due to the increased
complexity of the Jaco task (both in terms of extracting relevant
information from the sensory inputs, and the difficulty of the
actuation).

Secondly, there is a greater spread of values in layer 1 ablations
(Figure \ref{fig:ablations}a,b) versus layer 2 (Figure
\ref{fig:ablations}c,d). In particular, there appear to be a few highly
important units in layer 1, resulting in highly skewed distributions. We
believe this supports what we observe in the activation maximisation
plots (Figure \ref{fig:act_max_fetch} and Figure
\ref{fig:act_max_jaco}), where there is a greater diversity in the layer
1 filters.

While we can observe a greater variability in the noisy environment
(Figure \ref{fig:ablations}b,d), variability seems to be most correlated
with low performance. Intriguingly, when performance is suboptimal,
ablations can even improve performance beyond the baseline results. We
note that performance is only suboptimal when the agent has not been
trained under the corresponding condition---either the Fetch agent
trained with DR and without proprioception is tested on the standard
environment (Figure \ref{fig:ablations}a,c), or when any of the agents
are tested in the noisy environment (Figure \ref{fig:ablations}b,d).
This suggests a degree of overfitting to the training conditions.

One of our original hypotheses was that DR might force the learned
representations to become more redundant---as quantified by reduced
variability under unit ablation---but the results do not support this.
Instead, the baseline performance of the agents trained with DR is
simply higher than that of the agents trained without DR in the noisy
environment.

\hypertarget{layer-re-initialisation-1}{%
\subsection{Layer Re-initialisation}\label{layer-re-initialisation-1}}

\label{sec:layer_ablations}

\begin{figure*}
  \centering
  \begin{subfigure}{0.32\textwidth}
    \includegraphics[width=\textwidth]{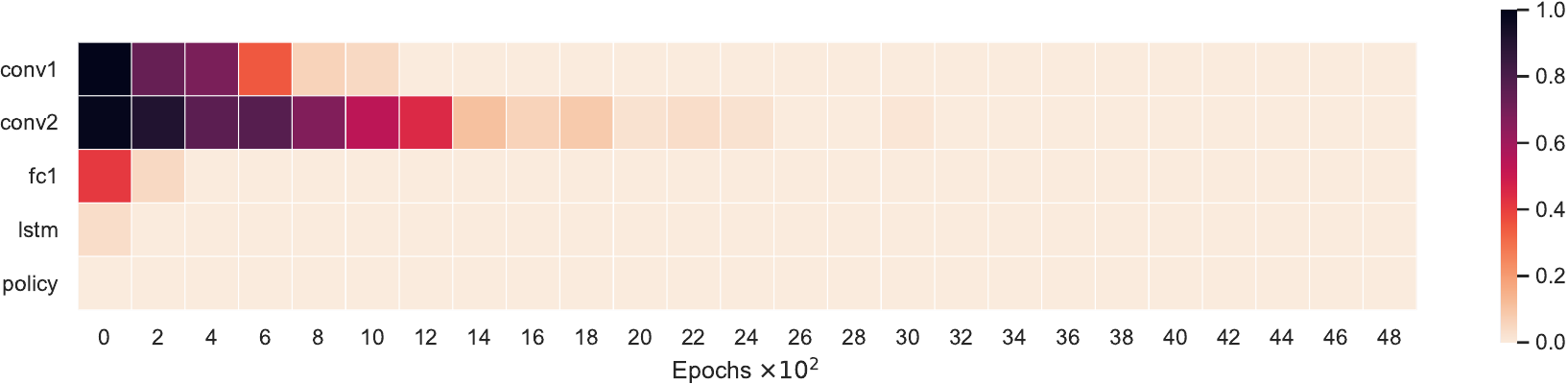}
    \caption{Robustness (no DR, no proprioception)}
  \end{subfigure}
  \begin{subfigure}{0.32\textwidth}
    \includegraphics[width=\textwidth]{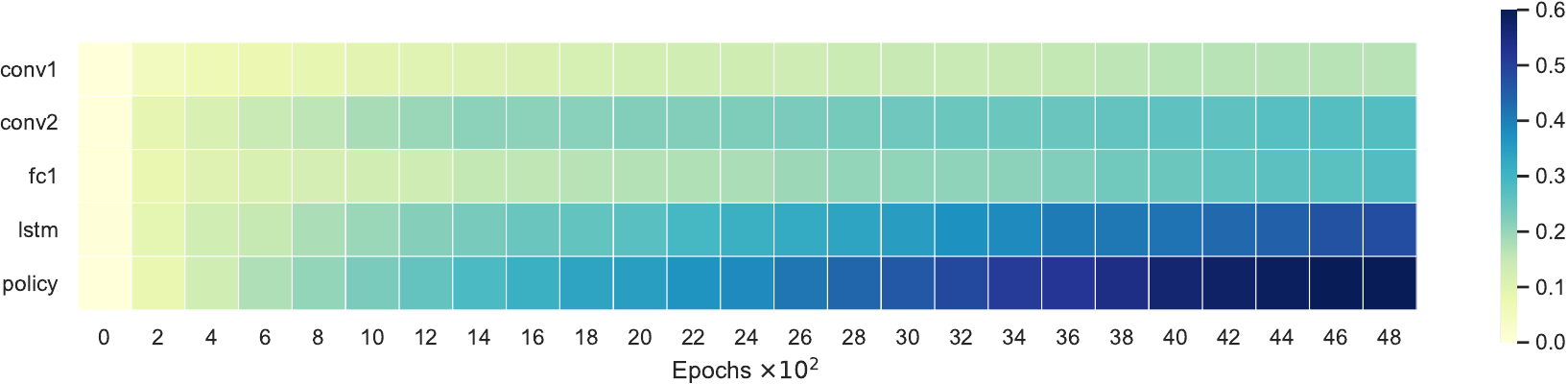}
    \caption{$\ell_\infty$-norm (no DR, no proprioception)}
  \end{subfigure}
  \begin{subfigure}{0.32\textwidth}
    \includegraphics[width=\textwidth]{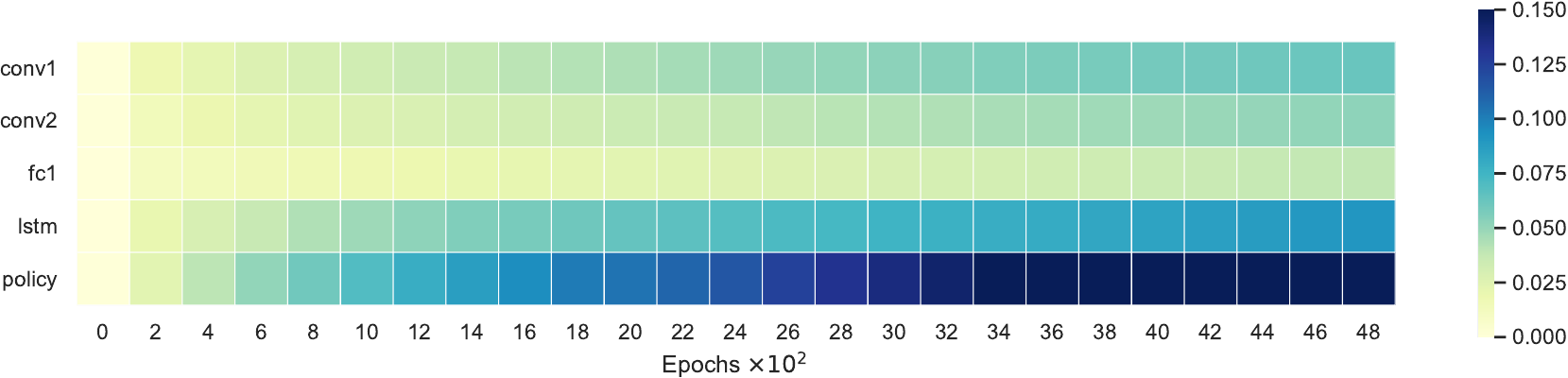}
    \caption{$\ell_2$-norm (no DR, no proprioception)}
  \end{subfigure}
  \begin{subfigure}{0.32\textwidth}
    \includegraphics[width=\textwidth]{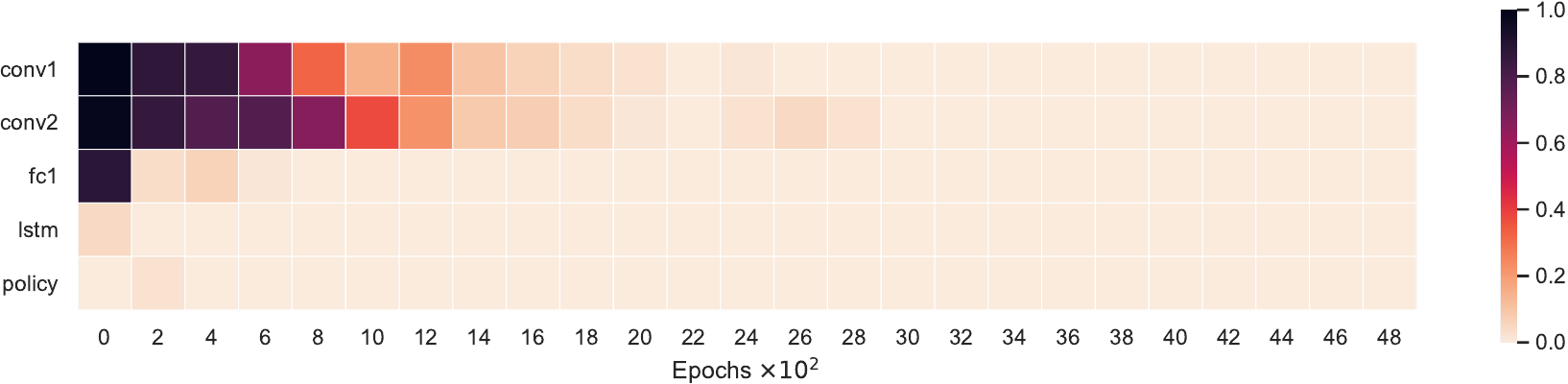}
    \caption{Robustness (no DR, proprioception)}
  \end{subfigure}
  \begin{subfigure}{0.32\textwidth}
    \includegraphics[width=\textwidth]{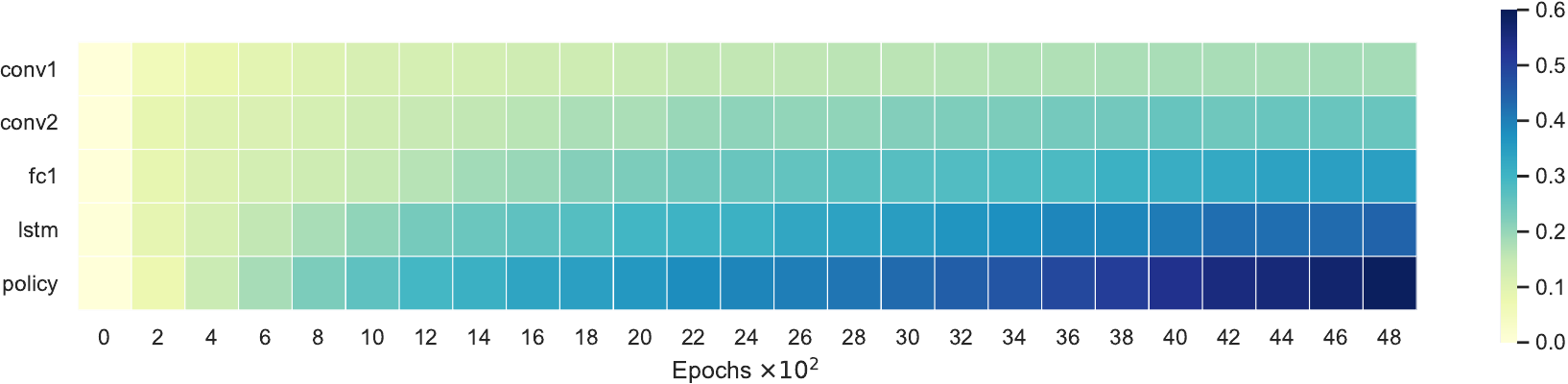}
    \caption{$\ell_\infty$-norm (no DR, proprioception)}
  \end{subfigure}
  \begin{subfigure}{0.32\textwidth}
    \includegraphics[width=\textwidth]{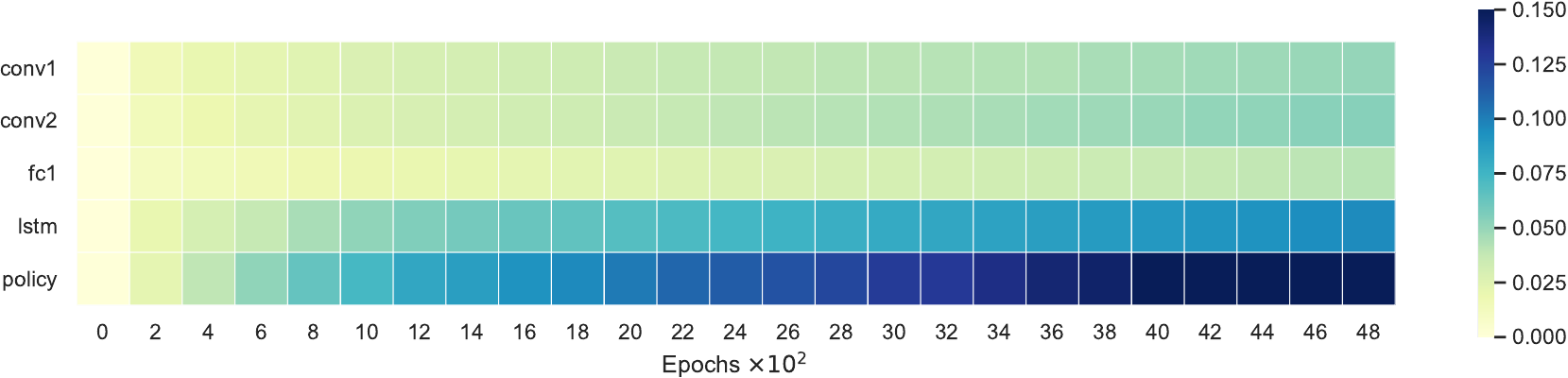}
    \caption{$\ell_2$-norm (no DR, proprioception)}
  \end{subfigure}
  \begin{subfigure}{0.32\textwidth}
    \includegraphics[width=\textwidth]{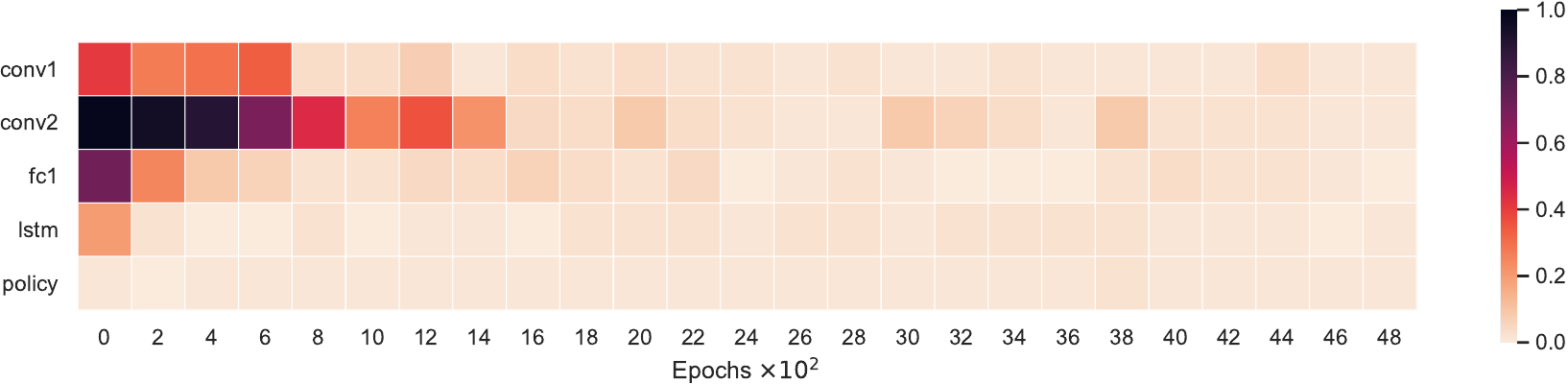}
    \caption{Robustness (DR, no proprioception)}
  \end{subfigure}
  \begin{subfigure}{0.32\textwidth}
    \includegraphics[width=\textwidth]{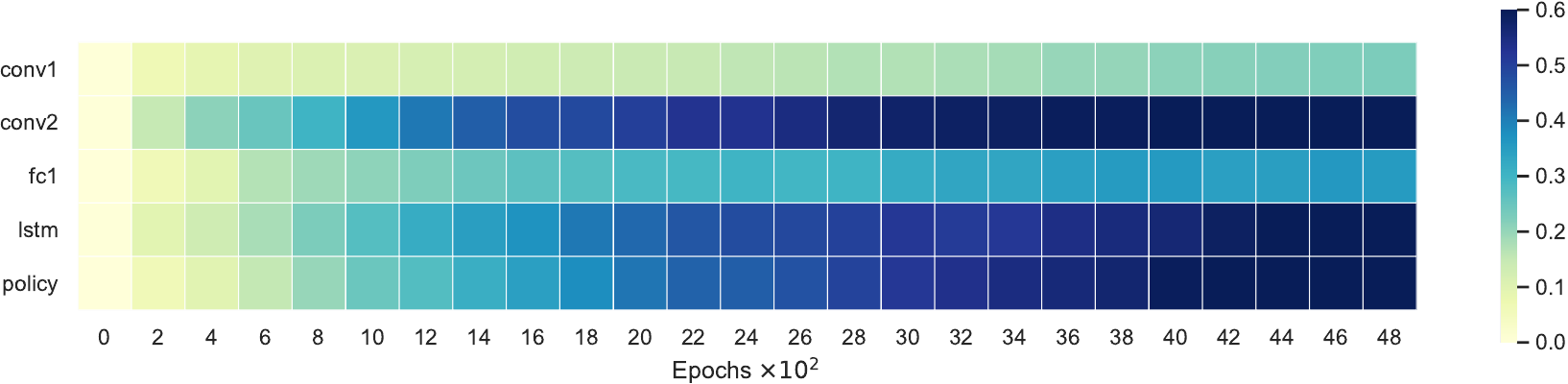}
    \caption{$\ell_\infty$-norm (DR, no proprioception)}
  \end{subfigure}
  \begin{subfigure}{0.32\textwidth}
    \includegraphics[width=\textwidth]{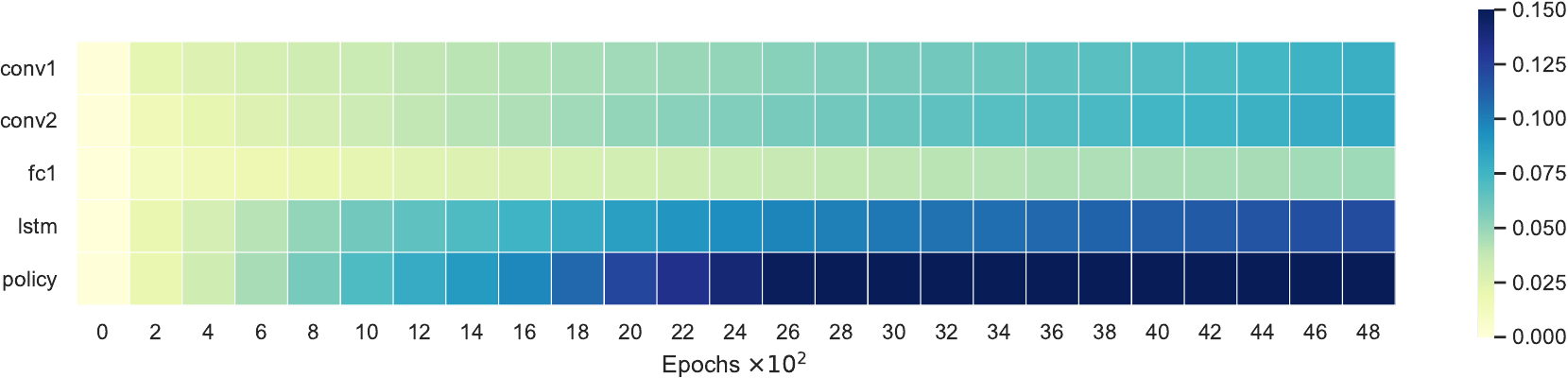}
    \caption{$\ell_2$-norm (DR, no proprioception)}
  \end{subfigure}
  \begin{subfigure}{0.32\textwidth}
    \includegraphics[width=\textwidth]{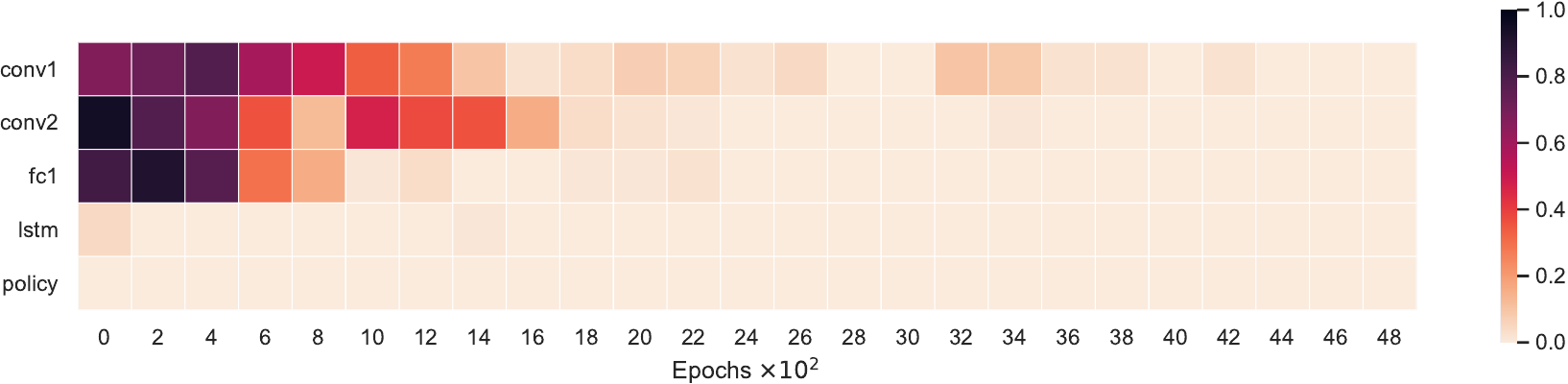}
    \caption{Robustness (DR, proprioception)}
  \end{subfigure}
  \begin{subfigure}{0.32\textwidth}
    \includegraphics[width=\textwidth]{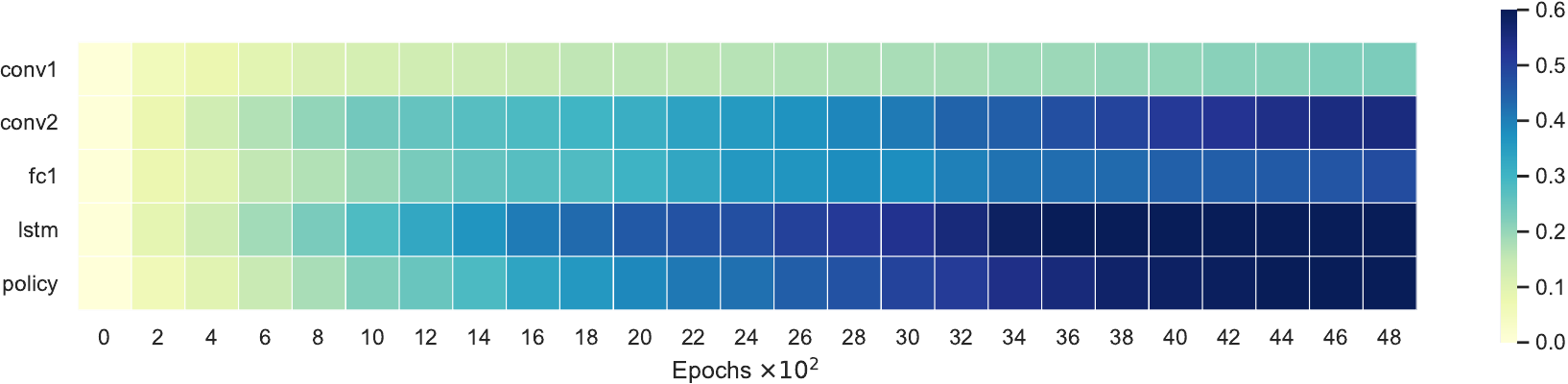}
    \caption{$\ell_\infty$-norm (DR, proprioception)}
  \end{subfigure}
  \begin{subfigure}{0.32\textwidth}
    \includegraphics[width=\textwidth]{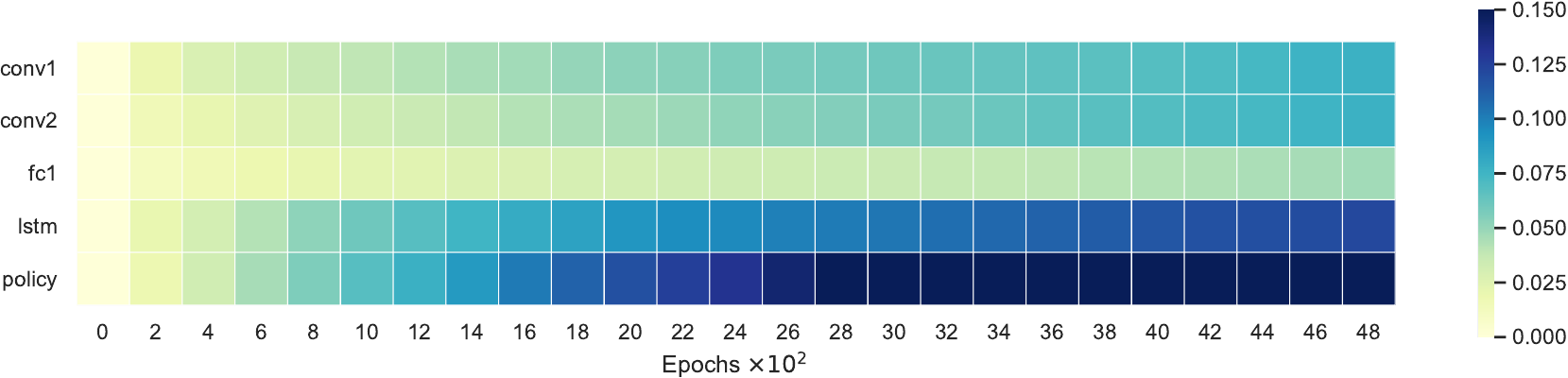}
    \caption{$\ell_2$-norm (DR, proprioception)}
  \end{subfigure}
  \caption{Re-initialisation robustness (1 for complete failure, and 0 for complete success), and change in $\ell_\infty$- and $\ell_2$-norm of parameters of Fetch agents trained with (g-l) and without (a-f) DR, and with (d-f, j-l) and without (a-c, g-i) proprioceptive inputs. Plots truncated to show detail during initial epochs. The best Fetch model was chosen for each training condition.}
  \label{fig:fetch_robustness}
\end{figure*}

\begin{figure*}
  \centering
  \begin{subfigure}{0.32\textwidth}
    \includegraphics[width=\textwidth]{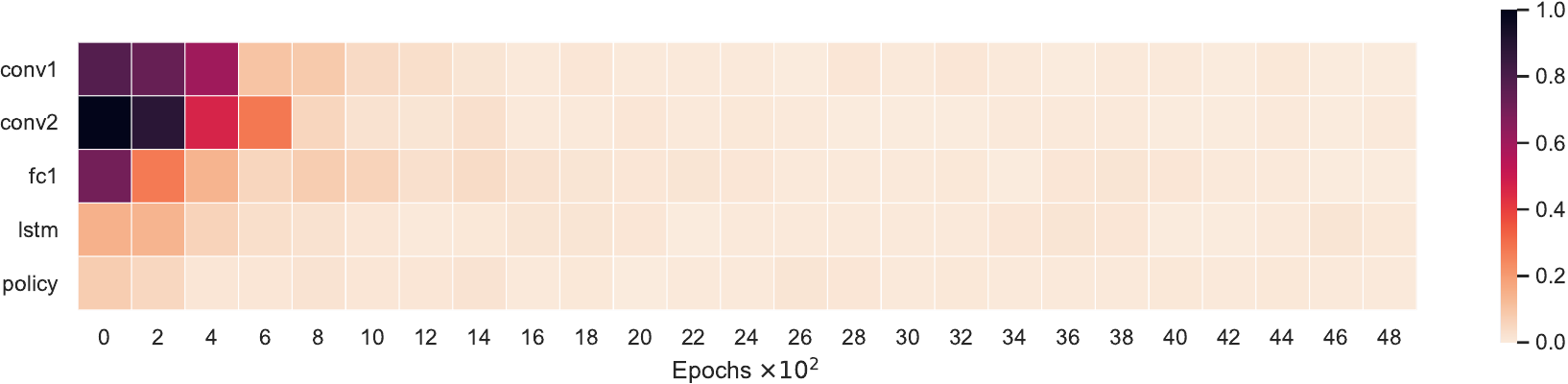}
    \caption{Robustness (no DR, no proprioception)}
  \end{subfigure}
  \begin{subfigure}{0.32\textwidth}
    \includegraphics[width=\textwidth]{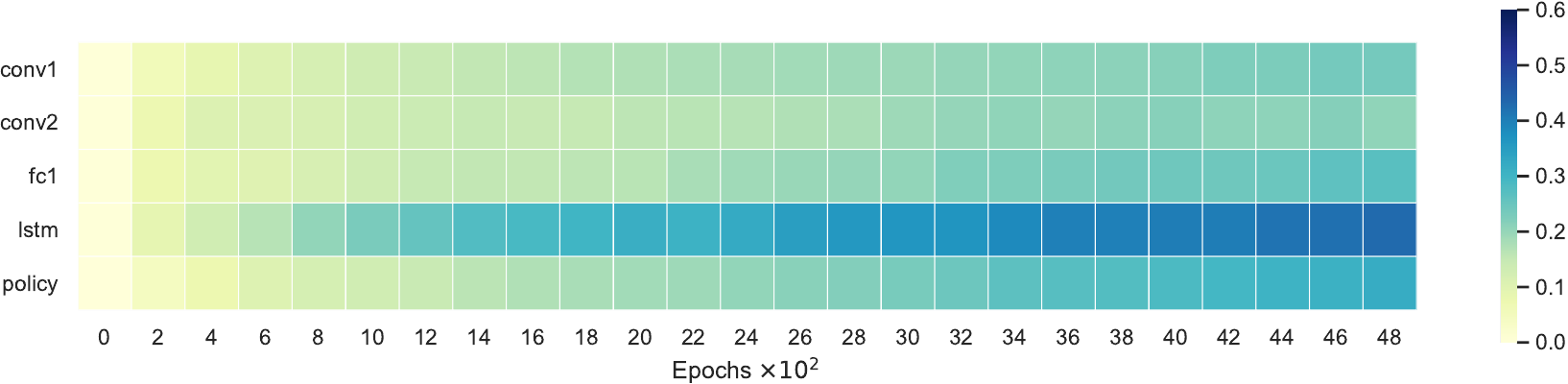}
    \caption{$\ell_\infty$-norm (no DR, no proprioception)}
  \end{subfigure}
  \begin{subfigure}{0.32\textwidth}
    \includegraphics[width=\textwidth]{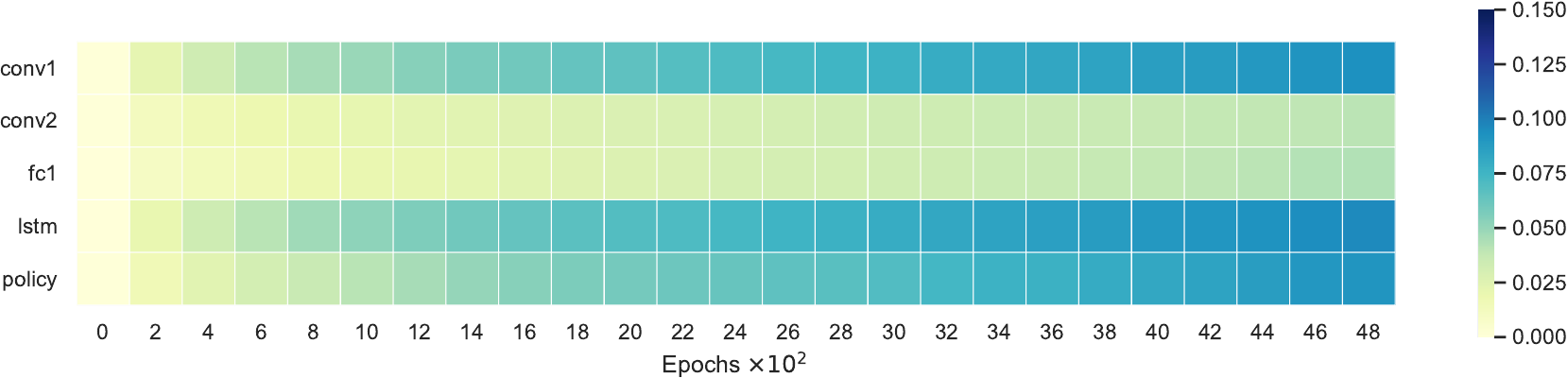}
    \caption{$\ell_2$-norm (no DR, no proprioception)}
  \end{subfigure}
  \begin{subfigure}{0.32\textwidth}
    \includegraphics[width=\textwidth]{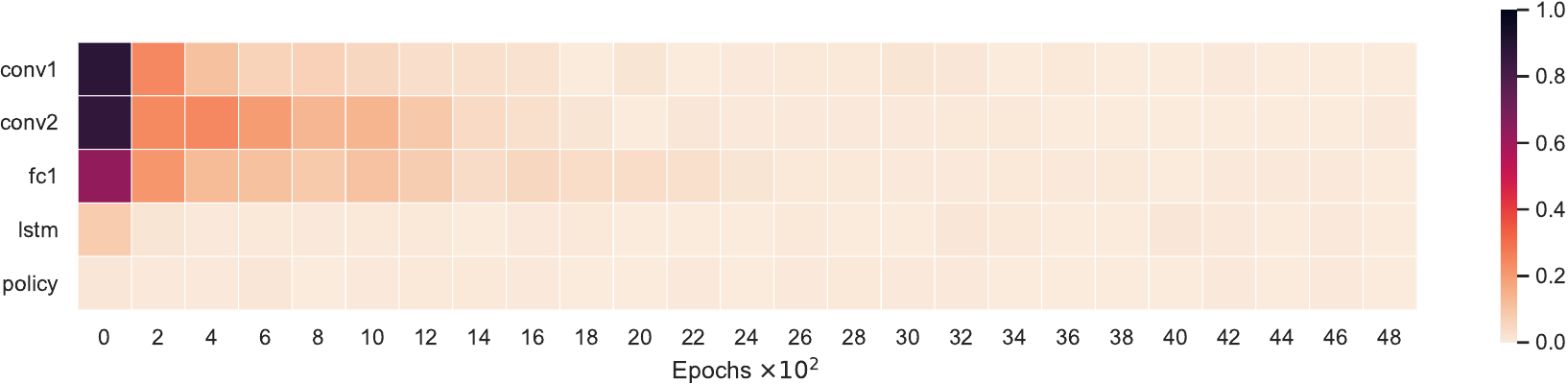}
    \caption{Robustness (no DR, proprioception)}
  \end{subfigure}
  \begin{subfigure}{0.32\textwidth}
    \includegraphics[width=\textwidth]{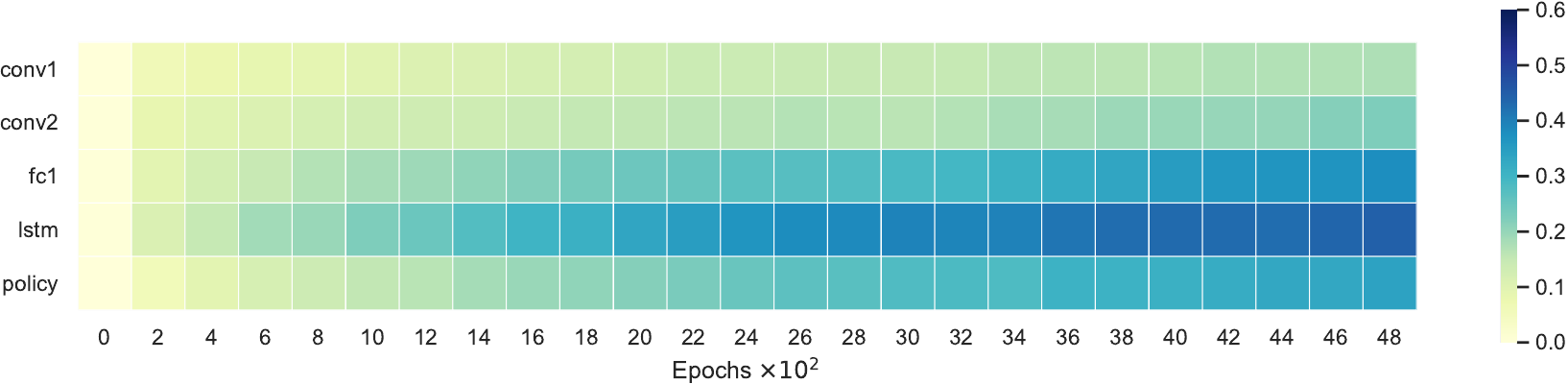}
    \caption{$\ell_\infty$-norm (no DR, proprioception)}
  \end{subfigure}
  \begin{subfigure}{0.32\textwidth}
    \includegraphics[width=\textwidth]{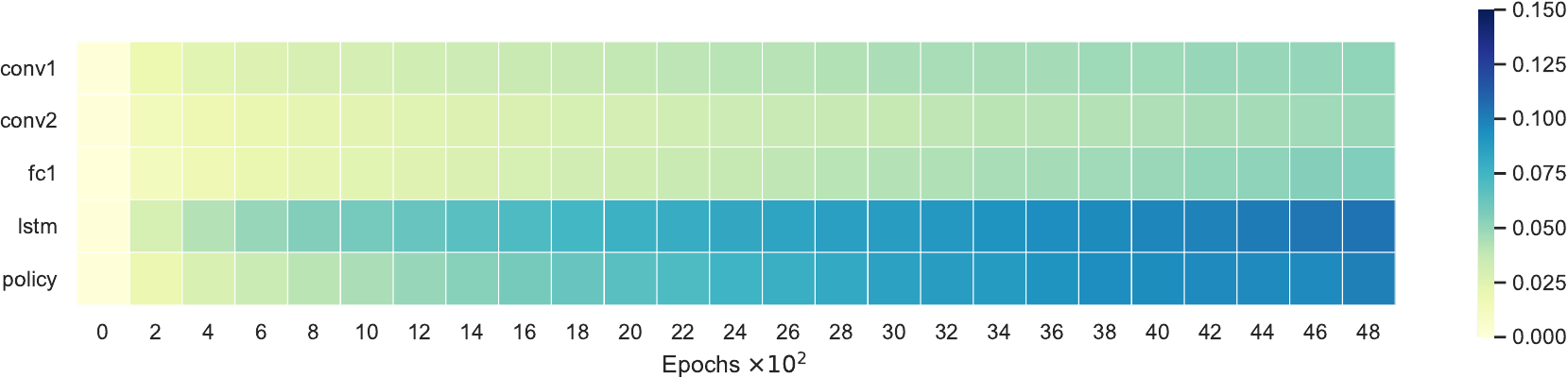}
    \caption{$\ell_2$-norm (no DR, proprioception)}
  \end{subfigure}
  \begin{subfigure}{0.32\textwidth}
    \includegraphics[width=\textwidth]{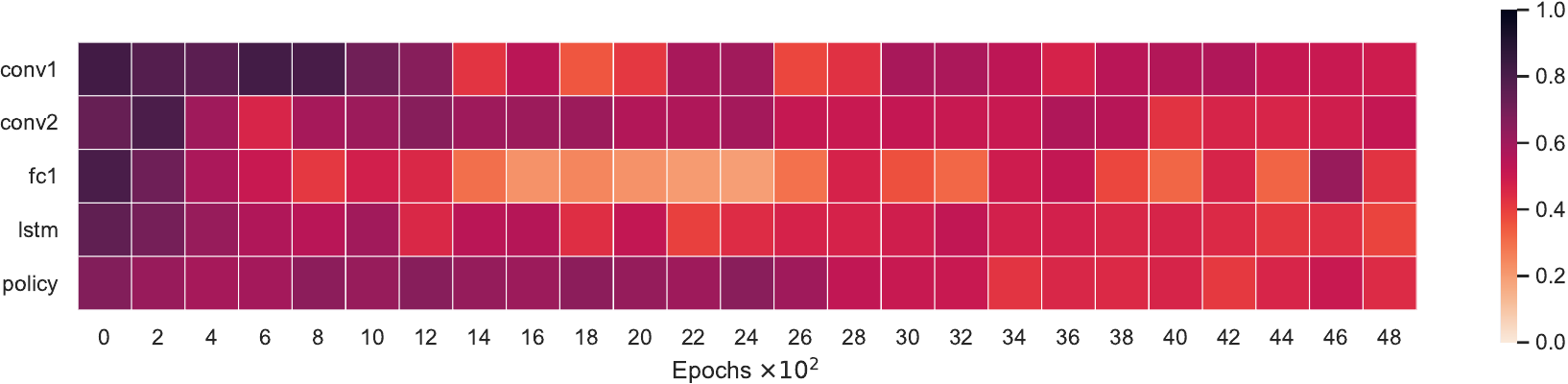}
    \caption{Robustness (DR, no proprioception)}
  \end{subfigure}
  \begin{subfigure}{0.32\textwidth}
    \includegraphics[width=\textwidth]{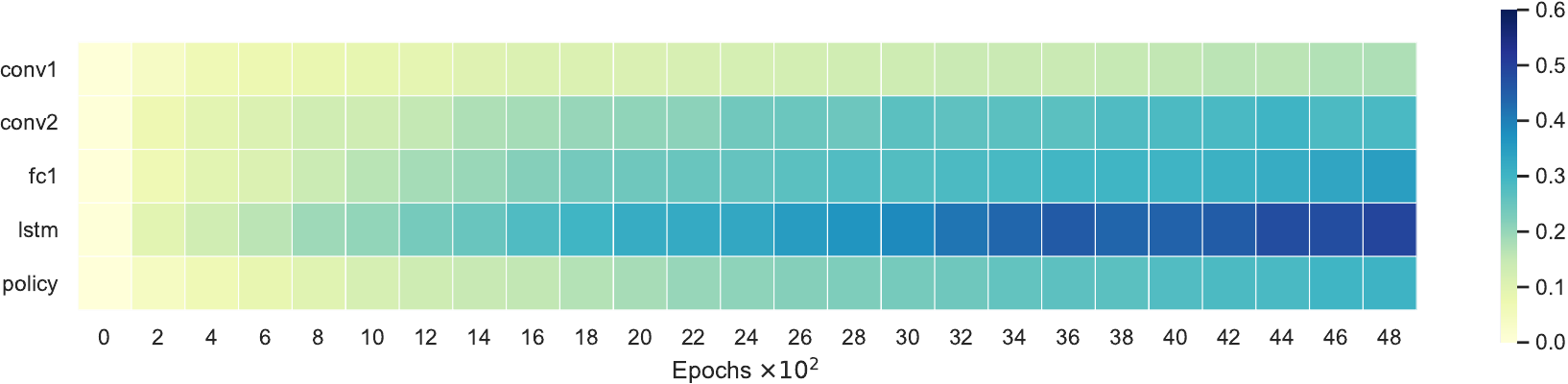}
    \caption{$\ell_\infty$-norm (DR, no proprioception)}
  \end{subfigure}
  \begin{subfigure}{0.32\textwidth}
    \includegraphics[width=\textwidth]{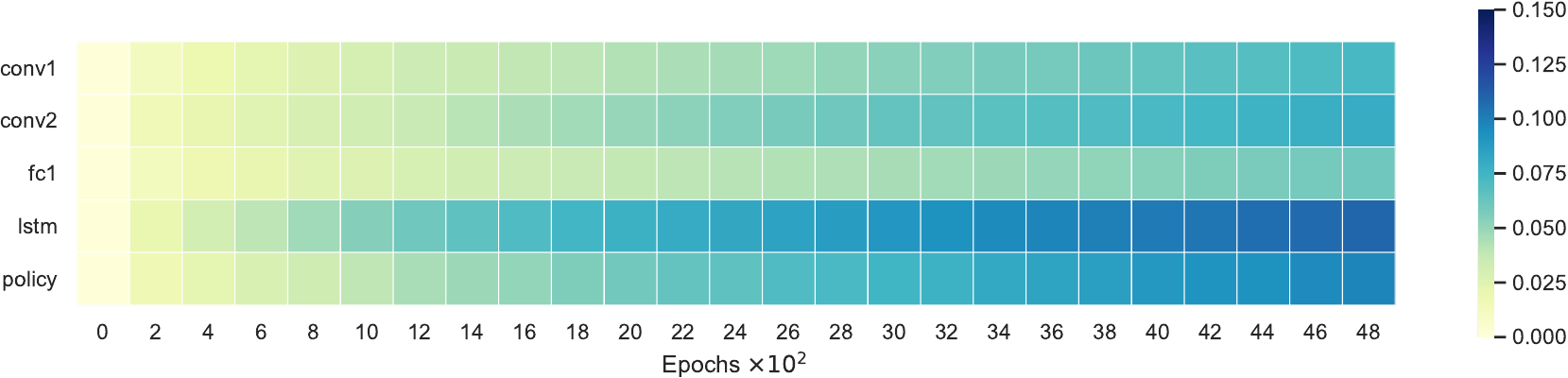}
    \caption{$\ell_2$-norm (DR, no proprioception)}
  \end{subfigure}
  \begin{subfigure}{0.32\textwidth}
    \includegraphics[width=\textwidth]{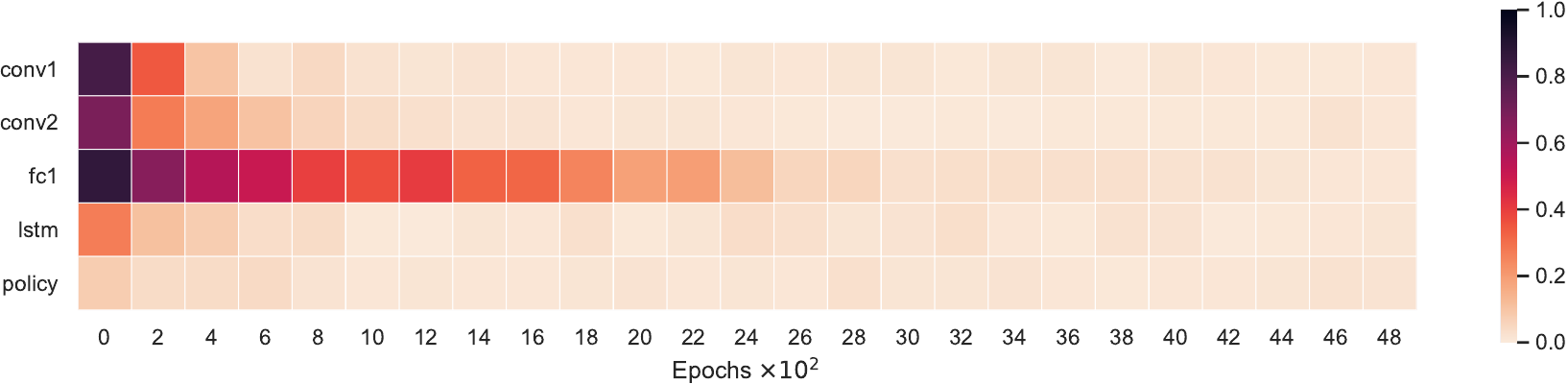}
    \caption{Robustness (DR, proprioception)}
  \end{subfigure}
  \begin{subfigure}{0.32\textwidth}
    \includegraphics[width=\textwidth]{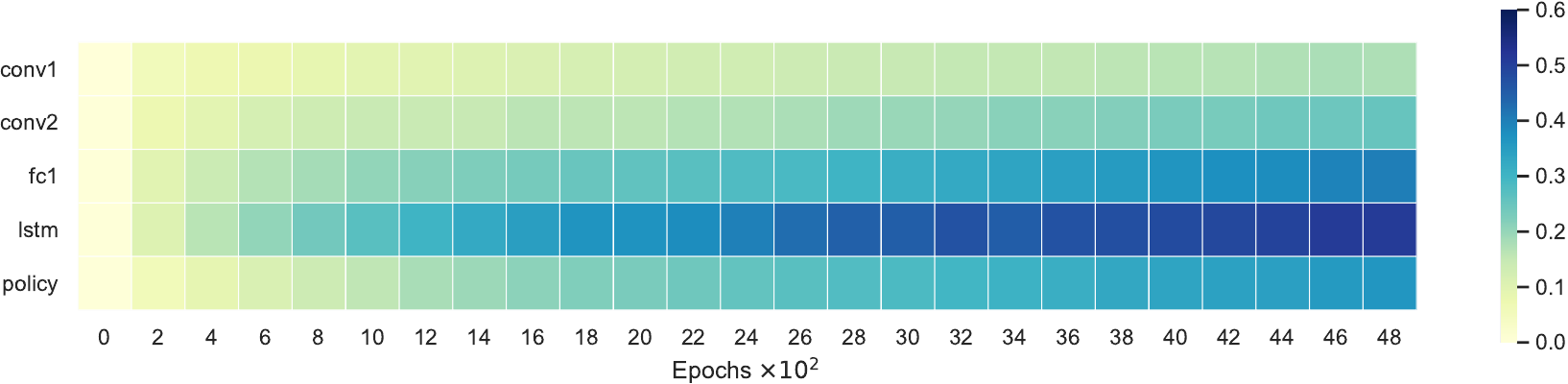}
    \caption{$\ell_\infty$-norm (DR, proprioception)}
  \end{subfigure}
  \begin{subfigure}{0.32\textwidth}
    \includegraphics[width=\textwidth]{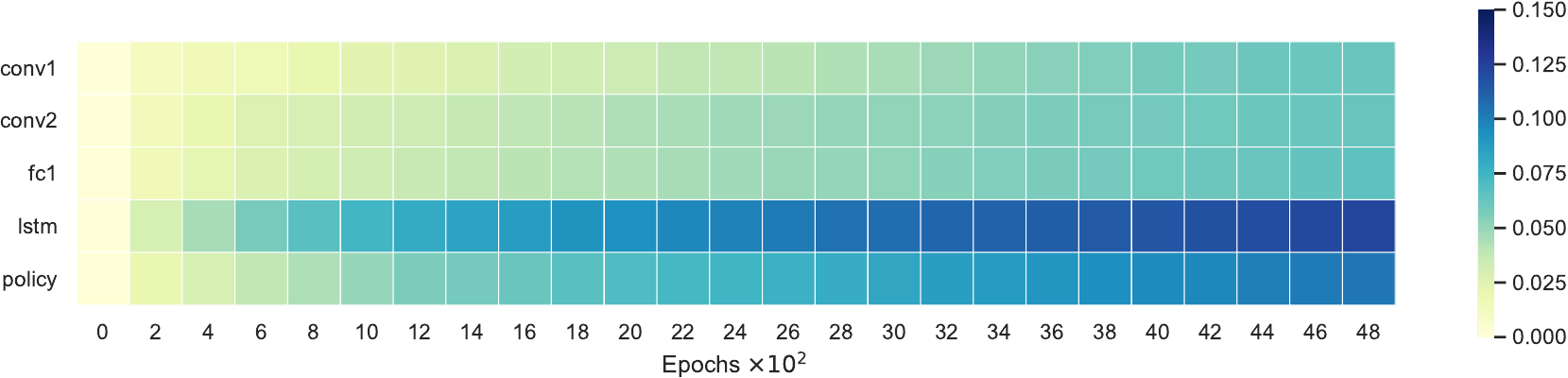}
    \caption{$\ell_2$-norm (DR, proprioception)}
  \end{subfigure}
  \caption{Re-initialisation robustness (1 indicates complete failure, and 0 indicates complete success), and change in $\ell_\infty$- and $\ell_2$-norm of parameters of Jaco agents trained with (g-l) and without (a-f) DR, and with (d-f, j-l) and without (a-c, g-i) proprioceptive inputs. Plots were truncated to show detail during initial epochs. The best Jaco model was chosen for each training condition. Note that the final failure rate of the best Jaco agent trained with DR and without proprioception on the standard environment is around 20\%. The re-initialisation robustness plot for this condition (g) indicates that all layers are necessary and that training continues to improve performance in the epochs depicted and beyond.}
  \label{fig:jaco_robustness}
\end{figure*}

Moving on from unit ablations, we now show the re-initialisation
robustness, as well as the change in \(\ell_\infty\)- and
\(\ell_2\)-norms of the parameters of my trained Fetch and Jaco agents
in Figures \ref{fig:fetch_robustness} and \ref{fig:jaco_robustness},
respectively. We use re-initialisation robustness to study the effect of
task complexity (training with and without DR, and with and without
proprioceptive inputs), but with networks of similar capacity. Our
results are mostly in line with Zhang et al.
\citeyearpar{zhang2019all}---despite continual changes in the weights
during training (as measured by weight norms), the latter layers of the
network are robust to re-initialisation after a few epochs of training,
and in the case of the Fetch agents, the policy layer is robust to
re-initialisation to the original set of weights. The agents trained
with DR are less robust to re-initialisation during
early-to-intermediate stages of training, implying that meaningful
changes in the learned representations occur for longer periods within
the entirety of training. In particular, the Jaco agent trained with DR
and without proprioception continues to improve for a significantly
longer duration than all other models.

For nearly all agents, the recurrent layer is quite robust to
re-initialisation to the original set of weights (despite noticeable
changes in the weights as measured by both the \(\ell_\infty\)-
\(\ell_2\)-norms)---while this does not necessarily indicate that the
agents do not utilise information over time, it does imply that training
the recurrent connections is largely unnecessary for these tasks---a
hypothesis we test further in Subsection \ref{sec:recurrent_ablation}.
While the fully-connected layer benefits from training during the
initial epoch across all models, it takes particularly long to train in
the case of the Jaco agent trained with DR and proprioception (Figure
\ref{fig:jaco_robustness}j), indicating the difficulty of the task.

\hypertarget{recurrent-ablation-1}{%
\subsection{Recurrent Ablation}\label{recurrent-ablation-1}}

\label{sec:recurrent_ablation}

To test how useful the LSTM is, we set the hidden and cell states to
constant values and re-evaluated all models. Rather than naively zeroing
the hidden states, which may not be representative of the values during
rollouts, we instead use the empirical average values, as calculated
over the normal execution of the models in testing. Table
\ref{tbl:hidden_ablation} shows the results of this ablation---there is
a slight effect for agents trained without DR, but a significant effect
for agents trained with DR. This indicates that recurrent processing may
not be necessary for solving either robotic task without DR, but it is
useful when DR is active.

\begin{table}
  \caption{Test performance of all models with standard operation versus constant (empirical average) hidden states. Checkmarks and crosses indicate enabling/disabling DR and proprioceptive inputs (Prop.), respectively. Statistics are calculated over all models (seeds) and test target locations.}
  \label{tbl:hidden_ablation}
  \centering
  \begin{tabular}{c|cc|cc}
    \toprule
    Robot & DR & Prop. & Standard & Constant Hidden\\
    \midrule
    Fetch & \textcolor{red}{\xmark}   & \textcolor{red}{\xmark}   & 1.000$\pm 0.000$ & 0.988$\pm 0.016$\\
    Fetch & \textcolor{red}{\xmark}   & \textcolor{green}{\cmark} & 1.000$\pm 0.000$ & 0.990$\pm 0.012$\\
    Fetch & \textcolor{green}{\cmark} & \textcolor{red}{\xmark}   & 0.983$\pm 0.004$ & 0.658$\pm 0.106$\\
    Fetch & \textcolor{green}{\cmark} & \textcolor{green}{\cmark} & 0.997$\pm 0.002$ & 0.838$\pm 0.051$\\
    \hline
    Jaco  & \textcolor{red}{\xmark}   & \textcolor{red}{\xmark}   & 0.995$\pm 0.003$ & 0.919$\pm 0.032$\\
    Jaco  & \textcolor{red}{\xmark}   & \textcolor{green}{\cmark} & 0.995$\pm 0.001$ & 0.943$\pm 0.022$\\
    Jaco  & \textcolor{green}{\cmark} & \textcolor{red}{\xmark}   & 0.650$\pm 0.056$ & 0.422$\pm 0.083$\\
    Jaco  & \textcolor{green}{\cmark} & \textcolor{green}{\cmark} & 0.991$\pm 0.004$ & 0.746$\pm 0.060$\\
    \bottomrule
  \end{tabular}
\end{table}

\hypertarget{entanglement-1}{%
\subsection{Entanglement}\label{entanglement-1}}

\label{sec:activation_analysis}

\begin{figure*}
  \centering
  \begin{subfigure}{0.22\textwidth}
    \includegraphics[width=\textwidth]{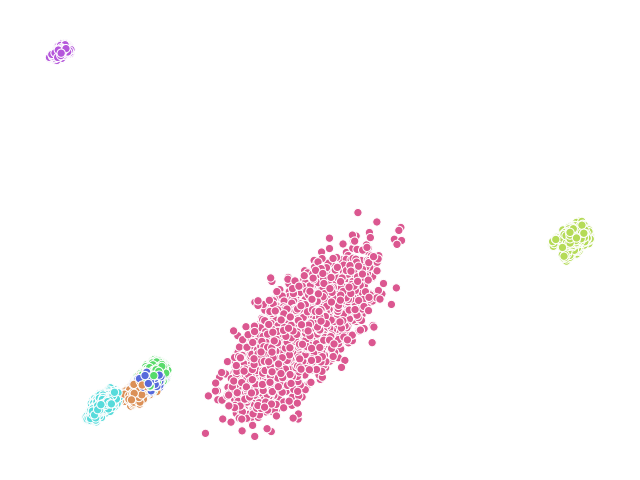}
    \caption{Conv. 1 (PCA; no DR)}
  \end{subfigure}
  \begin{subfigure}{0.22\textwidth}
    \includegraphics[width=\textwidth]{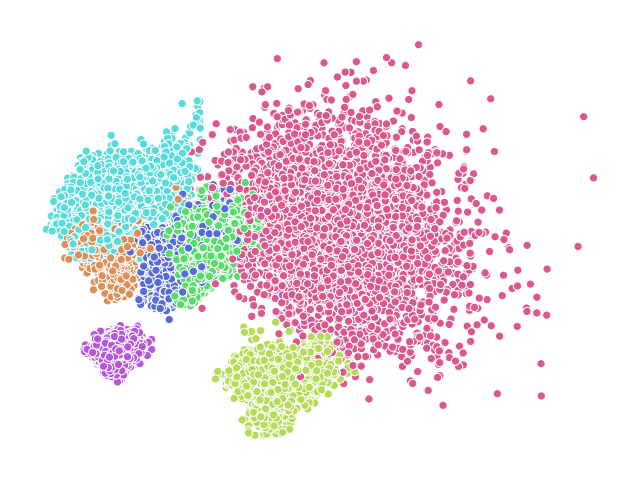}
    \caption{Conv. 2 (PCA; no DR)}
  \end{subfigure}
  \begin{subfigure}{0.22\textwidth}
    \includegraphics[width=\textwidth]{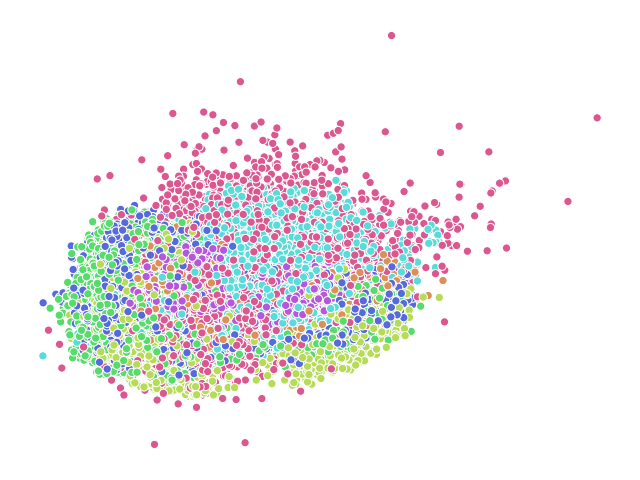}
    \caption{FC (PCA; no DR)}
  \end{subfigure}
  \begin{subfigure}{0.22\textwidth}
    \includegraphics[width=\textwidth]{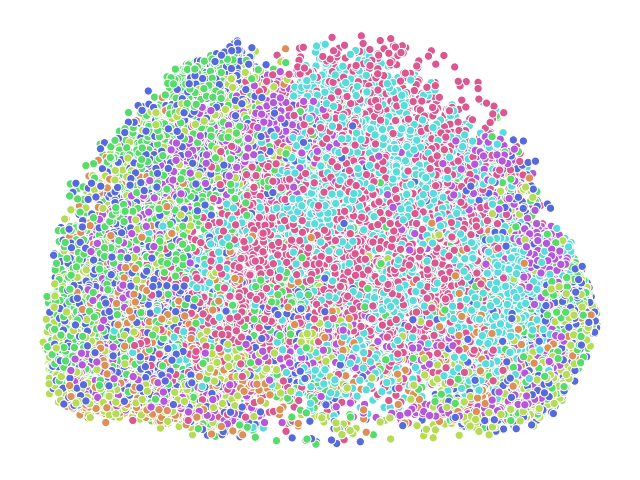}
    \caption{LSTM (PCA; no DR)}
  \end{subfigure}

  \begin{subfigure}{0.22\textwidth}
    \includegraphics[width=\textwidth]{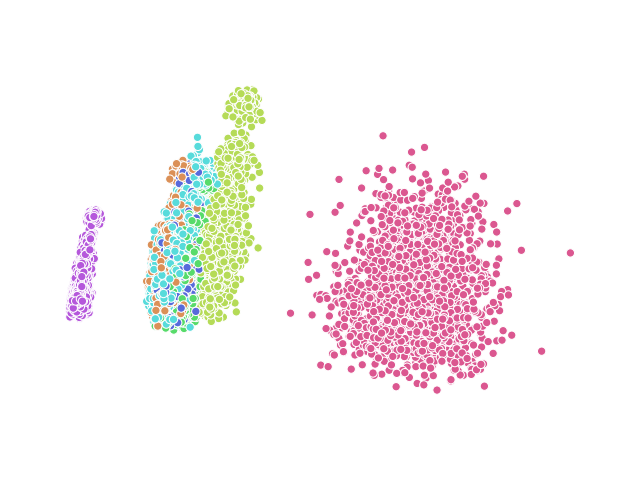}
    \caption{Conv. 1 (PCA; DR)}
  \end{subfigure}
  \begin{subfigure}{0.22\textwidth}
    \includegraphics[width=\textwidth]{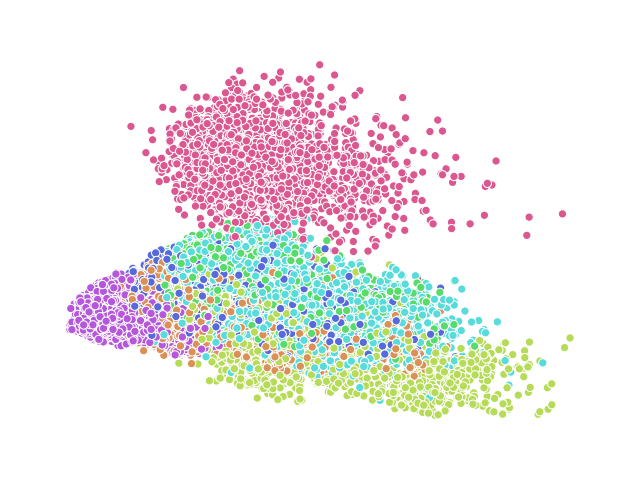}
    \caption{Conv. 2 (PCA; DR)}
  \end{subfigure}
  \begin{subfigure}{0.22\textwidth}
    \includegraphics[width=\textwidth]{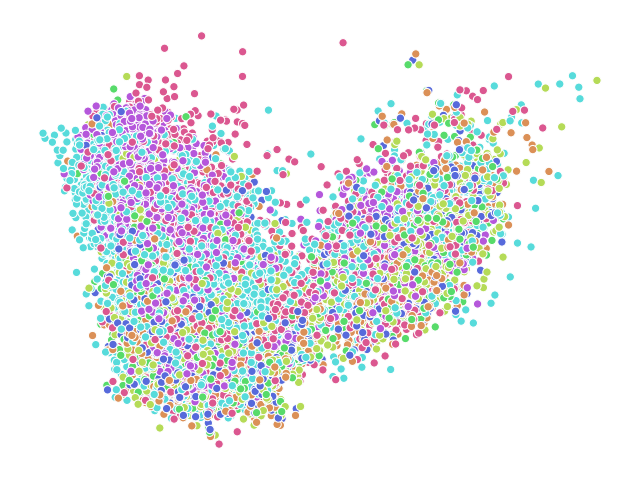}
    \caption{FC (PCA; DR)}
  \end{subfigure}
  \begin{subfigure}{0.22\textwidth}
    \includegraphics[width=\textwidth]{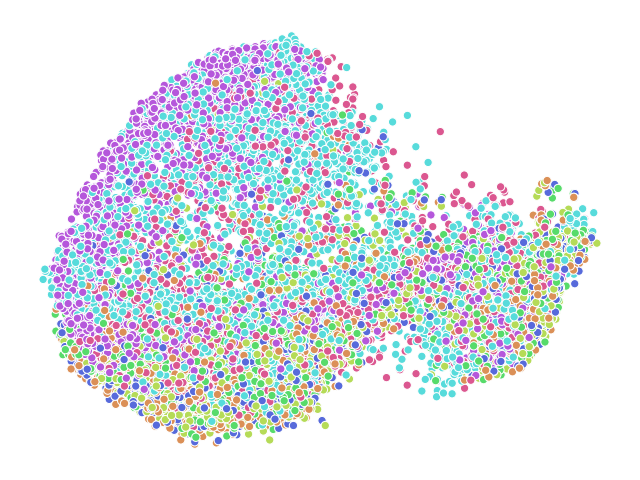}
    \caption{LSTM (PCA; DR)}
  \end{subfigure}

  \begin{subfigure}{0.22\textwidth}
    \includegraphics[width=\textwidth]{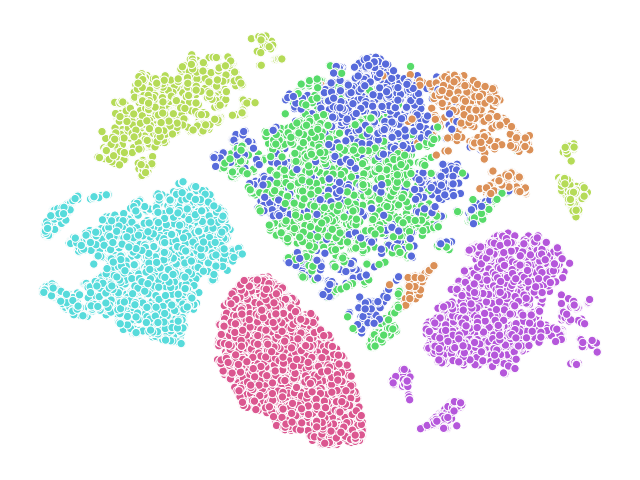}
    \caption{Conv. 1 (t-SNE; no DR)}
  \end{subfigure}
  \begin{subfigure}{0.22\textwidth}
    \includegraphics[width=\textwidth]{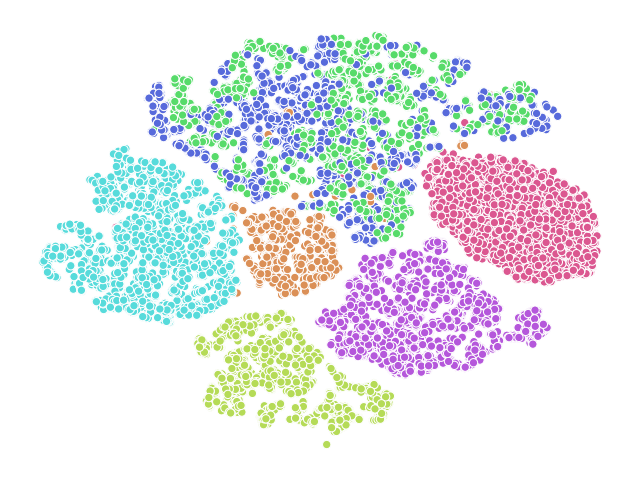}
    \caption{Conv. 2 (t-SNE; no DR)}
  \end{subfigure}
  \begin{subfigure}{0.22\textwidth}
    \includegraphics[width=\textwidth]{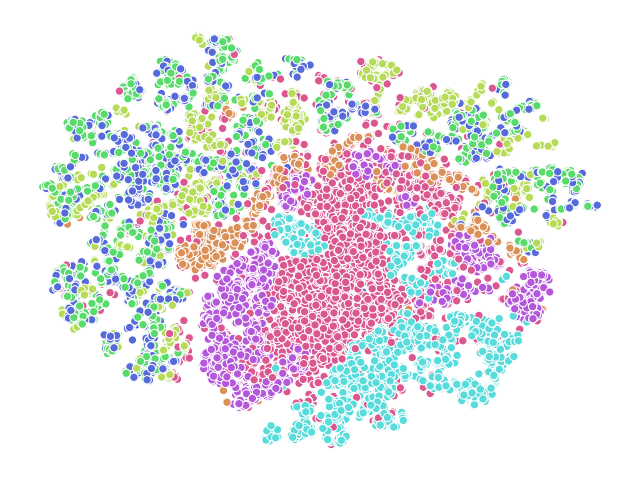}
    \caption{FC (t-SNE; no DR)}
  \end{subfigure}
  \begin{subfigure}{0.22\textwidth}
    \includegraphics[width=\textwidth]{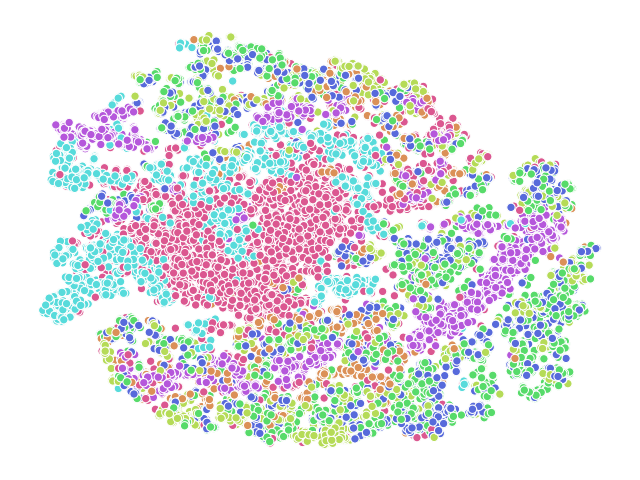}
    \caption{LSTM (t-SNE; no DR)}
  \end{subfigure}

  \begin{subfigure}{0.22\textwidth}
    \includegraphics[width=\textwidth]{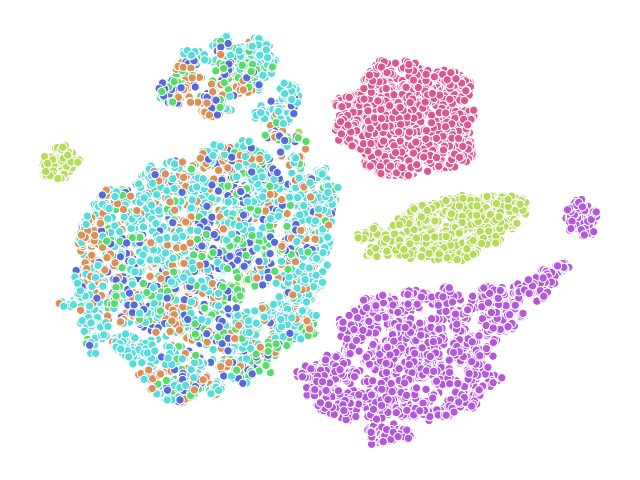}
    \caption{Conv. 1 (t-SNE; DR)}
  \end{subfigure}
  \begin{subfigure}{0.22\textwidth}
    \includegraphics[width=\textwidth]{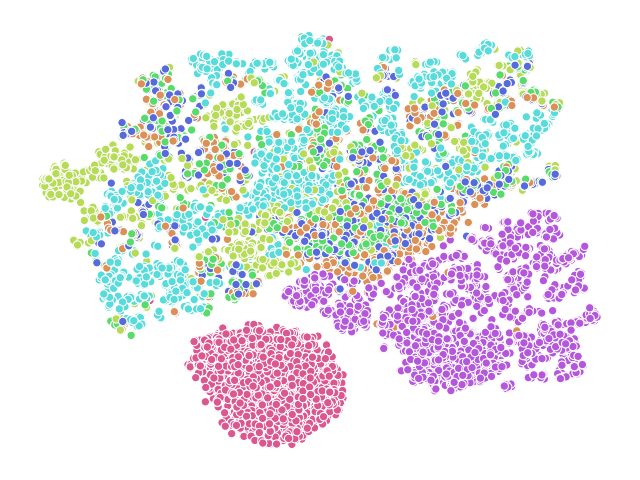}
    \caption{Conv. 2 (t-SNE; DR)}
  \end{subfigure}
  \begin{subfigure}{0.22\textwidth}
    \includegraphics[width=\textwidth]{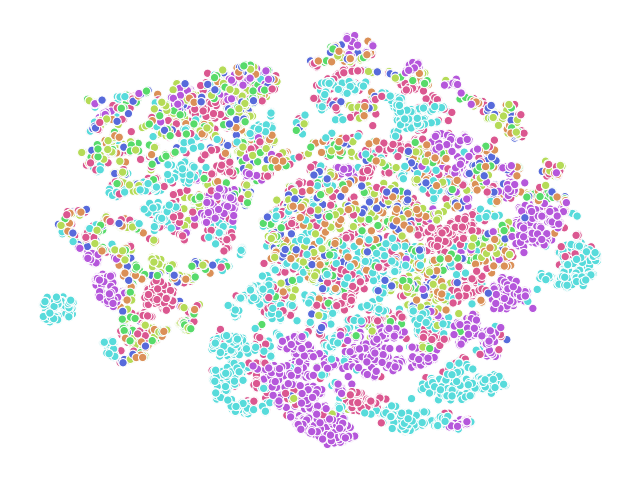}
    \caption{FC (t-SNE; DR)}
  \end{subfigure}
  \begin{subfigure}{0.22\textwidth}
    \includegraphics[width=\textwidth]{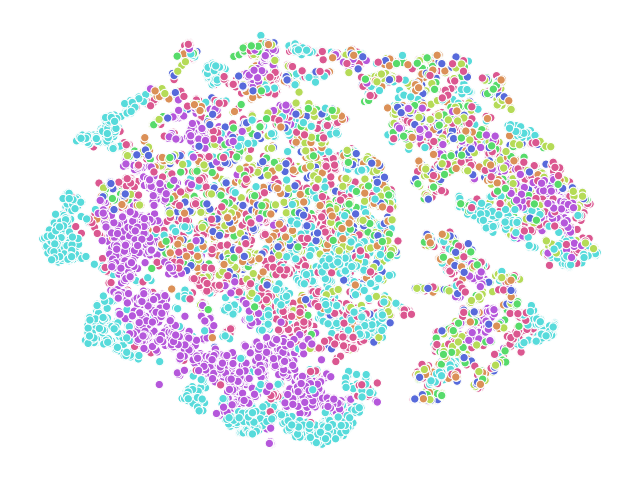}
    \caption{LSTM (t-SNE; DR)}
  \end{subfigure}

  \begin{subfigure}{0.22\textwidth}
    \includegraphics[width=\textwidth]{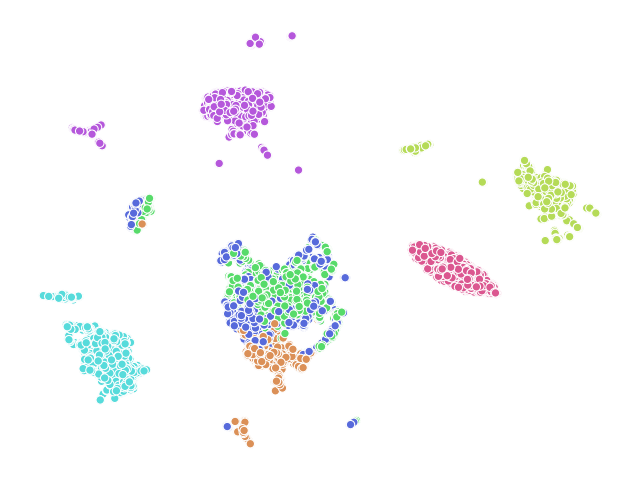}
    \caption{Conv. 1 (UMAP; no DR)}
  \end{subfigure}
  \begin{subfigure}{0.22\textwidth}
    \includegraphics[width=\textwidth]{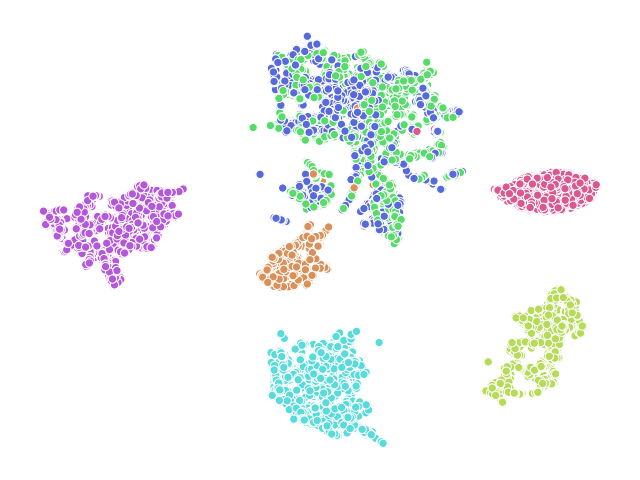}
    \caption{Conv. 2 (UMAP; no DR)}
  \end{subfigure}
  \begin{subfigure}{0.22\textwidth}
    \includegraphics[width=\textwidth]{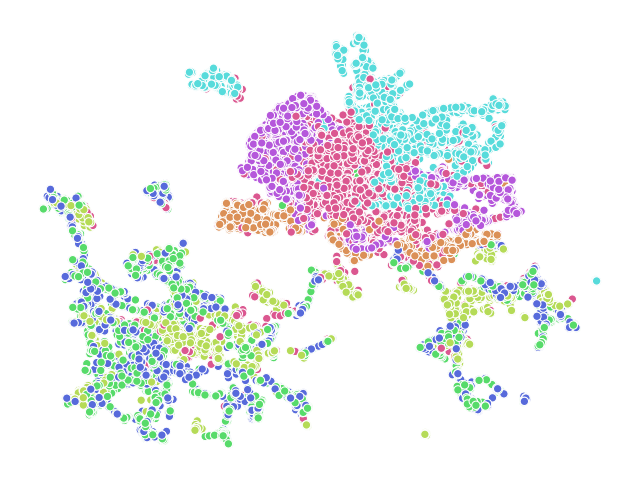}
    \caption{FC (UMAP; no DR)}
  \end{subfigure}
  \begin{subfigure}{0.22\textwidth}
    \includegraphics[width=\textwidth]{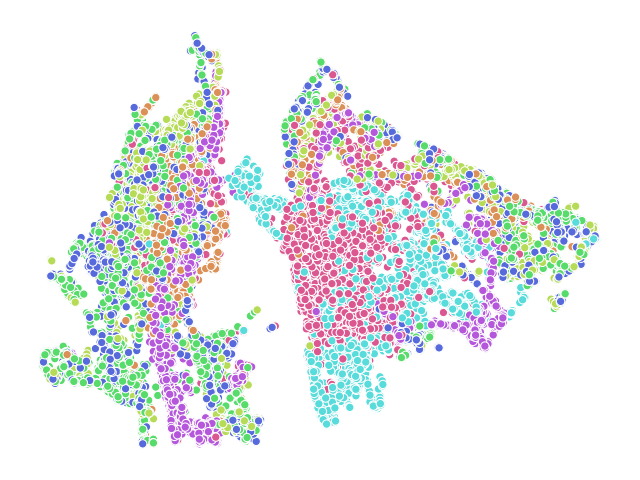}
    \caption{LSTM (UMAP; no DR)}
  \end{subfigure}

  \begin{subfigure}{0.22\textwidth}
    \includegraphics[width=\textwidth]{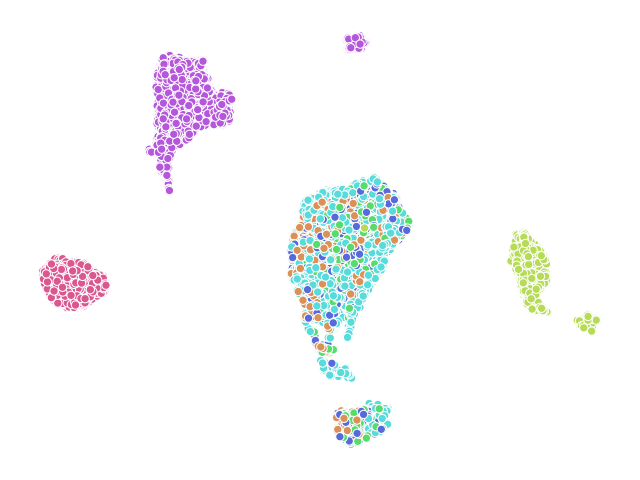}
    \caption{Conv. 1 (UMAP; DR)}
  \end{subfigure}
  \begin{subfigure}{0.22\textwidth}
    \includegraphics[width=\textwidth]{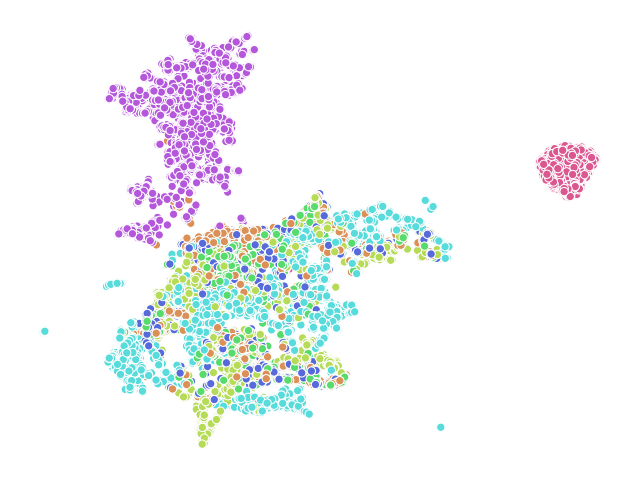}
    \caption{Conv. 2 (UMAP; DR)}
  \end{subfigure}
  \begin{subfigure}{0.22\textwidth}
    \includegraphics[width=\textwidth]{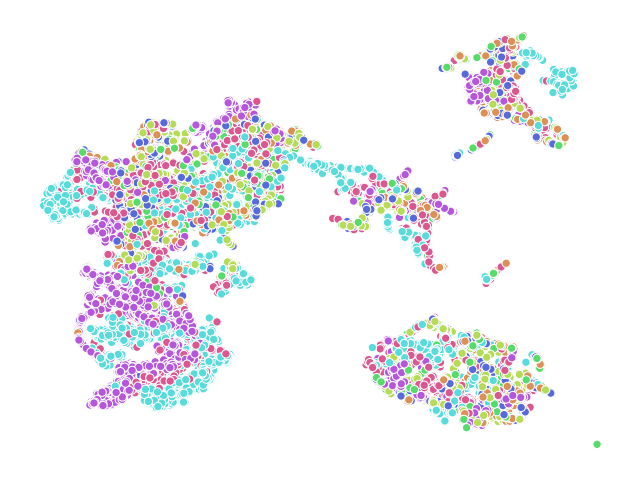}
    \caption{FC (UMAP; DR)}
  \end{subfigure}
  \begin{subfigure}{0.22\textwidth}
    \includegraphics[width=\textwidth]{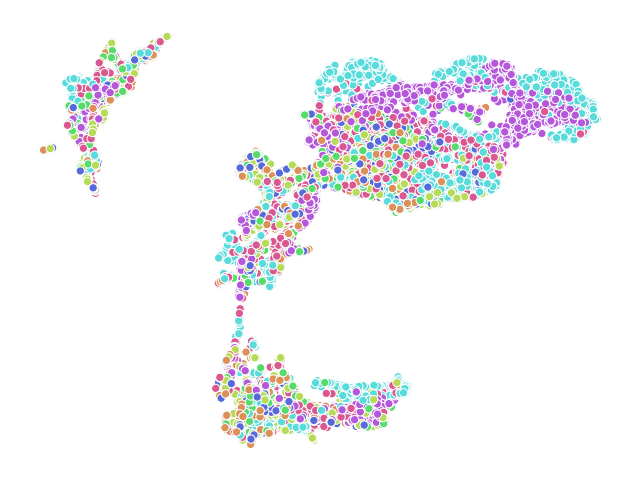}
    \caption{LSTM (UMAP; DR)}
  \end{subfigure}
  \caption{Embeddings for trained Jaco agents with proprioceptive inputs with and without DR. Test conditions that are entangled with the normal observations (orange) typically include changing the colour (dark blue) or shape (green) of the target, shifting the camera (light blue), and, for DR, adding reflections (yellow). Global changes---adding Gaussian noise (red) or changing the global lighting (purple)---are the least entangled with the normal observations.}
  \label{fig:act_embeddings}
\end{figure*}

\begin{table}
  \caption{Entanglement scores of different agents, for the first and second convolutional (conv.), fully-connected (FC) and LSTM layer, calculated over different testing conditions as classes (with $T = 0$). Checkmarks and crosses indicate enabling/disabling DR and proprioceptive inputs (Prop.), respectively.}
  \label{tbl:entanglement}
  \centering
  \resizebox{\linewidth}{!}{
  \begin{tabular}{c|cc|cccc}
    \toprule
    Robot & DR & Prop. & 1\textsuperscript{st} Conv. & 2\textsuperscript{nd} Conv. & FC & LSTM\\
    \midrule
    Fetch & \textcolor{red}{\xmark}   & \textcolor{red}{\xmark}   & 0.11 & 0.30 & 0.56 & 0.68 \\
    Fetch & \textcolor{red}{\xmark}   & \textcolor{green}{\cmark} & 0.12 & 0.30 & 0.45 & 0.45 \\
    Fetch & \textcolor{green}{\cmark} & \textcolor{red}{\xmark}   & 0.23 & 0.38 & 0.62 & 0.92 \\
    Fetch & \textcolor{green}{\cmark} & \textcolor{green}{\cmark} & 0.24 & 0.41 & 0.58 & 1.15 \\
    \hline
    Jaco  & \textcolor{red}{\xmark}   & \textcolor{red}{\xmark}   & 0.14 & 0.29 & 0.52 & 0.68 \\
    Jaco  & \textcolor{red}{\xmark}   & \textcolor{green}{\cmark} & 0.11 & 0.08 & 0.43 & 0.66 \\
    Jaco  & \textcolor{green}{\cmark} & \textcolor{red}{\xmark}   & 0.41 & 0.37 & 0.55 & 0.73 \\
    Jaco  & \textcolor{green}{\cmark} & \textcolor{green}{\cmark} & 0.65 & 0.56 & 1.21 & 1.37 \\
    \bottomrule
  \end{tabular}}
\end{table}

Firstly, we consider the quantitative analysis of activations from
different trained agents under the different training conditions. Table
\ref{tbl:entanglement} contains the entanglement scores
\citep{frosst2019analyzing} of the different trained agents, calculated
across the first 4 layers (not including the policy/value outputs); as
with the original work, we use a 2D t-SNE \citep{maaten2008visualizing}
embedding for the activations. There are two noticeable trends. Firstly,
the entanglement scores increase deeper into the network; this supports
the notion that the different testing conditions can result in very
different visual observations, but the difference between them
diminishes as they are further processed by the networks. Secondly, the
agents trained with DR have noticeably higher entanglement scores for
each layer as compared to their equivalents trained without DR. This
quantitatively supports the idea that DR makes agents largely invariant
to nuisance visual factors (as opposed to the agents finding different
strategies to cope with different visual conditions).

We can also qualitatively support these findings by visualising the same
activations in 2D (Figure \ref{fig:act_embeddings}). We use three common
embedding techniques in order to show different aspects of the data.
Firstly, we use PCA \citep{pearson1901liii}, which linearly embeds the
data into dimensions which explain the most variance in the original
data; as a result, linearly separable clusters have very different
global characteristics. Secondly, we use t-SNE
\citep{maaten2008visualizing}, which attempts to retain local structure
in the data by calculating pairwise similarities between datapoints and
creating a constrained graph layout in which distances in the original
high-dimensional and the low-dimensional projection are preserved as
much as possible. Thirdly, we use uniform manifold approximation and
projection (UMAP) \citep{mcinnes2018umap}, which operates similarly to
t-SNE at a high level, but better preserves global structure. Although
it is possible to tune t-SNE \citep{wattenberg2016use}, by default, UMAP
better shows relevant global structure.

\hypertarget{discussion}{%
\section{Discussion}\label{discussion}}

\label{sec:discussion}

A primary goal of these experiments was to uncover the effects of DR,
through a comparison between agents trained either with or without DR.
In line with prior work, DR improves performance across a wider
distribution of testing conditions. In particular, our implementation of
DR, which varied colours and textures, allowed generalisation to
scenarios with ``local'' perturbations, but was more variable when more
global changes were made to the setup; overall, agents trained with DR
were nearly always more robust than agents trained without (Subsection
\ref{sec:test_scenarios}). Adding DR to a task makes it more challenging
to solve, in terms of sample complexity, although under the current
experimental setup the models do not appear to require additional
architectural depth, as all agents\footnote{Except for the Jaco agent
  trained with DR and without proprioception, which has lower final
  performance under standard visuals.} are robust to re-initialisation
of the final (policy) layer (Subsection \ref{sec:layer_ablations}). The
application of entanglement \citep{frosst2019analyzing}, with respect to
visual perturbations, shows that throughout the network the
representations that are learned appear to be more invariant to these
changes in the visuals, as the embeddings of representations from the
different conditions have higher overlap (Subsection
\ref{sec:activation_analysis}).

At the lower levels of the networks, DR results in significant changes
in the \(\ell_1\)-norms of the convolutional filters (Subsection
\ref{sec:statistical_and_structural}), with more sophisticated feature
detectors (Subsection \ref{sec:activation_maximisation}). Supporting
this, visualising the saliency maps of the agents shows that DR agents
have more focused attention on task-specific features, such as the arm
or ball (Subsection \ref{sec:saliency_maps}). Counter to initial
expectations, we did not find that DR reduced the variability of
performance under convolutional filter ablations---the agents merely
have better baseline performance (Subsection \ref{sec:unit_ablations}).
Deeper within the networks, we found that DR caused the agents to
utilise the recurrent dynamics of the LSTM, whilst the agents trained
without DR were hardly impacted by keeping their recurrent state
constant (Subsection \ref{sec:recurrent_ablation}).

While we observe these general trends, it is notable that some of the
results are not \emph{a priori} as obvious. For example, even when
provided with proprioceptive inputs, the Fetch agent trained without DR
still uses its visual inputs for self-localisation (Subsection
\ref{sec:saliency_maps}), although the addition of DR removes this
observed effect. We believe that the relative simplicity of the Fetch
reaching task---including both sensing and actuation---leads to less
pronounced effects with DR (Subsection \ref{sec:test_scenarios}). The
most unexpected finding was that the performance of the Jaco agent
trained with DR and without proprioception dropped when shifting from DR
visuals to the standard simulator visuals, demonstrating that DR can
overfit (Subsection \ref{sec:domain_shift}). With proprioception the gap
disappears, which supports the idea that the form of input can have a
significant effect on generalisation in agents
\citep{hill2019emergent}---meriting further investigation.

This work has focused on understanding the effects of DR, but also has a
dual purpose, which is to inform research in an opposite sense: in
situations where DR is expensive or even infeasible, what approaches can
we take to improve generalisation in sim2real transfer? If certain
characteristics are positively correlated with DR training, explicitly
enforcing them---without requiring the DR pipeline---may also lead to
improved generalisation. For instance, Cobbe et al.
\citeyearpar{cobbe2018quantifying} showed that standard regularisation
techniques improve generalisation to a limited extent. Similarly, Pinto
et al. \citeyearpar{pinto2017robust} showed that adversarial training
could improve the robustness of DRL policies. In line with this,
enforcing greater spatial structure in the convolutional filters
(Subsection \ref{sec:spectral}), which is higher in agents trained with
DR (Subsection \ref{sec:statistical_and_structural}), could be used as a
novel regularisation objective.

A broader goal of these experiments was to assess the suitability of
interpretability methods within the context of DRL. Beyond noticing
limitations as discussed in previous works
\citep{mahendran2015understanding, kindermans2016investigating}, there
is a larger positive outcome from using a wide suite of interpretability
techniques. Firstly, when used together they can cross-check the
validity of each other's results. For example, supposedly ``dead'' units
in the Jaco model with DR and proprioceptive inputs do in fact worsen
performance when ablated (Subsection \ref{sec:activation_maximisation}).
Additionally, although the LSTM layer within DR agents are robust to
re-initialisation at early stages of training (Subsection
\ref{sec:layer_ablations}), the recurrent ablations show that the agents
depend heavily on recurrent processing (Subsection
\ref{sec:recurrent_ablation}). Secondly, the complementary answers these
techniques provide leads to a better understanding of the model as a
whole. For instance, unit ablations (Subsection
\ref{sec:unit_ablations}) can be related to diversity in activation
maximisation (Subsection \ref{sec:activation_maximisation}), and
entanglement (Subsection \ref{sec:activation_analysis}) can explain the
generalisation of agents trained with DR (Subsection
\ref{sec:test_scenarios}). Given these benefits, we therefore recommend
a holistic approach of interpretability techniques to be able to draw
correct and informative conclusions.

\FloatBarrier


\bibliographystyle{cas-model2-names}
\bibliography{references}

\begin{thebibliography}{115}
\expandafter\ifx\csname natexlab\endcsname\relax\def\natexlab#1{#1}\fi
\providecommand{\url}[1]{\texttt{#1}}
\providecommand{\href}[2]{#2}
\providecommand{\path}[1]{#1}
\providecommand{\DOIprefix}{doi:}
\providecommand{\ArXivprefix}{arXiv:}
\providecommand{\URLprefix}{URL: }
\providecommand{\Pubmedprefix}{pmid:}
\providecommand{\doi}[1]{\href{http://dx.doi.org/#1}{\path{#1}}}
\providecommand{\Pubmed}[1]{\href{pmid:#1}{\path{#1}}}
\providecommand{\bibinfo}[2]{#2}
\ifx\xfnm\relax \def\xfnm[#1]{\unskip,\space#1}\fi
\bibitem[{Andrychowicz et~al.(2018)Andrychowicz, Baker, Chociej, Jozefowicz,
  McGrew, Pachocki, Petron, Plappert, Powell, Ray
  et~al.}]{andrychowicz2018learning}
\bibinfo{author}{Andrychowicz, M.}, \bibinfo{author}{Baker, B.},
  \bibinfo{author}{Chociej, M.}, \bibinfo{author}{Jozefowicz, R.},
  \bibinfo{author}{McGrew, B.}, \bibinfo{author}{Pachocki, J.},
  \bibinfo{author}{Petron, A.}, \bibinfo{author}{Plappert, M.},
  \bibinfo{author}{Powell, G.}, \bibinfo{author}{Ray, A.}, et~al.,
  \bibinfo{year}{2018}.
\newblock \bibinfo{title}{Learning dexterous in-hand manipulation}.
\newblock \bibinfo{journal}{arXiv preprint arXiv:1808.00177} .
\bibitem[{Arulkumaran et~al.(2017)Arulkumaran, Deisenroth, Brundage and
  Bharath}]{arulkumaran2017deep}
\bibinfo{author}{Arulkumaran, K.}, \bibinfo{author}{Deisenroth, M.P.},
  \bibinfo{author}{Brundage, M.}, \bibinfo{author}{Bharath, A.A.},
  \bibinfo{year}{2017}.
\newblock \bibinfo{title}{Deep reinforcement learning: A brief survey}.
\newblock \bibinfo{journal}{IEEE Signal Processing Magazine}
  \bibinfo{volume}{34}, \bibinfo{pages}{26--38}.
\bibitem[{Aubry and Russell(2015)}]{aubry2015understanding}
\bibinfo{author}{Aubry, M.}, \bibinfo{author}{Russell, B.C.},
  \bibinfo{year}{2015}.
\newblock \bibinfo{title}{Understanding deep features with computer-generated
  imagery}, in: \bibinfo{booktitle}{International Conference on Computer
  Vision}, pp. \bibinfo{pages}{2875--2883}.
\bibitem[{Bach et~al.(2015)Bach, Binder, Montavon, Klauschen, M{\"u}ller and
  Samek}]{bach2015pixel}
\bibinfo{author}{Bach, S.}, \bibinfo{author}{Binder, A.},
  \bibinfo{author}{Montavon, G.}, \bibinfo{author}{Klauschen, F.},
  \bibinfo{author}{M{\"u}ller, K.R.}, \bibinfo{author}{Samek, W.},
  \bibinfo{year}{2015}.
\newblock \bibinfo{title}{On pixel-wise explanations for non-linear classifier
  decisions by layer-wise relevance propagation}.
\newblock \bibinfo{journal}{PloS one} \bibinfo{volume}{10},
  \bibinfo{pages}{e0130140}.
\bibitem[{Barto et~al.(1983)Barto, Sutton and Anderson}]{barto1983neuronlike}
\bibinfo{author}{Barto, A.G.}, \bibinfo{author}{Sutton, R.S.},
  \bibinfo{author}{Anderson, C.W.}, \bibinfo{year}{1983}.
\newblock \bibinfo{title}{Neuronlike adaptive elements that can solve difficult
  learning control problems}.
\newblock \bibinfo{journal}{IEEE Transactions on Systems, Man, and Cybernetics}
  , \bibinfo{pages}{834--846}.
\bibitem[{Bellido and Fiesler(1993)}]{bellido1993backpropagation}
\bibinfo{author}{Bellido, I.}, \bibinfo{author}{Fiesler, E.},
  \bibinfo{year}{1993}.
\newblock \bibinfo{title}{Do backpropagation trained neural networks have
  normal weight distributions?}, in: \bibinfo{booktitle}{International
  Conference on Artificial Neural Networks}, \bibinfo{organization}{Springer}.
  pp. \bibinfo{pages}{772--775}.
\bibitem[{Bousmalis et~al.(2018)Bousmalis, Irpan, Wohlhart, Bai, Kelcey,
  Kalakrishnan, Downs, Ibarz, Pastor, Konolige et~al.}]{bousmalis2018using}
\bibinfo{author}{Bousmalis, K.}, \bibinfo{author}{Irpan, A.},
  \bibinfo{author}{Wohlhart, P.}, \bibinfo{author}{Bai, Y.},
  \bibinfo{author}{Kelcey, M.}, \bibinfo{author}{Kalakrishnan, M.},
  \bibinfo{author}{Downs, L.}, \bibinfo{author}{Ibarz, J.},
  \bibinfo{author}{Pastor, P.}, \bibinfo{author}{Konolige, K.}, et~al.,
  \bibinfo{year}{2018}.
\newblock \bibinfo{title}{Using simulation and domain adaptation to improve
  efficiency of deep robotic grasping}, in: \bibinfo{booktitle}{International
  Conference on Robotics and Automation}, \bibinfo{organization}{IEEE}. pp.
  \bibinfo{pages}{4243--4250}.
\bibitem[{Brockman et~al.(2016)Brockman, Cheung, Pettersson, Schneider,
  Schulman, Tang and Zaremba}]{brockman2016openai}
\bibinfo{author}{Brockman, G.}, \bibinfo{author}{Cheung, V.},
  \bibinfo{author}{Pettersson, L.}, \bibinfo{author}{Schneider, J.},
  \bibinfo{author}{Schulman, J.}, \bibinfo{author}{Tang, J.},
  \bibinfo{author}{Zaremba, W.}, \bibinfo{year}{2016}.
\newblock \bibinfo{title}{{OpenAI Gym}}.
\newblock \bibinfo{journal}{arXiv preprint arXiv:1606.01540} .
\bibitem[{Chebotar et~al.(2018)Chebotar, Handa, Makoviychuk, Macklin, Issac,
  Ratliff and Fox}]{chebotar2018closing}
\bibinfo{author}{Chebotar, Y.}, \bibinfo{author}{Handa, A.},
  \bibinfo{author}{Makoviychuk, V.}, \bibinfo{author}{Macklin, M.},
  \bibinfo{author}{Issac, J.}, \bibinfo{author}{Ratliff, N.},
  \bibinfo{author}{Fox, D.}, \bibinfo{year}{2018}.
\newblock \bibinfo{title}{Closing the sim-to-real loop: Adapting simulation
  randomization with real world experience}.
\newblock \bibinfo{journal}{arXiv preprint arXiv:1810.05687} .
\bibitem[{Cobbe et~al.(2018)Cobbe, Klimov, Hesse, Kim and
  Schulman}]{cobbe2018quantifying}
\bibinfo{author}{Cobbe, K.}, \bibinfo{author}{Klimov, O.},
  \bibinfo{author}{Hesse, C.}, \bibinfo{author}{Kim, T.},
  \bibinfo{author}{Schulman, J.}, \bibinfo{year}{2018}.
\newblock \bibinfo{title}{Quantifying generalization in reinforcement
  learning}.
\newblock \bibinfo{journal}{arXiv preprint arXiv:1812.02341} .
\bibitem[{Craven and Shavlik(1996)}]{craven1996extracting}
\bibinfo{author}{Craven, M.}, \bibinfo{author}{Shavlik, J.W.},
  \bibinfo{year}{1996}.
\newblock \bibinfo{title}{Extracting tree-structured representations of trained
  networks}, in: \bibinfo{booktitle}{Advances in Neural Information Processing
  Systems}, pp. \bibinfo{pages}{24--30}.
\bibitem[{Deisenroth et~al.(2013)Deisenroth, Neumann, Peters
  et~al.}]{deisenroth2013survey}
\bibinfo{author}{Deisenroth, M.P.}, \bibinfo{author}{Neumann, G.},
  \bibinfo{author}{Peters, J.}, et~al., \bibinfo{year}{2013}.
\newblock \bibinfo{title}{A survey on policy search for robotics}.
\newblock \bibinfo{journal}{Foundations and Trends{\textregistered} in
  Robotics} \bibinfo{volume}{2}, \bibinfo{pages}{1--142}.
\bibitem[{Donahue et~al.(2014)Donahue, Jia, Vinyals, Hoffman, Zhang, Tzeng and
  Darrell}]{donahue2014decaf}
\bibinfo{author}{Donahue, J.}, \bibinfo{author}{Jia, Y.},
  \bibinfo{author}{Vinyals, O.}, \bibinfo{author}{Hoffman, J.},
  \bibinfo{author}{Zhang, N.}, \bibinfo{author}{Tzeng, E.},
  \bibinfo{author}{Darrell, T.}, \bibinfo{year}{2014}.
\newblock \bibinfo{title}{Decaf: A deep convolutional activation feature for
  generic visual recognition}, in: \bibinfo{booktitle}{International Conference
  on Machine Learning}, pp. \bibinfo{pages}{647--655}.
\bibitem[{Doshi-Velez and Kim(2017)}]{doshi2017towards}
\bibinfo{author}{Doshi-Velez, F.}, \bibinfo{author}{Kim, B.},
  \bibinfo{year}{2017}.
\newblock \bibinfo{title}{Towards a rigorous science of interpretable machine
  learning}.
\newblock \bibinfo{journal}{arXiv preprint arXiv:1702.08608} .
\bibitem[{Elman(1989)}]{elman1989representation}
\bibinfo{author}{Elman, J.L.}, \bibinfo{year}{1989}.
\newblock \bibinfo{title}{Representation and structure in connectionist
  models}.
\newblock \bibinfo{type}{Technical Report}. Univ. of California at San Diego,
  La Jolla Center For Research In Language.
\bibitem[{Erhan et~al.(2009)Erhan, Bengio, Courville and
  Vincent}]{erhan2009visualizing}
\bibinfo{author}{Erhan, D.}, \bibinfo{author}{Bengio, Y.},
  \bibinfo{author}{Courville, A.}, \bibinfo{author}{Vincent, P.},
  \bibinfo{year}{2009}.
\newblock \bibinfo{title}{Visualizing higher-layer features of a deep network}.
\newblock \bibinfo{type}{Technical Report} \bibinfo{number}{1341}. University
  of Montreal.
\bibitem[{Fran{\c{c}}ois-Lavet et~al.(2018)Fran{\c{c}}ois-Lavet, Henderson,
  Islam, Bellemare, Pineau et~al.}]{franccois2018introduction}
\bibinfo{author}{Fran{\c{c}}ois-Lavet, V.}, \bibinfo{author}{Henderson, P.},
  \bibinfo{author}{Islam, R.}, \bibinfo{author}{Bellemare, M.G.},
  \bibinfo{author}{Pineau, J.}, et~al., \bibinfo{year}{2018}.
\newblock \bibinfo{title}{An introduction to deep reinforcement learning}.
\newblock \bibinfo{journal}{Foundations and Trends{\textregistered} in Machine
  Learning} \bibinfo{volume}{11}, \bibinfo{pages}{219--354}.
\bibitem[{Freitas(2014)}]{freitas2014comprehensible}
\bibinfo{author}{Freitas, A.A.}, \bibinfo{year}{2014}.
\newblock \bibinfo{title}{Comprehensible classification models: a position
  paper}.
\newblock \bibinfo{journal}{Explorations Newsletter} \bibinfo{volume}{15},
  \bibinfo{pages}{1--10}.
\bibitem[{Frosst et~al.(2019)Frosst, Papernot and Hinton}]{frosst2019analyzing}
\bibinfo{author}{Frosst, N.}, \bibinfo{author}{Papernot, N.},
  \bibinfo{author}{Hinton, G.}, \bibinfo{year}{2019}.
\newblock \bibinfo{title}{Analyzing and improving representations with the soft
  nearest neighbor loss}, in: \bibinfo{booktitle}{International Conference on
  Machine Learning}, pp. \bibinfo{pages}{2012--2020}.
\bibitem[{Gers et~al.(2000)Gers, Schmidhuber and Cummins}]{gers2000learning}
\bibinfo{author}{Gers, F.A.}, \bibinfo{author}{Schmidhuber, J.},
  \bibinfo{author}{Cummins, F.}, \bibinfo{year}{2000}.
\newblock \bibinfo{title}{Learning to forget: Continual prediction with lstm}.
\newblock \bibinfo{journal}{Neural Computation} \bibinfo{volume}{12},
  \bibinfo{pages}{2451--2471}.
\bibitem[{Girshick et~al.(2014)Girshick, Donahue, Darrell and
  Malik}]{girshick2014rich}
\bibinfo{author}{Girshick, R.}, \bibinfo{author}{Donahue, J.},
  \bibinfo{author}{Darrell, T.}, \bibinfo{author}{Malik, J.},
  \bibinfo{year}{2014}.
\newblock \bibinfo{title}{Rich feature hierarchies for accurate object
  detection and semantic segmentation}, in: \bibinfo{booktitle}{Conference on
  Computer Vision and Pattern Recognition}, pp. \bibinfo{pages}{580--587}.
\bibitem[{Glorot and Bengio(2010)}]{glorot2010understanding}
\bibinfo{author}{Glorot, X.}, \bibinfo{author}{Bengio, Y.},
  \bibinfo{year}{2010}.
\newblock \bibinfo{title}{Understanding the difficulty of training deep
  feedforward neural networks}, in: \bibinfo{booktitle}{International
  Conference on Artificial Intelligence and Statistics}, pp.
  \bibinfo{pages}{249--256}.
\bibitem[{Greydanus et~al.(2018)Greydanus, Koul, Dodge and
  Fern}]{greydanus2018visualizing}
\bibinfo{author}{Greydanus, S.}, \bibinfo{author}{Koul, A.},
  \bibinfo{author}{Dodge, J.}, \bibinfo{author}{Fern, A.},
  \bibinfo{year}{2018}.
\newblock \bibinfo{title}{Visualizing and understanding atari agents}, in:
  \bibinfo{booktitle}{International Conference on Machine Learning}, pp.
  \bibinfo{pages}{1787--1796}.
\bibitem[{Gu et~al.(2017)Gu, Holly, Lillicrap and Levine}]{gu2017deep}
\bibinfo{author}{Gu, S.}, \bibinfo{author}{Holly, E.},
  \bibinfo{author}{Lillicrap, T.}, \bibinfo{author}{Levine, S.},
  \bibinfo{year}{2017}.
\newblock \bibinfo{title}{Deep reinforcement learning for robotic manipulation
  with asynchronous off-policy updates}, in: \bibinfo{booktitle}{International
  Conference on Robotics and Automation}, \bibinfo{organization}{IEEE}. pp.
  \bibinfo{pages}{3389--3396}.
\bibitem[{Guidotti et~al.(2018)Guidotti, Monreale, Ruggieri, Turini, Giannotti
  and Pedreschi}]{guidotti2018survey}
\bibinfo{author}{Guidotti, R.}, \bibinfo{author}{Monreale, A.},
  \bibinfo{author}{Ruggieri, S.}, \bibinfo{author}{Turini, F.},
  \bibinfo{author}{Giannotti, F.}, \bibinfo{author}{Pedreschi, D.},
  \bibinfo{year}{2018}.
\newblock \bibinfo{title}{A survey of methods for explaining black box models}.
\newblock \bibinfo{journal}{ACM Computing Surveys} \bibinfo{volume}{51},
  \bibinfo{pages}{93}.
\bibitem[{Hamel and Eck(2010)}]{hamel2010learning}
\bibinfo{author}{Hamel, P.}, \bibinfo{author}{Eck, D.}, \bibinfo{year}{2010}.
\newblock \bibinfo{title}{Learning features from music audio with deep belief
  networks.}, in: \bibinfo{booktitle}{ISMIR}, \bibinfo{organization}{Utrecht,
  The Netherlands}. pp. \bibinfo{pages}{339--344}.
\bibitem[{Han et~al.(2015)Han, Pool, Tran and Dally}]{han2015learning}
\bibinfo{author}{Han, S.}, \bibinfo{author}{Pool, J.}, \bibinfo{author}{Tran,
  J.}, \bibinfo{author}{Dally, W.}, \bibinfo{year}{2015}.
\newblock \bibinfo{title}{Learning both weights and connections for efficient
  neural network}, in: \bibinfo{booktitle}{Advances in Neural Information
  Processing Systems}, pp. \bibinfo{pages}{1135--1143}.
\bibitem[{Hansen and Salamon(1990)}]{hansen1990neural}
\bibinfo{author}{Hansen, L.K.}, \bibinfo{author}{Salamon, P.},
  \bibinfo{year}{1990}.
\newblock \bibinfo{title}{Neural network ensembles}.
\newblock \bibinfo{journal}{IEEE Transactions on Pattern Analysis \& Machine
  Intelligence} , \bibinfo{pages}{993--1001}.
\bibitem[{Hanson and Burr(1990)}]{hanson1990connectionist}
\bibinfo{author}{Hanson, S.J.}, \bibinfo{author}{Burr, D.J.},
  \bibinfo{year}{1990}.
\newblock \bibinfo{title}{What connectionist models learn: Learning and
  representation in connectionist networks}.
\newblock \bibinfo{journal}{Behavioral and Brain Sciences}
  \bibinfo{volume}{13}, \bibinfo{pages}{471--489}.
\bibitem[{Hanson and Pratt(1989)}]{hanson1989comparing}
\bibinfo{author}{Hanson, S.J.}, \bibinfo{author}{Pratt, L.Y.},
  \bibinfo{year}{1989}.
\newblock \bibinfo{title}{Comparing biases for minimal network construction
  with back-propagation}, in: \bibinfo{booktitle}{Advances in Neural
  Information Processing Systems}, pp. \bibinfo{pages}{177--185}.
\bibitem[{Hassibi and Stork(1993)}]{hassibi1993second}
\bibinfo{author}{Hassibi, B.}, \bibinfo{author}{Stork, D.G.},
  \bibinfo{year}{1993}.
\newblock \bibinfo{title}{Second order derivatives for network pruning: Optimal
  brain surgeon}, in: \bibinfo{booktitle}{Advances in Neural Information
  Processing Systems}, pp. \bibinfo{pages}{164--171}.
\bibitem[{He et~al.(2015)He, Zhang, Ren and Sun}]{he2015delving}
\bibinfo{author}{He, K.}, \bibinfo{author}{Zhang, X.}, \bibinfo{author}{Ren,
  S.}, \bibinfo{author}{Sun, J.}, \bibinfo{year}{2015}.
\newblock \bibinfo{title}{Delving deep into rectifiers: Surpassing human-level
  performance on imagenet classification}, in:
  \bibinfo{booktitle}{International Conference on Computer Vision}, pp.
  \bibinfo{pages}{1026--1034}.
\bibitem[{Hill et~al.(2019)Hill, Lampinen, Schneider, Clark, Botvinick,
  McClelland and Santoro}]{hill2019emergent}
\bibinfo{author}{Hill, F.}, \bibinfo{author}{Lampinen, A.},
  \bibinfo{author}{Schneider, R.}, \bibinfo{author}{Clark, S.},
  \bibinfo{author}{Botvinick, M.}, \bibinfo{author}{McClelland, J.L.},
  \bibinfo{author}{Santoro, A.}, \bibinfo{year}{2019}.
\newblock \bibinfo{title}{Emergent systematic generalization in a situated
  agent}.
\newblock \bibinfo{journal}{arXiv preprint arXiv:1910.00571} .
\bibitem[{Hinton and Shallice(1991)}]{hinton1991lesioning}
\bibinfo{author}{Hinton, G.E.}, \bibinfo{author}{Shallice, T.},
  \bibinfo{year}{1991}.
\newblock \bibinfo{title}{Lesioning an attractor network: Investigations of
  acquired dyslexia.}
\newblock \bibinfo{journal}{Psychological review} \bibinfo{volume}{98},
  \bibinfo{pages}{74}.
\bibitem[{Hochreiter and Schmidhuber(1997)}]{hochreiter1997long}
\bibinfo{author}{Hochreiter, S.}, \bibinfo{author}{Schmidhuber, J.},
  \bibinfo{year}{1997}.
\newblock \bibinfo{title}{Long short-term memory}.
\newblock \bibinfo{journal}{Neural Computation} \bibinfo{volume}{9},
  \bibinfo{pages}{1735--1780}.
\bibitem[{Ilyas et~al.(2018)Ilyas, Engstrom, Santurkar, Tsipras, Janoos,
  Rudolph and Madry}]{ilyas2018deep}
\bibinfo{author}{Ilyas, A.}, \bibinfo{author}{Engstrom, L.},
  \bibinfo{author}{Santurkar, S.}, \bibinfo{author}{Tsipras, D.},
  \bibinfo{author}{Janoos, F.}, \bibinfo{author}{Rudolph, L.},
  \bibinfo{author}{Madry, A.}, \bibinfo{year}{2018}.
\newblock \bibinfo{title}{Are deep policy gradient algorithms truly policy
  gradient algorithms?}
\newblock \bibinfo{journal}{arXiv preprint arXiv:1811.02553} .
\bibitem[{Jakobi et~al.(1995)Jakobi, Husbands and Harvey}]{jakobi1995noise}
\bibinfo{author}{Jakobi, N.}, \bibinfo{author}{Husbands, P.},
  \bibinfo{author}{Harvey, I.}, \bibinfo{year}{1995}.
\newblock \bibinfo{title}{Noise and the reality gap: The use of simulation in
  evolutionary robotics}, in: \bibinfo{booktitle}{European Conference on
  Artificial Life}, \bibinfo{organization}{Springer}. pp.
  \bibinfo{pages}{704--720}.
\bibitem[{James et~al.(2017)James, Davison and Johns}]{james2017transferring}
\bibinfo{author}{James, S.}, \bibinfo{author}{Davison, A.J.},
  \bibinfo{author}{Johns, E.}, \bibinfo{year}{2017}.
\newblock \bibinfo{title}{Transferring end-to-end visuomotor control from
  simulation to real world for a multi-stage task}, in:
  \bibinfo{booktitle}{Conference on Robot Learning}, pp.
  \bibinfo{pages}{334--343}.
\bibitem[{Justesen et~al.(2018)Justesen, Torrado, Bontrager, Khalifa, Togelius
  and Risi}]{justesen2018procedural}
\bibinfo{author}{Justesen, N.}, \bibinfo{author}{Torrado, R.R.},
  \bibinfo{author}{Bontrager, P.}, \bibinfo{author}{Khalifa, A.},
  \bibinfo{author}{Togelius, J.}, \bibinfo{author}{Risi, S.},
  \bibinfo{year}{2018}.
\newblock \bibinfo{title}{Procedural level generation improves generality of
  deep reinforcement learning}.
\newblock \bibinfo{journal}{arXiv preprint arXiv:1806.10729} .
\bibitem[{Khabou et~al.(1999)Khabou, Gader and Shi}]{khabou1999entropy}
\bibinfo{author}{Khabou, M.A.}, \bibinfo{author}{Gader, P.D.},
  \bibinfo{author}{Shi, H.}, \bibinfo{year}{1999}.
\newblock \bibinfo{title}{Entropy optimized morphological shared-weight neural
  networks}.
\newblock \bibinfo{journal}{Optical Engineering} \bibinfo{volume}{38},
  \bibinfo{pages}{263--274}.
\bibitem[{Kindermans et~al.(2017)Kindermans, Hooker, Adebayo, Alber,
  Sch{\"u}tt, D{\"a}hne, Erhan and Kim}]{kindermans2017reliability}
\bibinfo{author}{Kindermans, P.J.}, \bibinfo{author}{Hooker, S.},
  \bibinfo{author}{Adebayo, J.}, \bibinfo{author}{Alber, M.},
  \bibinfo{author}{Sch{\"u}tt, K.T.}, \bibinfo{author}{D{\"a}hne, S.},
  \bibinfo{author}{Erhan, D.}, \bibinfo{author}{Kim, B.}, \bibinfo{year}{2017}.
\newblock \bibinfo{title}{The (un) reliability of saliency methods}, in:
  \bibinfo{booktitle}{NeurIPS Interpreting, Explaining and Visualizing Deep
  Learning Workshop}.
\bibitem[{Kindermans et~al.(2016)Kindermans, Sch{\"u}tt, M{\"u}ller and
  D{\"a}hne}]{kindermans2016investigating}
\bibinfo{author}{Kindermans, P.J.}, \bibinfo{author}{Sch{\"u}tt, K.},
  \bibinfo{author}{M{\"u}ller, K.R.}, \bibinfo{author}{D{\"a}hne, S.},
  \bibinfo{year}{2016}.
\newblock \bibinfo{title}{Investigating the influence of noise and distractors
  on the interpretation of neural networks}, in: \bibinfo{booktitle}{NeurIPS
  Interpretable Machine Learning in Complex Systems Workshop}.
\bibitem[{Kingma and Ba(2014)}]{kingma2014adam}
\bibinfo{author}{Kingma, D.P.}, \bibinfo{author}{Ba, J.}, \bibinfo{year}{2014}.
\newblock \bibinfo{title}{Adam: A method for stochastic optimization}.
\newblock \bibinfo{journal}{arXiv preprint arXiv:1412.6980} .
\bibitem[{Kragic and Vincze(2009)}]{kragic2009vision}
\bibinfo{author}{Kragic, D.}, \bibinfo{author}{Vincze, M.},
  \bibinfo{year}{2009}.
\newblock \bibinfo{title}{Vision for robotics}.
\newblock \bibinfo{journal}{Foundations and Trends in Robotics}
  \bibinfo{volume}{1}, \bibinfo{pages}{1--78}.
\bibitem[{Krizhevsky et~al.(2012)Krizhevsky, Sutskever and
  Hinton}]{krizhevsky2012imagenet}
\bibinfo{author}{Krizhevsky, A.}, \bibinfo{author}{Sutskever, I.},
  \bibinfo{author}{Hinton, G.E.}, \bibinfo{year}{2012}.
\newblock \bibinfo{title}{Imagenet classification with deep convolutional
  neural networks}, in: \bibinfo{booktitle}{Advances in Neural Information
  Processing Systems}, pp. \bibinfo{pages}{1097--1105}.
\bibitem[{Krkic et~al.(1996)Krkic, Roberts, Rezek and Jordan}]{krkic1996eeg}
\bibinfo{author}{Krkic, M.}, \bibinfo{author}{Roberts, S.J.},
  \bibinfo{author}{Rezek, I.}, \bibinfo{author}{Jordan, C.},
  \bibinfo{year}{1996}.
\newblock \bibinfo{title}{Eeg-based assessment of anaesthetic depth using
  neural networks} .
\bibitem[{Lapuschkin et~al.(2019)Lapuschkin, W{\"a}ldchen, Binder, Montavon,
  Samek and M{\"u}ller}]{lapuschkin2019unmasking}
\bibinfo{author}{Lapuschkin, S.}, \bibinfo{author}{W{\"a}ldchen, S.},
  \bibinfo{author}{Binder, A.}, \bibinfo{author}{Montavon, G.},
  \bibinfo{author}{Samek, W.}, \bibinfo{author}{M{\"u}ller, K.R.},
  \bibinfo{year}{2019}.
\newblock \bibinfo{title}{Unmasking clever hans predictors and assessing what
  machines really learn}.
\newblock \bibinfo{journal}{Nature Communications} \bibinfo{volume}{10},
  \bibinfo{pages}{1096}.
\bibitem[{LeCun et~al.(1998)LeCun, Bottou, Orr and
  M{\"u}ller}]{lecun1998efficient}
\bibinfo{author}{LeCun, Y.}, \bibinfo{author}{Bottou, L.},
  \bibinfo{author}{Orr, G.B.}, \bibinfo{author}{M{\"u}ller, K.R.},
  \bibinfo{year}{1998}.
\newblock \bibinfo{title}{Efficient backprop}, in: \bibinfo{booktitle}{Neural
  Networks: Tricks of the Trade}. \bibinfo{publisher}{Springer}, pp.
  \bibinfo{pages}{9--50}.
\bibitem[{LeCun et~al.(1990)LeCun, Denker and Solla}]{lecun1990optimal}
\bibinfo{author}{LeCun, Y.}, \bibinfo{author}{Denker, J.S.},
  \bibinfo{author}{Solla, S.A.}, \bibinfo{year}{1990}.
\newblock \bibinfo{title}{Optimal brain damage}, in:
  \bibinfo{booktitle}{Advances in Neural Information Processing Systems}, pp.
  \bibinfo{pages}{598--605}.
\bibitem[{Levine et~al.(2016)Levine, Finn, Darrell and Abbeel}]{levine2016end}
\bibinfo{author}{Levine, S.}, \bibinfo{author}{Finn, C.},
  \bibinfo{author}{Darrell, T.}, \bibinfo{author}{Abbeel, P.},
  \bibinfo{year}{2016}.
\newblock \bibinfo{title}{End-to-end training of deep visuomotor policies}.
\newblock \bibinfo{journal}{The Journal of Machine Learning Research}
  \bibinfo{volume}{17}, \bibinfo{pages}{1334--1373}.
\bibitem[{Levine et~al.(2018)Levine, Pastor, Krizhevsky, Ibarz and
  Quillen}]{levine2018learning}
\bibinfo{author}{Levine, S.}, \bibinfo{author}{Pastor, P.},
  \bibinfo{author}{Krizhevsky, A.}, \bibinfo{author}{Ibarz, J.},
  \bibinfo{author}{Quillen, D.}, \bibinfo{year}{2018}.
\newblock \bibinfo{title}{Learning hand-eye coordination for robotic grasping
  with deep learning and large-scale data collection}.
\newblock \bibinfo{journal}{The International Journal of Robotics Research}
  \bibinfo{volume}{37}, \bibinfo{pages}{421--436}.
\bibitem[{Li et~al.(2017)Li, Kadav, Durdanovic, Samet and Graf}]{li2017pruning}
\bibinfo{author}{Li, H.}, \bibinfo{author}{Kadav, A.},
  \bibinfo{author}{Durdanovic, I.}, \bibinfo{author}{Samet, H.},
  \bibinfo{author}{Graf, H.P.}, \bibinfo{year}{2017}.
\newblock \bibinfo{title}{Pruning filters for efficient convnets}, in:
  \bibinfo{booktitle}{International Conference on Learning Representations}.
\bibitem[{Lin et~al.(2013)Lin, Chen and Yan}]{lin2013network}
\bibinfo{author}{Lin, M.}, \bibinfo{author}{Chen, Q.}, \bibinfo{author}{Yan,
  S.}, \bibinfo{year}{2013}.
\newblock \bibinfo{title}{Network in network}.
\newblock \bibinfo{journal}{International Conference on Learning
  Representations} .
\bibitem[{Liu et~al.(2018)Liu, Xu, Peng and Xiong}]{liu2018frequency}
\bibinfo{author}{Liu, Z.}, \bibinfo{author}{Xu, J.}, \bibinfo{author}{Peng,
  X.}, \bibinfo{author}{Xiong, R.}, \bibinfo{year}{2018}.
\newblock \bibinfo{title}{Frequency-domain dynamic pruning for convolutional
  neural networks}, in: \bibinfo{booktitle}{Advances in Neural Information
  Processing Systems}, pp. \bibinfo{pages}{1043--1053}.
\bibitem[{Lundberg and Lee(2017)}]{lundberg2017unified}
\bibinfo{author}{Lundberg, S.M.}, \bibinfo{author}{Lee, S.I.},
  \bibinfo{year}{2017}.
\newblock \bibinfo{title}{A unified approach to interpreting model
  predictions}, in: \bibinfo{booktitle}{Advances in Neural Information
  Processing Systems}, pp. \bibinfo{pages}{4765--4774}.
\bibitem[{Luo and Wu(2017)}]{luo2017entropy}
\bibinfo{author}{Luo, J.H.}, \bibinfo{author}{Wu, J.}, \bibinfo{year}{2017}.
\newblock \bibinfo{title}{An entropy-based pruning method for cnn compression}.
\newblock \bibinfo{journal}{arXiv preprint arXiv:1706.05791} .
\bibitem[{Maaten and Hinton(2008)}]{maaten2008visualizing}
\bibinfo{author}{Maaten, L.v.d.}, \bibinfo{author}{Hinton, G.},
  \bibinfo{year}{2008}.
\newblock \bibinfo{title}{Visualizing data using t-sne}.
\newblock \bibinfo{journal}{Journal of machine learning research}
  \bibinfo{volume}{9}, \bibinfo{pages}{2579--2605}.
\bibitem[{Madry et~al.(2018)Madry, Makelov, Schmidt, Tsipras and
  Vladu}]{madry2018towards}
\bibinfo{author}{Madry, A.}, \bibinfo{author}{Makelov, A.},
  \bibinfo{author}{Schmidt, L.}, \bibinfo{author}{Tsipras, D.},
  \bibinfo{author}{Vladu, A.}, \bibinfo{year}{2018}.
\newblock \bibinfo{title}{Towards deep learning models resistant to adversarial
  attacks}, in: \bibinfo{booktitle}{International Conference on Learning
  Representations}.
\bibitem[{Mahendran and Vedaldi(2015)}]{mahendran2015understanding}
\bibinfo{author}{Mahendran, A.}, \bibinfo{author}{Vedaldi, A.},
  \bibinfo{year}{2015}.
\newblock \bibinfo{title}{Understanding deep image representations by inverting
  them}, in: \bibinfo{booktitle}{Conference on Computer Vision and Pattern
  Recognition}, pp. \bibinfo{pages}{5188--5196}.
\bibitem[{Martinez-Gomez et~al.(2014)Martinez-Gomez, Fernandez-Caballero,
  Garcia-Varea, Rodriguez and Romero-Gonzalez}]{martinez2014taxonomy}
\bibinfo{author}{Martinez-Gomez, J.}, \bibinfo{author}{Fernandez-Caballero,
  A.}, \bibinfo{author}{Garcia-Varea, I.}, \bibinfo{author}{Rodriguez, L.},
  \bibinfo{author}{Romero-Gonzalez, C.}, \bibinfo{year}{2014}.
\newblock \bibinfo{title}{A taxonomy of vision systems for ground mobile
  robots}.
\newblock \bibinfo{journal}{International Journal of Advanced Robotic Systems}
  \bibinfo{volume}{11}, \bibinfo{pages}{111}.
\bibitem[{McInnes et~al.(2018)McInnes, Healy and Melville}]{mcinnes2018umap}
\bibinfo{author}{McInnes, L.}, \bibinfo{author}{Healy, J.},
  \bibinfo{author}{Melville, J.}, \bibinfo{year}{2018}.
\newblock \bibinfo{title}{Umap: Uniform manifold approximation and projection
  for dimension reduction}.
\newblock \bibinfo{journal}{arXiv preprint arXiv:1802.03426} .
\bibitem[{Misra et~al.(2004)Misra, Ikbal, Bourlard and
  Hermansky}]{misra2004spectral}
\bibinfo{author}{Misra, H.}, \bibinfo{author}{Ikbal, S.},
  \bibinfo{author}{Bourlard, H.}, \bibinfo{author}{Hermansky, H.},
  \bibinfo{year}{2004}.
\newblock \bibinfo{title}{Spectral entropy based feature for robust asr}, in:
  \bibinfo{booktitle}{International Conference on Acoustics, Speech, and Signal
  Processing}, \bibinfo{organization}{IEEE}. pp. \bibinfo{pages}{I--193}.
\bibitem[{Mnih et~al.(2016)Mnih, Badia, Mirza, Graves, Lillicrap, Harley,
  Silver and Kavukcuoglu}]{mnih2016asynchronous}
\bibinfo{author}{Mnih, V.}, \bibinfo{author}{Badia, A.P.},
  \bibinfo{author}{Mirza, M.}, \bibinfo{author}{Graves, A.},
  \bibinfo{author}{Lillicrap, T.}, \bibinfo{author}{Harley, T.},
  \bibinfo{author}{Silver, D.}, \bibinfo{author}{Kavukcuoglu, K.},
  \bibinfo{year}{2016}.
\newblock \bibinfo{title}{Asynchronous methods for deep reinforcement
  learning}, in: \bibinfo{booktitle}{International Conference on Machine
  Learning}, pp. \bibinfo{pages}{1928--1937}.
\bibitem[{Mnih et~al.(2015)Mnih, Kavukcuoglu, Silver, Rusu, Veness, Bellemare,
  Graves, Riedmiller, Fidjeland, Ostrovski et~al.}]{mnih2015human}
\bibinfo{author}{Mnih, V.}, \bibinfo{author}{Kavukcuoglu, K.},
  \bibinfo{author}{Silver, D.}, \bibinfo{author}{Rusu, A.A.},
  \bibinfo{author}{Veness, J.}, \bibinfo{author}{Bellemare, M.G.},
  \bibinfo{author}{Graves, A.}, \bibinfo{author}{Riedmiller, M.},
  \bibinfo{author}{Fidjeland, A.K.}, \bibinfo{author}{Ostrovski, G.}, et~al.,
  \bibinfo{year}{2015}.
\newblock \bibinfo{title}{Human-level control through deep reinforcement
  learning}.
\newblock \bibinfo{journal}{Nature} \bibinfo{volume}{518},
  \bibinfo{pages}{529}.
\bibitem[{Mohamed et~al.(2012)Mohamed, Hinton and
  Penn}]{mohamed2012understanding}
\bibinfo{author}{Mohamed, A.r.}, \bibinfo{author}{Hinton, G.},
  \bibinfo{author}{Penn, G.}, \bibinfo{year}{2012}.
\newblock \bibinfo{title}{Understanding how deep belief networks perform
  acoustic modelling}.
\newblock \bibinfo{journal}{Neural Networks} , \bibinfo{pages}{6--9}.
\bibitem[{Morch et~al.(1995)Morch, Kjems, Hansen, Svarer, Law, Lautrup,
  Strother and Rehm}]{morch1995visualization}
\bibinfo{author}{Morch, N.}, \bibinfo{author}{Kjems, U.},
  \bibinfo{author}{Hansen, L.K.}, \bibinfo{author}{Svarer, C.},
  \bibinfo{author}{Law, I.}, \bibinfo{author}{Lautrup, B.},
  \bibinfo{author}{Strother, S.}, \bibinfo{author}{Rehm, K.},
  \bibinfo{year}{1995}.
\newblock \bibinfo{title}{Visualization of neural networks using saliency
  maps}, in: \bibinfo{booktitle}{International Conference on Neural Networks},
  \bibinfo{organization}{IEEE}. pp. \bibinfo{pages}{2085--2090}.
\bibitem[{Mordvintsev et~al.(2015)Mordvintsev, Olah and
  Tyka}]{mordvintsev2015inceptionism}
\bibinfo{author}{Mordvintsev, A.}, \bibinfo{author}{Olah, C.},
  \bibinfo{author}{Tyka, M.}, \bibinfo{year}{2015}.
\newblock \bibinfo{title}{Inceptionism: Going deeper into neural networks} .
\bibitem[{Nair and Hinton(2010)}]{nair2010rectified}
\bibinfo{author}{Nair, V.}, \bibinfo{author}{Hinton, G.E.},
  \bibinfo{year}{2010}.
\newblock \bibinfo{title}{Rectified linear units improve restricted boltzmann
  machines}, in: \bibinfo{booktitle}{International Conference on Machine
  Learning}, pp. \bibinfo{pages}{807--814}.
\bibitem[{Nguyen et~al.(2015)Nguyen, Yosinski and Clune}]{nguyen2015deep}
\bibinfo{author}{Nguyen, A.}, \bibinfo{author}{Yosinski, J.},
  \bibinfo{author}{Clune, J.}, \bibinfo{year}{2015}.
\newblock \bibinfo{title}{Deep neural networks are easily fooled: High
  confidence predictions for unrecognizable images}, in:
  \bibinfo{booktitle}{Conference on Computer Vision and Pattern Recognition},
  pp. \bibinfo{pages}{427--436}.
\bibitem[{Odena et~al.(2016)Odena, Dumoulin and Olah}]{odena2016deconvolution}
\bibinfo{author}{Odena, A.}, \bibinfo{author}{Dumoulin, V.},
  \bibinfo{author}{Olah, C.}, \bibinfo{year}{2016}.
\newblock \bibinfo{title}{Deconvolution and checkerboard artifacts}.
\newblock \bibinfo{journal}{Distill} \URLprefix
  \url{http://distill.pub/2016/deconv-checkerboard},
  \DOIprefix\doi{10.23915/distill.00003}.
\bibitem[{Olah et~al.(2017)Olah, Mordvintsev and Schubert}]{olah2017feature}
\bibinfo{author}{Olah, C.}, \bibinfo{author}{Mordvintsev, A.},
  \bibinfo{author}{Schubert, L.}, \bibinfo{year}{2017}.
\newblock \bibinfo{title}{Feature visualization}.
\newblock \bibinfo{journal}{Distill} \DOIprefix\doi{10.23915/distill.00007}.
  \bibinfo{note}{https://distill.pub/2017/feature-visualization}.
\bibitem[{Packer et~al.(2018)Packer, Gao, Kos, Kr{\"a}henb{\"u}hl, Koltun and
  Song}]{packer2018assessing}
\bibinfo{author}{Packer, C.}, \bibinfo{author}{Gao, K.}, \bibinfo{author}{Kos,
  J.}, \bibinfo{author}{Kr{\"a}henb{\"u}hl, P.}, \bibinfo{author}{Koltun, V.},
  \bibinfo{author}{Song, D.}, \bibinfo{year}{2018}.
\newblock \bibinfo{title}{Assessing generalization in deep reinforcement
  learning}.
\newblock \bibinfo{journal}{arXiv preprint arXiv:1810.12282} .
\bibitem[{Pascanu et~al.(2013)Pascanu, Mikolov and
  Bengio}]{pascanu2013difficulty}
\bibinfo{author}{Pascanu, R.}, \bibinfo{author}{Mikolov, T.},
  \bibinfo{author}{Bengio, Y.}, \bibinfo{year}{2013}.
\newblock \bibinfo{title}{On the difficulty of training recurrent neural
  networks}, in: \bibinfo{booktitle}{International Conference on Machine
  Learning}, pp. \bibinfo{pages}{1310--1318}.
\bibitem[{Paszke et~al.(2017)Paszke, Gross, Chintala, Chanan, Yang, DeVito,
  Lin, Desmaison, Antiga and Lerer}]{paszke2017automatic}
\bibinfo{author}{Paszke, A.}, \bibinfo{author}{Gross, S.},
  \bibinfo{author}{Chintala, S.}, \bibinfo{author}{Chanan, G.},
  \bibinfo{author}{Yang, E.}, \bibinfo{author}{DeVito, Z.},
  \bibinfo{author}{Lin, Z.}, \bibinfo{author}{Desmaison, A.},
  \bibinfo{author}{Antiga, L.}, \bibinfo{author}{Lerer, A.},
  \bibinfo{year}{2017}.
\newblock \bibinfo{title}{Automatic differentiation in pytorch} .
\bibitem[{Pearson(1901)}]{pearson1901liii}
\bibinfo{author}{Pearson, K.}, \bibinfo{year}{1901}.
\newblock \bibinfo{title}{Liii. on lines and planes of closest fit to systems
  of points in space}.
\newblock \bibinfo{journal}{The London, Edinburgh, and Dublin Philosophical
  Magazine and Journal of Science} \bibinfo{volume}{2},
  \bibinfo{pages}{559--572}.
\bibitem[{Peng et~al.(2018)Peng, Andrychowicz, Zaremba and
  Abbeel}]{peng2018sim}
\bibinfo{author}{Peng, X.B.}, \bibinfo{author}{Andrychowicz, M.},
  \bibinfo{author}{Zaremba, W.}, \bibinfo{author}{Abbeel, P.},
  \bibinfo{year}{2018}.
\newblock \bibinfo{title}{Sim-to-real transfer of robotic control with dynamics
  randomization}, in: \bibinfo{booktitle}{International Conference on Robotics
  and Automation}, \bibinfo{organization}{IEEE}. pp. \bibinfo{pages}{1--8}.
\bibitem[{Pinto et~al.(2017)Pinto, Davidson, Sukthankar and
  Gupta}]{pinto2017robust}
\bibinfo{author}{Pinto, L.}, \bibinfo{author}{Davidson, J.},
  \bibinfo{author}{Sukthankar, R.}, \bibinfo{author}{Gupta, A.},
  \bibinfo{year}{2017}.
\newblock \bibinfo{title}{Robust adversarial reinforcement learning}, in:
  \bibinfo{booktitle}{International Conference on Machine Learning}, pp.
  \bibinfo{pages}{2817--2826}.
\bibitem[{Plappert et~al.(2018)Plappert, Andrychowicz, Ray, McGrew, Baker,
  Powell, Schneider, Tobin, Chociej, Welinder et~al.}]{plappert2018multi}
\bibinfo{author}{Plappert, M.}, \bibinfo{author}{Andrychowicz, M.},
  \bibinfo{author}{Ray, A.}, \bibinfo{author}{McGrew, B.},
  \bibinfo{author}{Baker, B.}, \bibinfo{author}{Powell, G.},
  \bibinfo{author}{Schneider, J.}, \bibinfo{author}{Tobin, J.},
  \bibinfo{author}{Chociej, M.}, \bibinfo{author}{Welinder, P.}, et~al.,
  \bibinfo{year}{2018}.
\newblock \bibinfo{title}{Multi-goal reinforcement learning: Challenging
  robotics environments and request for research}.
\newblock \bibinfo{journal}{arXiv preprint arXiv:1802.09464} .
\bibitem[{Rauber et~al.(2017)Rauber, Fadel, Falcao and
  Telea}]{rauber2017visualizing}
\bibinfo{author}{Rauber, P.E.}, \bibinfo{author}{Fadel, S.G.},
  \bibinfo{author}{Falcao, A.X.}, \bibinfo{author}{Telea, A.C.},
  \bibinfo{year}{2017}.
\newblock \bibinfo{title}{Visualizing the hidden activity of artificial neural
  networks}.
\newblock \bibinfo{journal}{IEEE Transactions on Visualization and Computer
  Graphics} \bibinfo{volume}{23}, \bibinfo{pages}{101--110}.
\bibitem[{Reed(1993)}]{reed1993pruning}
\bibinfo{author}{Reed, R.}, \bibinfo{year}{1993}.
\newblock \bibinfo{title}{Pruning algorithms-a survey}.
\newblock \bibinfo{journal}{IEEE Transactions on Neural Networks}
  \bibinfo{volume}{4}, \bibinfo{pages}{740--747}.
\bibitem[{Ribeiro et~al.(2016)Ribeiro, Singh and Guestrin}]{ribeiro2016should}
\bibinfo{author}{Ribeiro, M.T.}, \bibinfo{author}{Singh, S.},
  \bibinfo{author}{Guestrin, C.}, \bibinfo{year}{2016}.
\newblock \bibinfo{title}{Why should i trust you?: Explaining the predictions
  of any classifier}, in: \bibinfo{booktitle}{International Conference on
  Knowledge Discovery and Data Mining}, \bibinfo{organization}{ACM}. pp.
  \bibinfo{pages}{1135--1144}.
\bibitem[{Rusu et~al.(2017)Rusu, Ve{\v{c}}er{\'\i}k, Roth{\"o}rl, Heess,
  Pascanu and Hadsell}]{rusu2017sim}
\bibinfo{author}{Rusu, A.A.}, \bibinfo{author}{Ve{\v{c}}er{\'\i}k, M.},
  \bibinfo{author}{Roth{\"o}rl, T.}, \bibinfo{author}{Heess, N.},
  \bibinfo{author}{Pascanu, R.}, \bibinfo{author}{Hadsell, R.},
  \bibinfo{year}{2017}.
\newblock \bibinfo{title}{Sim-to-real robot learning from pixels with
  progressive nets}, in: \bibinfo{booktitle}{Conference on Robot Learning}, pp.
  \bibinfo{pages}{262--270}.
\bibitem[{Sadeghi and Levine(2017)}]{sadeghi2017cad2rl}
\bibinfo{author}{Sadeghi, F.}, \bibinfo{author}{Levine, S.},
  \bibinfo{year}{2017}.
\newblock \bibinfo{title}{Cad2rl: Real single-image flight without a single
  real image}, in: \bibinfo{booktitle}{Robotics: Science and Systems}.
\bibitem[{Salakhutdinov and Hinton(2007)}]{salakhutdinov2007learning}
\bibinfo{author}{Salakhutdinov, R.}, \bibinfo{author}{Hinton, G.},
  \bibinfo{year}{2007}.
\newblock \bibinfo{title}{Learning a nonlinear embedding by preserving class
  neighbourhood structure}, in: \bibinfo{booktitle}{Artificial Intelligence and
  Statistics}, pp. \bibinfo{pages}{412--419}.
\bibitem[{Saxe et~al.(2014)Saxe, McClelland and Ganguli}]{saxe2014exact}
\bibinfo{author}{Saxe, A.M.}, \bibinfo{author}{McClelland, J.L.},
  \bibinfo{author}{Ganguli, S.}, \bibinfo{year}{2014}.
\newblock \bibinfo{title}{Exact solutions to the nonlinear dynamics of learning
  in deep linear neural networks}, in: \bibinfo{booktitle}{International
  Conference on Learning Representations}.
\bibitem[{Schulman et~al.(2015)Schulman, Moritz, Levine, Jordan and
  Abbeel}]{schulman2015high}
\bibinfo{author}{Schulman, J.}, \bibinfo{author}{Moritz, P.},
  \bibinfo{author}{Levine, S.}, \bibinfo{author}{Jordan, M.},
  \bibinfo{author}{Abbeel, P.}, \bibinfo{year}{2015}.
\newblock \bibinfo{title}{High-dimensional continuous control using generalized
  advantage estimation}.
\newblock \bibinfo{journal}{arXiv preprint arXiv:1506.02438} .
\bibitem[{Schulman et~al.(2017)Schulman, Wolski, Dhariwal, Radford and
  Klimov}]{schulman2017proximal}
\bibinfo{author}{Schulman, J.}, \bibinfo{author}{Wolski, F.},
  \bibinfo{author}{Dhariwal, P.}, \bibinfo{author}{Radford, A.},
  \bibinfo{author}{Klimov, O.}, \bibinfo{year}{2017}.
\newblock \bibinfo{title}{Proximal policy optimization algorithms}.
\newblock \bibinfo{journal}{arXiv preprint arXiv:1707.06347} .
\bibitem[{Selvaraju et~al.(2017)Selvaraju, Cogswell, Das, Vedantam, Parikh and
  Batra}]{selvaraju2017grad}
\bibinfo{author}{Selvaraju, R.R.}, \bibinfo{author}{Cogswell, M.},
  \bibinfo{author}{Das, A.}, \bibinfo{author}{Vedantam, R.},
  \bibinfo{author}{Parikh, D.}, \bibinfo{author}{Batra, D.},
  \bibinfo{year}{2017}.
\newblock \bibinfo{title}{Grad-cam: Visual explanations from deep networks via
  gradient-based localization}, in: \bibinfo{booktitle}{International
  Conference on Computer Vision}, pp. \bibinfo{pages}{618--626}.
\bibitem[{Shrikumar et~al.(2017)Shrikumar, Greenside and
  Kundaje}]{shrikumar2017learning}
\bibinfo{author}{Shrikumar, A.}, \bibinfo{author}{Greenside, P.},
  \bibinfo{author}{Kundaje, A.}, \bibinfo{year}{2017}.
\newblock \bibinfo{title}{Learning important features through propagating
  activation differences}, in: \bibinfo{booktitle}{International Conference on
  Machine Learning}, \bibinfo{organization}{JMLR. org}. pp.
  \bibinfo{pages}{3145--3153}.
\bibitem[{Sietsma and Dow(1988)}]{sietsma1988neural}
\bibinfo{author}{Sietsma, J.}, \bibinfo{author}{Dow, R.J.},
  \bibinfo{year}{1988}.
\newblock \bibinfo{title}{Neural net pruning - why and how}, in:
  \bibinfo{booktitle}{International Conference on Neural Networks}, pp.
  \bibinfo{pages}{325--333}.
\bibitem[{Sietsma and Dow(1991)}]{sietsma1991creating}
\bibinfo{author}{Sietsma, J.}, \bibinfo{author}{Dow, R.J.},
  \bibinfo{year}{1991}.
\newblock \bibinfo{title}{Creating artificial neural networks that generalize}.
\newblock \bibinfo{journal}{Neural Networks} \bibinfo{volume}{4},
  \bibinfo{pages}{67--79}.
\bibitem[{Simonyan et~al.(2013)Simonyan, Vedaldi and
  Zisserman}]{simonyan2013deep}
\bibinfo{author}{Simonyan, K.}, \bibinfo{author}{Vedaldi, A.},
  \bibinfo{author}{Zisserman, A.}, \bibinfo{year}{2013}.
\newblock \bibinfo{title}{Deep inside convolutional networks: Visualising image
  classification models and saliency maps}.
\newblock \bibinfo{journal}{arXiv preprint arXiv:1312.6034} .
\bibitem[{Springenberg et~al.(2015)Springenberg, Dosovitskiy, Brox and
  Riedmiller}]{springenberg2015striving}
\bibinfo{author}{Springenberg, J.}, \bibinfo{author}{Dosovitskiy, A.},
  \bibinfo{author}{Brox, T.}, \bibinfo{author}{Riedmiller, M.},
  \bibinfo{year}{2015}.
\newblock \bibinfo{title}{Striving for simplicity: The all convolutional net},
  in: \bibinfo{booktitle}{International Conference on Learning Representations
  (Workshop Track)}.
\bibitem[{Srinivasan et~al.(2005)Srinivasan, Eswaran, Sriraam and
  N}]{srinivasan2005artificial}
\bibinfo{author}{Srinivasan, V.}, \bibinfo{author}{Eswaran, C.},
  \bibinfo{author}{Sriraam}, \bibinfo{author}{N}, \bibinfo{year}{2005}.
\newblock \bibinfo{title}{Artificial neural network based epileptic detection
  using time-domain and frequency-domain features}.
\newblock \bibinfo{journal}{Journal of Medical Systems} \bibinfo{volume}{29},
  \bibinfo{pages}{647--660}.
\bibitem[{Sturmfels et~al.(2020)Sturmfels, Lundberg and
  Lee}]{sturmfels2020visualizing}
\bibinfo{author}{Sturmfels, P.}, \bibinfo{author}{Lundberg, S.},
  \bibinfo{author}{Lee, S.I.}, \bibinfo{year}{2020}.
\newblock \bibinfo{title}{Visualizing the impact of feature attribution
  baselines}.
\newblock \bibinfo{journal}{Distill} \bibinfo{volume}{5}, \bibinfo{pages}{e21}.
\bibitem[{Such et~al.(2018)Such, Madhavan, Liu, Wang, Castro, Li, Schubert,
  Bellemare, Clune and Lehman}]{such2018atari}
\bibinfo{author}{Such, F.}, \bibinfo{author}{Madhavan, V.},
  \bibinfo{author}{Liu, R.}, \bibinfo{author}{Wang, R.},
  \bibinfo{author}{Castro, P.}, \bibinfo{author}{Li, Y.},
  \bibinfo{author}{Schubert, L.}, \bibinfo{author}{Bellemare, M.G.},
  \bibinfo{author}{Clune, J.}, \bibinfo{author}{Lehman, J.},
  \bibinfo{year}{2018}.
\newblock \bibinfo{title}{An atari model zoo for analyzing, visualizing, and
  comparing deep reinforcement learning agents}, in:
  \bibinfo{booktitle}{NeurIPS Deep RL Workshop}.
\bibitem[{Sundararajan et~al.(2017)Sundararajan, Taly and
  Yan}]{sundararajan2017axiomatic}
\bibinfo{author}{Sundararajan, M.}, \bibinfo{author}{Taly, A.},
  \bibinfo{author}{Yan, Q.}, \bibinfo{year}{2017}.
\newblock \bibinfo{title}{Axiomatic attribution for deep networks}, in:
  \bibinfo{booktitle}{International Conference on Machine Learning},
  \bibinfo{organization}{JMLR. org}. pp. \bibinfo{pages}{3319--3328}.
\bibitem[{Sutton and Barto(2018)}]{sutton2018reinforcement}
\bibinfo{author}{Sutton, R.S.}, \bibinfo{author}{Barto, A.G.},
  \bibinfo{year}{2018}.
\newblock \bibinfo{title}{Reinforcement learning: An introduction}.
\newblock \bibinfo{publisher}{MIT press}.
\bibitem[{Tobin et~al.(2017)Tobin, Fong, Ray, Schneider, Zaremba and
  Abbeel}]{tobin2017domain}
\bibinfo{author}{Tobin, J.}, \bibinfo{author}{Fong, R.}, \bibinfo{author}{Ray,
  A.}, \bibinfo{author}{Schneider, J.}, \bibinfo{author}{Zaremba, W.},
  \bibinfo{author}{Abbeel, P.}, \bibinfo{year}{2017}.
\newblock \bibinfo{title}{Domain randomization for transferring deep neural
  networks from simulation to the real world}, in:
  \bibinfo{booktitle}{International Conference on Intelligent Robots and
  Systems}, \bibinfo{organization}{IEEE}. pp. \bibinfo{pages}{23--30}.
\bibitem[{Todorov et~al.(2012)Todorov, Erez and Tassa}]{todorov2012mujoco}
\bibinfo{author}{Todorov, E.}, \bibinfo{author}{Erez, T.},
  \bibinfo{author}{Tassa, Y.}, \bibinfo{year}{2012}.
\newblock \bibinfo{title}{Mujoco: A physics engine for model-based control},
  in: \bibinfo{booktitle}{International Conference on Intelligent Robots and
  Systems}, \bibinfo{organization}{IEEE}. pp. \bibinfo{pages}{5026--5033}.
\bibitem[{Tzeng et~al.(2015)Tzeng, Devin, Hoffman, Finn, Peng, Levine, Saenko
  and Darrell}]{tzeng2015towards}
\bibinfo{author}{Tzeng, E.}, \bibinfo{author}{Devin, C.},
  \bibinfo{author}{Hoffman, J.}, \bibinfo{author}{Finn, C.},
  \bibinfo{author}{Peng, X.}, \bibinfo{author}{Levine, S.},
  \bibinfo{author}{Saenko, K.}, \bibinfo{author}{Darrell, T.},
  \bibinfo{year}{2015}.
\newblock \bibinfo{title}{Towards adapting deep visuomotor representations from
  simulated to real environments}.
\newblock \bibinfo{journal}{arXiv preprint arXiv:1511.07111}
  \bibinfo{volume}{2}.
\bibitem[{Uesato et~al.(2018)Uesato, Kumar, Szepesvari, Erez, Ruderman,
  Anderson, Heess, Kohli et~al.}]{uesato2018rigorous}
\bibinfo{author}{Uesato, J.}, \bibinfo{author}{Kumar, A.},
  \bibinfo{author}{Szepesvari, C.}, \bibinfo{author}{Erez, T.},
  \bibinfo{author}{Ruderman, A.}, \bibinfo{author}{Anderson, K.},
  \bibinfo{author}{Heess, N.}, \bibinfo{author}{Kohli, P.}, et~al.,
  \bibinfo{year}{2018}.
\newblock \bibinfo{title}{Rigorous agent evaluation: An adversarial approach to
  uncover catastrophic failures}.
\newblock \bibinfo{journal}{arXiv preprint arXiv:1812.01647} .
\bibitem[{Wattenberg et~al.(2016)Wattenberg, Vi{\'e}gas and
  Johnson}]{wattenberg2016use}
\bibinfo{author}{Wattenberg, M.}, \bibinfo{author}{Vi{\'e}gas, F.},
  \bibinfo{author}{Johnson, I.}, \bibinfo{year}{2016}.
\newblock \bibinfo{title}{How to use t-sne effectively}.
\newblock \bibinfo{journal}{Distill} \bibinfo{volume}{1}, \bibinfo{pages}{e2}.
\bibitem[{Wierstra et~al.(2007)Wierstra, Foerster, Peters and
  Schmidhuber}]{wierstra2007solving}
\bibinfo{author}{Wierstra, D.}, \bibinfo{author}{Foerster, A.},
  \bibinfo{author}{Peters, J.}, \bibinfo{author}{Schmidhuber, J.},
  \bibinfo{year}{2007}.
\newblock \bibinfo{title}{Solving deep memory pomdps with recurrent policy
  gradients}, in: \bibinfo{booktitle}{International Conference on Artificial
  Neural Networks}, \bibinfo{organization}{Springer}. pp.
  \bibinfo{pages}{697--706}.
\bibitem[{Williams and Peng(1991)}]{williams1991function}
\bibinfo{author}{Williams, R.J.}, \bibinfo{author}{Peng, J.},
  \bibinfo{year}{1991}.
\newblock \bibinfo{title}{Function optimization using connectionist
  reinforcement learning algorithms}.
\newblock \bibinfo{journal}{Connection Science} \bibinfo{volume}{3},
  \bibinfo{pages}{241--268}.
\bibitem[{Witty et~al.(2018)Witty, Lee, Tosch, Atrey, Littman and
  Jensen}]{witty2018measuring}
\bibinfo{author}{Witty, S.}, \bibinfo{author}{Lee, J.K.},
  \bibinfo{author}{Tosch, E.}, \bibinfo{author}{Atrey, A.},
  \bibinfo{author}{Littman, M.}, \bibinfo{author}{Jensen, D.},
  \bibinfo{year}{2018}.
\newblock \bibinfo{title}{Measuring and characterizing generalization in deep
  reinforcement learning}.
\newblock \bibinfo{journal}{arXiv preprint arXiv:1812.02868} .
\bibitem[{Yamins and DiCarlo(2016)}]{yamins2016using}
\bibinfo{author}{Yamins, D.L.}, \bibinfo{author}{DiCarlo, J.J.},
  \bibinfo{year}{2016}.
\newblock \bibinfo{title}{Using goal-driven deep learning models to understand
  sensory cortex}.
\newblock \bibinfo{journal}{Nature Neuroscience} \bibinfo{volume}{19},
  \bibinfo{pages}{356}.
\bibitem[{Zeiler and Fergus(2014)}]{zeiler2014visualizing}
\bibinfo{author}{Zeiler, M.D.}, \bibinfo{author}{Fergus, R.},
  \bibinfo{year}{2014}.
\newblock \bibinfo{title}{Visualizing and understanding convolutional
  networks}, in: \bibinfo{booktitle}{European Conference on Computer Vision},
  \bibinfo{organization}{Springer}. pp. \bibinfo{pages}{818--833}.
\bibitem[{Zhang et~al.(2018a)Zhang, Ballas and Pineau}]{zhang2018dissection}
\bibinfo{author}{Zhang, A.}, \bibinfo{author}{Ballas, N.},
  \bibinfo{author}{Pineau, J.}, \bibinfo{year}{2018}a.
\newblock \bibinfo{title}{A dissection of overfitting and generalization in
  continuous reinforcement learning}.
\newblock \bibinfo{journal}{arXiv preprint arXiv:1806.07937} .
\bibitem[{Zhang et~al.(2019)Zhang, Bengio and Singer}]{zhang2019all}
\bibinfo{author}{Zhang, C.}, \bibinfo{author}{Bengio, S.},
  \bibinfo{author}{Singer, Y.}, \bibinfo{year}{2019}.
\newblock \bibinfo{title}{Are all layers created equal?}
\newblock \bibinfo{journal}{arXiv preprint arXiv:1902.01996} .
\bibitem[{Zhang et~al.(2018b)Zhang, Vinyals, Munos and Bengio}]{zhang2018study}
\bibinfo{author}{Zhang, C.}, \bibinfo{author}{Vinyals, O.},
  \bibinfo{author}{Munos, R.}, \bibinfo{author}{Bengio, S.},
  \bibinfo{year}{2018}b.
\newblock \bibinfo{title}{A study on overfitting in deep reinforcement
  learning}.
\newblock \bibinfo{journal}{arXiv preprint arXiv:1804.06893} .
\bibitem[{Zhao et~al.(2019)Zhao, Siguad, Stulp and
  Hospedales}]{zhao2019investigating}
\bibinfo{author}{Zhao, C.}, \bibinfo{author}{Siguad, O.},
  \bibinfo{author}{Stulp, F.}, \bibinfo{author}{Hospedales, T.M.},
  \bibinfo{year}{2019}.
\newblock \bibinfo{title}{Investigating generalisation in continuous deep
  reinforcement learning}.
\newblock \bibinfo{journal}{arXiv preprint arXiv:1902.07015} .
\bibitem[{Zheng et~al.(1996)Zheng, Qian and Clarke}]{zheng1996digital}
\bibinfo{author}{Zheng, B.}, \bibinfo{author}{Qian, W.},
  \bibinfo{author}{Clarke, L.P.}, \bibinfo{year}{1996}.
\newblock \bibinfo{title}{Digital mammography: mixed feature neural network
  with spectral entropy decision for detection of microcalcifications}.
\newblock \bibinfo{journal}{IEEE Transactions on Medical Imaging}
  \bibinfo{volume}{15}, \bibinfo{pages}{589--597}.
\bibitem[{Zhou et~al.(2016)Zhou, Khosla, Lapedriza, Oliva and
  Torralba}]{zhou2016learning}
\bibinfo{author}{Zhou, B.}, \bibinfo{author}{Khosla, A.},
  \bibinfo{author}{Lapedriza, A.}, \bibinfo{author}{Oliva, A.},
  \bibinfo{author}{Torralba, A.}, \bibinfo{year}{2016}.
\newblock \bibinfo{title}{Learning deep features for discriminative
  localization}, in: \bibinfo{booktitle}{Conference on Computer Vision and
  Pattern Recognition}, pp. \bibinfo{pages}{2921--2929}.
\bibitem[{Zhu et~al.(2017)Zhu, Mottaghi, Kolve, Lim, Gupta, Fei-Fei and
  Farhadi}]{zhu2017target}
\bibinfo{author}{Zhu, Y.}, \bibinfo{author}{Mottaghi, R.},
  \bibinfo{author}{Kolve, E.}, \bibinfo{author}{Lim, J.J.},
  \bibinfo{author}{Gupta, A.}, \bibinfo{author}{Fei-Fei, L.},
  \bibinfo{author}{Farhadi, A.}, \bibinfo{year}{2017}.
\newblock \bibinfo{title}{Target-driven visual navigation in indoor scenes
  using deep reinforcement learning}, in: \bibinfo{booktitle}{International
  Conference on Robotics and Automation}, \bibinfo{organization}{IEEE}. pp.
  \bibinfo{pages}{3357--3364}.

\end{thebibliography}


%

\end{document}